%% file: icml2020arxiv.tex
\documentclass{article}
\usepackage{multicol}  

\usepackage{microtype}
\usepackage{graphicx}
\usepackage{subfigure}
\usepackage{booktabs} 
\usepackage{hyperref}
\usepackage{url}
\usepackage[utf8]{inputenc} 
\usepackage[T1]{fontenc}    
\usepackage{wrapfig,lipsum}
\usepackage{graphics} 
\usepackage{epsfig} 
\usepackage{mathptmx} 
\usepackage{caption}
\usepackage{blindtext}
\usepackage{times} 
\usepackage{amsmath} 
\usepackage{nicefrac}       
\usepackage{microtype}      
\usepackage{epstopdf}
\usepackage{color}
\usepackage{txfonts}
\usepackage{amssymb}
\usepackage{glossaries}
\AppendGraphicsExtensions{.tiff}
\AppendGraphicsExtensions{.tif}

\usepackage[accepted]{icml2020}

\icmltitlerunning{Estimating Model Uncertainty of Neural Networks in Sparse Information Form}

\begin{document}

\twocolumn[
\icmltitle{Estimating Model Uncertainty of Neural Networks  \\ in Sparse Information Form}




\begin{icmlauthorlist}
\icmlauthor{Jongseok Lee}{to}
\icmlauthor{Matthias Humt}{to}
\icmlauthor{Jianxiang Feng}{to,goo}
\icmlauthor{Rudolph Triebel}{to,goo}
\end{icmlauthorlist}

\icmlaffiliation{to}{Institute of Robotics and Mechatronics, German Aerospace Center (DLR), Wessling, Germany}
\icmlaffiliation{goo}{Computer Vision Group, Technical University of Munich (TU Munich), Garching, Germany}

\icmlcorrespondingauthor{Jongseok Lee}{jongseok.lee@dlr.de}

\icmlkeywords{Bayesian Deep Learning, Laplace Approximation, Low Rank Approximation, Model Uncertainty, Information Theory}

\vskip 0.3in
]



\printAffiliationsAndNotice{}  

\begin{abstract}
We present a sparse representation of model uncertainty for Deep Neural Networks (DNNs) where the parameter posterior is approximated with an inverse formulation of the Multivariate Normal Distribution (MND), also known as the \emph{information form}. The key insight of our work is that the information matrix, i.e. the inverse of the covariance matrix tends to be sparse in its spectrum. Therefore, dimensionality reduction techniques such as low rank approximations (LRA) can be effectively exploited. To achieve this, we develop a novel sparsification algorithm and derive a cost-effective analytical sampler. As a result, we show that the information form can be scalably applied to represent model uncertainty in DNNs. Our exhaustive theoretical analysis and empirical evaluations on various benchmarks show the competitiveness of our approach over the current methods.
\end{abstract}

\input{chapters/introduction.tex}
\input{chapters/background.tex}
\input{chapters/methods.tex}
\input{chapters/relatedworks.tex}
\input{chapters/results.tex}
\input{chapters/results_toy_uci.tex}

\input{chapters/results_active.tex}
\input{chapters/results_bdl.tex}
\input{chapters/results_imagenet.tex}
\input{chapters/conclusion.tex}

\section*{Acknowledgements}
We thank the anonymous reviewers and the area chairs for their time and thoughtful comments. Special thanks to Konstantin Kondak for many inspiring discussions, and Klaus Strobl for many pointers to the related works. The authors acknowledge the support of Helmholtz Association, the project ARCHES (contract number ZT-0033) and the EU-project AUTOPILOT (contract number 731993). Jianxiang Feng is supported by the Munich School for Data Science (MUDS) and Rudolph Triebel is a member of MUDS.


\bibliography{reference.bib}
\bibliographystyle{icml2020.bst}

\cleardoublepage

\appendix
\input{supplementary/organization.tex}
\input{supplementary/derivations.tex}
\input{supplementary/proofs.tex}
\input{supplementary/implementation_details.tex}
\input{supplementary/further_results.tex}

\end{document}

%% file: chapters/introduction.tex
\section{Introduction}
Whenever machine learning methods are used for safety-critical applications such as autonomous driving, it is crucial to provide a precise estimation of the failure probability of the learned predictor. Therefore, most of the current learning approaches return distributions rather than single, most-likely predictions. However, in case of DNNs, this true failure probability tends to be severely underestimated, leading to \emph{overconfident} predictions \citep{guo17calib}. The main reason for this is that DNNs are typically trained with a principle of \emph{maximum likelihood}, neglecting their \emph{epistemic} or model uncertainty with the point estimates of parameters. 

Imposing Gaussians on model uncertainty is arguably the most popular choice as Gaussians are for \textit{approximate inference} what linear maps are for algebra. For example, once the \textit{posterior distribution} is inferred, the majority of computations can be performed using the well known tools of linear algebra. For DNNs however, the space complexity of using MNDs is intractable as the covariance matrix scales quadratic to the number of parameters. Consequently, approximate inference on DNNs posterior often neglects the parameter correlations \citep{wu2018deterministic, NIPS2015_5666, Alex2011, Hernandez2015} or simplifies the covariance matrix into Kronecker products of two smaller matrices \citep{pmlr-v54-sun17b, louizos16, zhang2017noisy, Park2019} regardless of the inference principles such as variational inference.

Instead, inspired by \citet{Thrun03d}, we advocate to explore the dual and inverse formulation of MNDs:
%
%
%
\begin{align*} 
\bar{x} & \propto \text{exp}\left \{ -\frac{1}{2}(\bar{x}-\mu)^T\Sigma^{-1}(\bar{x}-\mu) \right \} \\
 &= \text{exp}\left \{ -\frac{1}{2}\bar{x}^T\Sigma^{-1} \bar{x} + \mu^T\Sigma^{-1} \bar{x} \right \} \\
 &= \text{exp}\left \{ -\frac{1}{2}\bar{x}^T\boldsymbol{I} \bar{x} + \mu^{IV} \bar{x} \right \}  \text{  or  } \bar{x} \sim \mathcal{N}^{-1}(\mu^{IV}, \boldsymbol{I})
\end{align*}

where the Gaussian random variable $\bar{x}$ is fully parameterized by the information vector $\mu^{IV}$ and matrix $\boldsymbol{I}$ as opposed to mean $\mu$ and covariance matrix $\Sigma$. Our major findings are that this so-called information form has important ramifications on developing scalable \textit{Bayesian Neural Networks}. Firstly, we point out that the approximate inference for this formulation can be simplified to scalable Laplace Approximation (LA) \citep{MacKay1992, Ritter2017ASL}, in which we improve the state-of-the-art Kronecker factored approximations of the information matrix \citep{George2018} by correcting the diagonal variance in parameter space.

More importantly, DNNs offer a natural spectral sparsity in the information matrix \citep{SagunEGDB18} as oppose to the covariance matrix. Intuitively, the information content of each parameters become weaker with increasing number of parameters and thus, sparse representations can be effectively exploited. To do so, we propose a novel low-rank representation of the given Kronecker factorization and devise a spectral sparsification algorithm that can preserve the Kronecker product in its eigenbasis. Based on this formulation, we further demonstrate low rank sampling computations which significantly reduces the space complexity of MND from $O(N^3)$ to $O(L^3)$ where L is the chosen low-rank dimension instead of parameter space lying in high dimensional N manifolds. Lastly, we also exhaustively perform both theoretical and empirical evaluations, yielding state-of-the-art results in both scalability and performance. 

Our main contribution is a novel sparse representation for DNNs posterior that is backed up by scalable mathematical foot-works - more specifically: (i) an approximate inference that estimates model uncertainty in information form (section \ref{sec:la}), (ii) a low-rank representation of Kronecker factored eigendecomposition (section \ref{sec:lra}), (iii) an algorithm to enable a LRA for the given representation of MNDs (algorithm \ref{algorithm1}) and (iv) derivation of a memory-wise tractable sampler (section \ref{sec:sampling:main}). With our theoretical (section \ref{sec:method:theory}) and experimental results (section \ref{sec:result}) we further showcase the state-of-the-art performance. Finally, a plug-in-and-play code is attached for enabling adoptions in practice.

%% file: chapters/background.tex
\section{Methodology}
\label{sec:method}
\subsection{Background and Notation}
\label{sec:background}
A neural network is a parameterized function $f_{\theta}:\mathbb{R}^{N_1}\rightarrow \mathbb{R}^{N_l}$ where $\theta \in \mathbb{R}^{N_{\theta}}$ are the weights and $N_\theta = N_1 + \dots + N_l$. 
This function $f_{\theta}$ is in fact a concatenation of $l$ layers, where each layer $\mathfrak{i} \in \left \{ 1, ..., l \right \}$ computes $h_\mathfrak{i} = W_\mathfrak{i} a_{\mathfrak{i}-1}$ and $a_\mathfrak{i} = \phi (h_{\mathfrak{i}-1})$. 
Here, $\phi$ is a nonlinear function, $a_\mathfrak{i}$ are activations, $h_\mathfrak{i}$ linear pre-activations, and $W_\mathfrak{i}$ are weight matrices. The bias terms are absorbed into $W_\mathfrak{i}$ by appending $1$ to each $a_\mathfrak{i}$. 
Thus, $\theta = \begin{bmatrix} vec(W_1)^{T} & ... & vec(W_l)^{T} \end{bmatrix}^{T}$ where $vec$ is the operator that stacks the columns of a matrix to a vector. Let $g_\mathfrak{i} = \delta h_\mathfrak{i}$, the gradient of $h_\mathfrak{i}$ w.r.t $\theta$. Using LA the posterior is approximated with a Gaussian. 
The mean is then given by the MAP estimate $\theta_{MAP}$ and the covariance by the Hessian of the log-likelihood $(H+\tau I)^{-1}$ assuming a Gaussian prior with precision $\tau$. 
Using loss functions such as MSE or cross entropy and piece-wise linear activation $a_\mathfrak{i}$ (e.g RELU), a good approximation of the Hessian is the Fisher information matrix (IM) $\boldsymbol{I} = \mathbb{E}\left [ \delta \theta \delta \theta^T \right ]$ for the backpropagated gradients $\delta \theta $ \footnote{The expectation herein is defined with respect to the paramerterized density function $p_\theta(y|x)$ assuming i.i.d. samples x.} and is typically scaled by the number of data points $\textsc{N}$ \citep{MartensG15}. 
IM is of size $N_{\theta}\times N_{\theta}$ resulting in too large matrix for moderately sized DNNs.

To make the computation tractable, it is first assumed that the weights across layers are uncorrelated, which corresponds to a block-diagonal form of $\boldsymbol{I}$ with blocks $\boldsymbol{I}_1, \dots, \boldsymbol{I}_l$. Then, each realisation of block $\boldsymbol{I}_i$ is represented as a Kronecker product $\delta \theta_\mathfrak{i} \delta \theta_\mathfrak{i}^T = a_{\mathfrak{i}-1}a_{\mathfrak{i}-1}^T \otimes g_\mathfrak{i}g_\mathfrak{i}^T$. Then, matrices $A_{\mathfrak{i}-1}$ and $G_\mathfrak{i}$ are assumed to be statistically independent:

\begin{equation}
\label{eq:3:4}
\boldsymbol{I}_{\mathfrak{i}} \approx \boldsymbol{I}_{\mathfrak{i}, {\text{kfac}}} = \mathbb{E}\left [a_{\mathfrak{i}-1}a_{\mathfrak{i}-1}^T\right ]\otimes \mathbb{E}\left [g_\mathfrak{i}g_\mathfrak{i}^T\right ] = A_{\mathfrak{i}-1}  \otimes G_{\mathfrak{i}}.  
\end{equation}

We refer to \citet{MartensG15} for details on KFAC. Here, $A_{\mathfrak{i}-1} \in \mathbb{R}^{n_\mathfrak{i} \times n_\mathfrak{i}}$ and $G_\mathfrak{i} \in \mathbb{R}^{m_\mathfrak{i} \times m_\mathfrak{i}}$, where the number of weights is $N_\mathfrak{i}=n_\mathfrak{i}m_\mathfrak{i}$. Typically IM is scaled by the number of data points $\textsc{N}$ and incorporates the prior $\tau$. The herein presented parameter posterior omits the addition of prior precision and scaling term for simplicity. Here, $\textsc{N}$ and $\tau$ are treated as hyperparameters \citep{Ritter2017ASL} similar to tempering in \citep{wenzel2020good}. KFAC scales to big data sets such as ImageNet \citep{Krizhevsky2012} with large DNNs \citep{Ba2017} and does not require changes in the training procedure when used for LA \citep{Ritter2017ASL}.

%% file: chapters/methods.tex
\subsection{Approximate Inference in Information Form}
\label{sec:la}
We first employ an eigenvalue correction in the Kronecker factored eigenbasis \citep{George2018} for LA. Layer indices $\mathfrak{i}$ are omitted and explanation applies layer-wise.

Let $\boldsymbol{I}=V_{\text{true}}\Lambda_{\text{true}} V_{\text{true}}^T$ be the true eigendecomposition of IM per layer. 
From this it follows $\Lambda_{\text{true}} = V_{\text{ture}}^{T}\boldsymbol{I}V_{\text{true}}= \mathbb{E}\left [ V_{\text{true}}^T\delta \theta \delta \theta^T V_{\text{true}}\right]$ and $\Lambda_{\text{true},ii} = \mathbb{E}\left [ (V_{\text{true}}^T \delta \theta)_{i}^2 \right ]$ where $i \in \left \{ 1, \cdots, N \right \}$ and $N$ is the number of parameters of this layer. 
Defining the eigendecomposition of $A$ and $G$ in \eqref{eq:3:4} as $A=U_AS_AU_A^T$ and $G=U_GS_GU_G^T$, it further follows $\boldsymbol{I}_{\text{kfac}} \approx A \otimes G = (U_A \otimes U_G)(S_A \otimes S_G)(U_A \otimes U_G)^T$ from the properties of the Kronecker product. 
Now, this approximation can be improved by replacing $(S_A \otimes S_G)$ with the eigenvalues $\Lambda_{\text{true}}$, where $V_{\text{true}}$ is approximated with $(U_A \otimes U_G)$ resulting in $\Lambda_{ii} = \mathbb{E}\left [ [(U_A \otimes U_G)^T \delta \theta]_{i}^2 \right ]$. We denote this as the eigenvalue corrected, Kronecker-factored eigenbasis (EFB):

\begin{equation}
    \boldsymbol{I}_{\text{efb}} = (U_A \otimes U_G)\Lambda (U_A \otimes U_G)^T
\end{equation}

This technique has many desirable properties. Notably, $\left \| \boldsymbol{I}-\boldsymbol{I}_{\text{efb}} \right \|_F \leq  \left \| \boldsymbol{I}-\boldsymbol{I}_{\text{kfac}} \right \|_F$ wrt. the Frobenius norm as the computation is more accurate by correcting the eigenvalues. 

However, there is an approximation in EFB since $ (U_A \otimes U_G)$ is still an approximation of the true eigenbasis $V_{\text{true}}$. Intuitively, EFB only performs a correction of the diagonal elements in the eigenbasis, but when mapping back to the parameter space this correction is again harmed by the inexact estimate of the eigenvectors. Although an exact estimation of the eigenvectors is infeasible, it is important to note that the diagonals of the exact IM $\boldsymbol{I}_{ii} = \mathbb{E} \left [ \delta \theta_i^2 \right ]$ can be computed efficiently using back-propagation. This motivates the idea to correct the approximation further as follows:

\begin{equation}
\label{eq:4:1}
\begin{aligned}
    \boldsymbol{I}_{\text{inf}} &= (U_A \otimes U_G)\Lambda (U_A \otimes U_G)^T + D \ \ \text{where} \\
   D_{ii} &= \mathbb{E} \left [ \delta \theta_i^2 \right ] - \sum_{j=1}^{nm} (v_{i, j} \sqrt{\Lambda_j})^2.
\end{aligned}
\end{equation}

In \eqref{eq:4:1}, we have represented $(U_A \otimes U_G)\Lambda (U_A \otimes U_G)_{ii}^T$ as $\sum_{j=1}^{nm} (v_{i, j} \sqrt{\Lambda_j})^2$  where $V = (U_A \otimes U_G) \in \mathbb{R}^{mn \times mn}$ is a Kronecker product with row elements $v_{i, j}$ (see definition 1 below). It follows from the properties of the Kronecker product that $i= m({\alpha}-1) + {\gamma}$. The derivation is shown in supplementary materials. Note that the Kronecker products are never directly evaluated but the diagonal matrix $D$ can be computed recursively, making it computationally feasible.

\textbf{Definition 1:} \textit{For $U_A \in \mathbb{R}^{n \times n}$ and $U_G \in \mathbb{R}^{m \times m}$, the Kronecker product of $V = U_A \otimes U_G \in \mathbb{R}^{mn \times mn}$ is given by $V_{i, j} = U_{A_{{\alpha, \beta}}}U_{G_{{\gamma, \zeta}}}$, with $i = m({\alpha}-1) + {\gamma}$ and $j = m({\beta}-1) + {\zeta}$. $\alpha \in \left \{1, \cdots, n  \right \}$ and $\beta \in \left \{1, \cdots, n  \right \}$ are row and column indices of $U_{A}$. So as $\gamma \in \left \{1, \cdots, m \right \}$ and $\zeta \in \left \{1, \cdots, m  \right \}$ for $U_{G}$.}

Now, the parameter posterior distribution can be represented in an information form $\mathcal{N}^{-1}$ of MND as shown below:

\begin{equation*}
\label{eq:4:2}
\begin{aligned}
    p(\theta \mid x,y) & \sim \mathcal{N}(\theta_{MAP},\boldsymbol{I}_{\text{inf}}^{-1}) \\
    &= \mathcal{N}^{-1}(\theta_{MAP}^{IV}, (U_A \otimes U_G)\Lambda (U_A \otimes U_G)^T + D).
\end{aligned}
\end{equation*}

This shows how an information form of MND can be computed using LA, and from its graphical interpretation, keeping the diagonals of IM exact has also a consequence of obtaining information content of the parameters accurate \citep{paskin2003thin}. We note that, similar insights have been studied for Bayesian tracking problems \citep{Thrun03d} with wide adoptions in practice \citep{bailey2006simultaneous,thrun2005multi,eustice2006exactly}.

For a full Bayesian analysis with DNNs, however, the samples of the resulting posterior are to be drawn from the information matrix instead of the covariance matrix. For this, an efficient sampling computation is proposed next.

\subsection{Model Uncertainty in Sparse Information Form}
\label{sec:lra}
Sampling from the posterior is crucial. For example, an important use-case of the parameter posterior is estimating the predictive uncertainty for test data ($x^*$,$y^*$) by a full Bayesian analysis with $K_{mc}$ samples. This step is typically approximated with Monte-carlo integration \citep{Gal2016Uncertainty}:


\begin{equation*}
\label{eq:marginalization}
    p(y^*|x^*, x, y) \approx \frac{1}{K_{mc}}\sum_{t=1}^{K_{mc}}y^*(x^*,\theta_{t}^s) \ \ \text{for } \theta_{t}^s \sim \mathcal{N}^{-1}(\theta_{MAP}^{IV}, \boldsymbol{I} _{\text{inf}}).
\end{equation*}

However, this operation is non-trivial as the sampling computation requires $O(N^3)$ complexity (the cost of inversion and finding a symmetrical factor) and for matrices that lie in a high dimensional space, it is computationally infeasible. While previous works \cite{MartensG15} showed that Kronecker products of two matrices can be exploited along a fidelity-cost trade-off, our main aim is to introduce an alternative form of the Gaussian posterior family. 

\begin{figure*}[tb]
\centering
\includegraphics[width=0.9\linewidth]{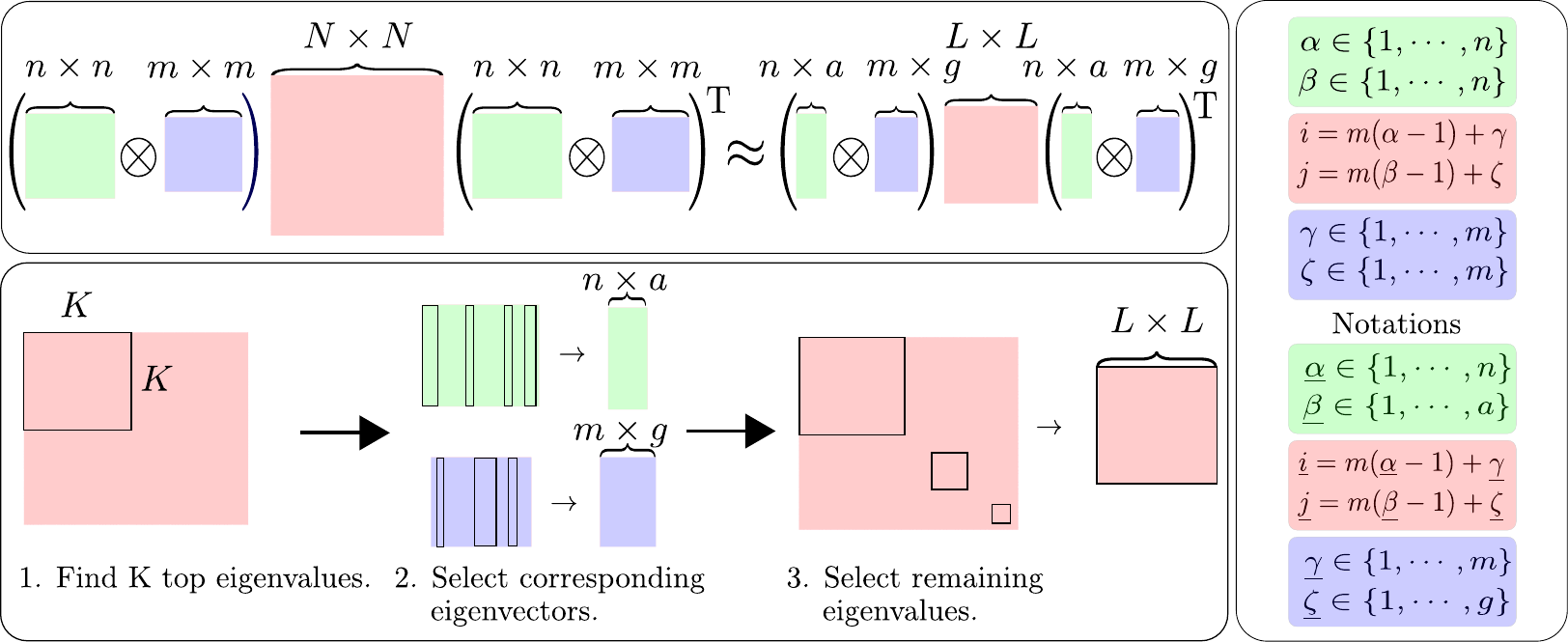}
\caption{\textbf{Illustration of algorithm \ref{algorithm1}.} A low rank approximation on Kronecker factored eigendecomposition that preserves Kronecker structure in eigenvectors have two benefits: (a) reducing directly $(U_A \otimes U_G)_{1:L}$ is memory-wise infeasible, and (b) sampling costs are drastically reduced as demonstrated in section \ref{sec:sampling:main}. Notations, low rank structure and a visualization of algorithm \ref{algorithm1} are depicted.}
\label{fig:lra:concept}
\end{figure*}

Observing that the eigenvalues of IM tends to be close to zero for the overparameterized DNNs \citep{SagunEGDB18}, information form can naturally leverage dimensionality reduction in IM (in oppose to the covariance matrix). Intuitively speaking, as more and more parameters are used to explain the same set of data, information of these parameters tends to be smaller, and we can make use of this tendency. 

To this end, we propose the low rank form in \eqref{eq:4:3} \footnote{D is added after LRA which is computed similar to \eqref{eq:4:1}.} as a first step, in which we preserve the Kronecker product in eigenvectors. Here, we highlight that the proposed form differs from conventional LRA which do not preserve the Kronecker product in eigenvectors (i.e $(U_{A} \otimes U_{G})_{1:L}$ for top L eigenvalues). Two main advantages of this representation are that it avoids the memory-wise expensive computation of evaluating the matrix $(U_{a} \otimes U_{g})$, where $U_a$ and $U_g$ are sub-matrices of $U_A$ and $U_G$, respectively. This formulation also results in sampling computation that is $O(L^3)$ in cost instead of $O(N^3)$, which we demonstrate later in section \ref{sec:sampling:main}.

\begin{equation}
\label{eq:4:3}
   \boldsymbol{I}_{\text{inf}} \approx \boldsymbol{\hat{I}}_{\text{inf}} = (U_{a} \otimes U_{g})\Lambda_{1:L} (U_{a} \otimes U_{g})^T + D
\end{equation}

Here, $\Lambda_{1:L}\in \mathbb{R}^{L \times L}$, $U_{a}\in \mathbb{R}^{m \times a}$ and $U_{g}\in \mathbb{R}^{n \times g}$ denote low rank form of corresponding eigenvalues and vectors (depicted in figure \ref{fig:lra:concept}). Naturally, it follows that $L = ag$, $N=mn$ and furthermore, the persevered rank $L$ corresponds to preserving top $K$ and additional $J$ eigenvalues (resulting in $L \geq K$, $L=ag=K+J$) as explained with an example.

\textbf{Why can't LRA directly be used?:} \textit{Let a matrix $E=U_{1:6}\Lambda_{1:6}U_{1:6}^T \in \mathbb{R}^{6 \times 6}$ with $U_{1:6}=[u_1 \cdots u_6] \in \mathbb{R}^{6 \times 6}$ with $\{u_i\}^{6}_{i=1}$ the eigenvectors and $\Lambda_{1:6} = \text{diag}(\lambda_1, \cdots, \lambda_6)\in \mathbb{R}^{6 \times 6}$ in a descending order. 
In this toy example, LRA with top 3 eigenvalues results: $E_{1:3} = U_{1:3}\Lambda_{1:3}U^{T}_{1:3}\in \mathbb{R}^{6 \times 6}$. 
Instead, if the eigenvector matrices are expressed in Kronecker product structure, $E_{\text{kron}} = (U_{A_{1:3}} \otimes U_{G_{1:2}})\Lambda_{1:6} (U_{A_{1:3}}\otimes U_{G_{1:2}})^T\in \mathbb{R}^{6 \times 6}$. 
For LRA, it's not trivial to directly preserve the top 3 eigenvalues $\Lambda_{1:3}$ and corresponding eigenvectors $(U_{A_{1:3}}\otimes U_{G_{1:2}})_{1:3}$. Because as $(U_{A_{1:a}}\otimes U_{G_{1:g}})_{1:3} = \begin{bmatrix} u_{A_1}\otimes u_{G_1} & u_{A_1}\otimes u_{G_2} & u_{A_2}\otimes u_{G_1}\end{bmatrix}$, preserving the eigenvectors with Kronecker structure results in having to store more eigenvectors: $U_{A_{1:2}}=[
u_{A_1} u_{A_2}
]$ and $U_{G_{1:2}}=[
u_{G_1} u_{G_2}
]$. Consequently, additional eigenvalue $\Lambda_{4}$ needs to be saved so that $E_{kron_{1:3}} = (U_{A_{1:2}} \otimes U_{G_{1:2}})\Lambda_{1:4}(U_{A_{1:2}} \otimes U_{G_{1:2}})^T \in \mathbb{R}^{6 \times 6}$.}

Then, how to achieve a LRA that preserves Kronecker structures in eigenvectors? For this, we propose algorithm \ref{algorithm1} (also illustrated in figure \ref{fig:lra:concept}). To select the additional eigenvectors and -values correctly, we need to introduce a definition on indexing rules of Kronecker factored diagonal matrices.

\textbf{Definition 2:} \textit{For diagonal matrices $S_A \in \mathbb{R}^{n \times n}$ and $S_G \in \mathbb{R}^{m \times m}$, the Kronecker product of $\Lambda = S_A \otimes S_G \in \mathbb{R}^{mn \times mn}$ is given by $\Lambda_{i} =  s_{\alpha \beta}s_{\gamma \zeta}$, where the indices $i = m(\beta-1) + \zeta$ with $\beta \in \left \{1, \cdots, m \right \}$ and $\zeta \in \left \{1, \cdots, n \right \}$. Then, given $i$ and $m$, $\beta = floor(\frac{i}{m})+1$ and given $\beta$, m, and $i$, $\zeta = i - m(\beta-1)$.}

\begin{algorithm}[tb]
   \caption{Spectral sparsification}
   \label{algorithm1}
\begin{algorithmic}
\STATE {\bfseries Input:} Matrices $U_{A}$, $U_{G}$, $\Lambda$ and Rank K. \\
\STATE {\bfseries Output:} Matrices: $U_{a}$, $U_{g}$, $\Lambda_{1:L}$.\\
\STATE 1. Find top K eigenvalues $\lambda_i$ on $\Lambda$ where $i \in \left \{ 1, \cdots, K \right \}$.
\STATE 2. For all $i$, find $\underline{\beta} = floor(\frac{i}{m})+1$ and $\underline{\zeta} = i - m(\underline{\beta}-1)$.\\ 
\STATE 3. Generate sub-matrices $U_a$ and $U_g$ by selecting \\ \hspace{0.3cm} eigenvectors in $U_A$ and $U_G$ according to $\underline{\beta}$ and $\underline{\zeta}$.  \\
\STATE 4. Find remaining eigenvalues $\lambda_j$ with $j = m({\underline{\beta}}-1)+{\underline{\zeta}}$.\\
\STATE 5. Concatenate and diagonalize selected eigenvalues\\ \hspace{0.3cm} $\Lambda_{1:L} = diag([\lambda_i, \lambda_j])$ for all $i$ and $j$.\\
\end{algorithmic}
\end{algorithm}

Notations in algorithm \ref{algorithm1} are also depicted in figure \ref{fig:lra:concept}. Now we explain this computation with a toy example below.

\textbf{Algorithm 1 for a toy example:} \textit{To explain, the same toy example can be revisited. Firstly, we aim to preserve the top 3 eigenvalues, $i \in \left \{ 1, 2, 3 \right \}$ which are indices of eigenvalues $\Lambda_{1:3}$ (step 1). Then, $\underline{\beta} \in \left \{ 1, 2 \right \}$ and $\underline{\zeta} \in \left \{ 1, 2 \right \}$ can be computed using definition 2 (step 2). This relation holds as $\Lambda$ is computed from $S_A \otimes S_G$, and thus, $U_A$ and $U_G$ are their corresponding eigenvectors respectively. Then we produce $U_{A_{1:2}}$ and $U_{G_{1:2}}$ according to $\underline{\beta}$ and $\underline{\zeta}$ (step 3). Again, in order to fulfill the Kronecker product operation, we need to find the eigenvalues $\underline{j} \in \left \{ 1, 2, 3, 4 \right \}$, and preserve $\Lambda_{1:4}$ (step 4$\&$5). This results in saving top 3 and additional 1 eigenvalues. Algorithm 1 provides the generalization of these steps and even if eigendecomposition does not come with a descending order, the same logic trivially applies.}

Next, we describe the last step of the sampling derivation.

\subsection{Low Rank Sampling Computations}
\label{sec:sampling:main}

Consider drawing samples $\theta_t^s$ $\in \mathbb{R}^{mn}$ from the representation:

\begin{equation}
    \theta_t^s \sim \mathcal{N}^{-1}(\theta_{\text{MAP}}^{IV}, (U_{a} \otimes U_{g})\Lambda_{1:L}(U_{a} \otimes U_{g})^T + D).
\end{equation}

Typically, drawing such samples $\theta_t^s$ requires finding a symmetrical factor of the covariance matrix (e.g. Chloesky decomposition) which is cubic in cost $O(N^3)$ (here $N = mn$). Furthermore, in our representation, it requires first an inversion of IM and then the computation of a symmetrical factor which overall constitutes two operations of cost $O(N^3)$. Clearly, if $N$ lies in a high dimension sampling becomes infeasible. Therefore, we need a sampling computation that performs these operations in the dimensions of low rank L.

Let us define $X^l \in \mathbb{R}^{mn}$ as the samples from a standard MND. Then, the samples $\theta_t^s$ can be computed analytically as, 

\begin{equation}
\label{eq:sampling:new1}
\begin{aligned}
    \theta_t^s &= \theta_{\text{MAP}} + F^cX^l \ \text{where}\\
    F^c & = D^{-\frac{1}{2}}\big(I_{nm} - D^{-\frac{1}{2}} (U_{a} \otimes U_{g})\Lambda_{1:L}^{\frac{1}{2}} \\
    & \underbrace{(C^{-1}+V_s^TV_s)^{-1}}_\text{cost: $O(L^3)\ll O(N^3)$}\Lambda_{1:L}^{\frac{1}{2}}(U_{a} \otimes U_{g})^TD^{-\frac{1}{2}} \big). 
\end{aligned}
\end{equation}

Firstly, the symmetrical factor $F^c \in \mathbb{R}^{mn \times mn}$ in \eqref{eq:sampling:new1} is a function of matrices that are feasible to store as they are diagonal or small Kronecker factored matrices. Furthermore,

\begin{equation*}
\begin{aligned}
    V_s & = D^{-\frac{1}{2}} (U_{a} \otimes U_{g})\Lambda_{1:L}^{\frac{1}{2}} \ \text{and} \ C = A_c^{-T}(B_c-I_{L})A_c^{-1}
\end{aligned}
\end{equation*}

with $A_c$ and $B_c$ being the Cholesky decomposed matrices of $V_s^TV_s \in \mathbb{R}^{L \times L}$ and $V_s^TV_s+I_L \in \mathbb{R}^{L \times L}$ such that:

\begin{equation*}
\begin{aligned}
    A_cA_c^T &= V_s^TV_s \ \text{and} \ B_cB_c^T = V_s^TV_s+I_L.
\end{aligned}
\end{equation*}

Consequently, the matrices in \eqref{eq:sampling:new1} are defined as $C \in \mathbb{R}^{L \times L}$, $(C^{-1}+V_s^TV_s) \in \mathbb{R}^{L \times L}$ and identity matrix $I_L \in \mathbb{R}^{L \times L}$. In this way, the two operations namely Cholesky decomposition and inversion that are cubic in cost $O(N^3)$ are reduced to the low rank dimension L with complexity $O(L^3)$. The complete derivation is in supplementary materials where we further show how the Kronecker structure in eigendecomposition can be exploited to compute $F^cX^l$ using the \textit{vec} trick.

\subsection{Theoretical Guarantees}
\label{sec:method:theory}
We outline theoretical guarantees of our approach below. Note that these properties hold regardless of data and model.

\textbf{Lemma 1: } \textit{Let $\boldsymbol{I}$ be the real information matrix, and let $\boldsymbol{I}_{\text{\text{inf}}}$ and $\boldsymbol{I}_{\text{efb}}$ be the INF and EFB estimates of it respectively. It is guaranteed to have $\left \| \boldsymbol{I} - \boldsymbol{I}_{\text{efb}} \right \|_F \geq \left \| \boldsymbol{I} - \boldsymbol{I}_{\text{\text{inf}}} \right \|_F$.}

\textbf{Corollary 1:} \textit{Let $\boldsymbol{I}_{\text{kfac}}$ and $\boldsymbol{I}_{\text{\text{inf}}}$ be KFAC and our estimates of real information matrix $\boldsymbol{I}$ respectively. Then, it is guaranteed to have $\left \| \boldsymbol{I} - \boldsymbol{I}_{\text{kfac}} \right \|_F \geq \left \| \boldsymbol{I} - \boldsymbol{I}_{\text{\text{inf}}} \right \|_F$.}

\textbf{Lemma 2: } \textit{Let $\boldsymbol{I}$ be the real information matrix, and let $\boldsymbol{\hat{I}}_{\text{\text{inf}}}$, $\boldsymbol{I}_{\text{efb}}$ and $\boldsymbol{I}_{\text{kfac}}$ be the low rank INF, EFB and KFAC estimates of it respectively. Then, it is guaranteed to have $\left \| \text{ diag}(\boldsymbol{I}) - \text{ diag}(\boldsymbol{I}_{\text{efb}}) \right \|_F \geq \left \| \text{ diag}(\boldsymbol{I}) - \text{ diag}(\boldsymbol{\hat{I}}_{\text{inf}}) \right \|_F = 0$ and $\left \| \text{ diag}(\boldsymbol{I}) - \text{diag}(\boldsymbol{I}_{\text{kfac}}) \right \|_F \geq \left \| \text{  diag}(\boldsymbol{I}) - \text{ diag}(\boldsymbol{\hat{I}}_{\text{inf}}) \right \|_F = 0$. Furthermore, if the eigenvalues of $\boldsymbol{\hat{I}}_{\text{\text{inf}}}$ contains all non-zero eigenvalues of $\boldsymbol{I}_{\text{\text{inf}}}$, it follows: $\left \| \boldsymbol{I} - \boldsymbol{I}_{\text{efb}} \right \|_F \geq \left \| \boldsymbol{I} - \boldsymbol{\hat{I}}_{\text{inf}} \right \|_F$.}

Proofs along with a further remarks and theoretical analysis on (i) error bounds of the proposed LRA and (ii) validity of the posterior can be found in supplementary materials.

\begin{figure}
\begin{center}
\includegraphics[width=0.9\linewidth]{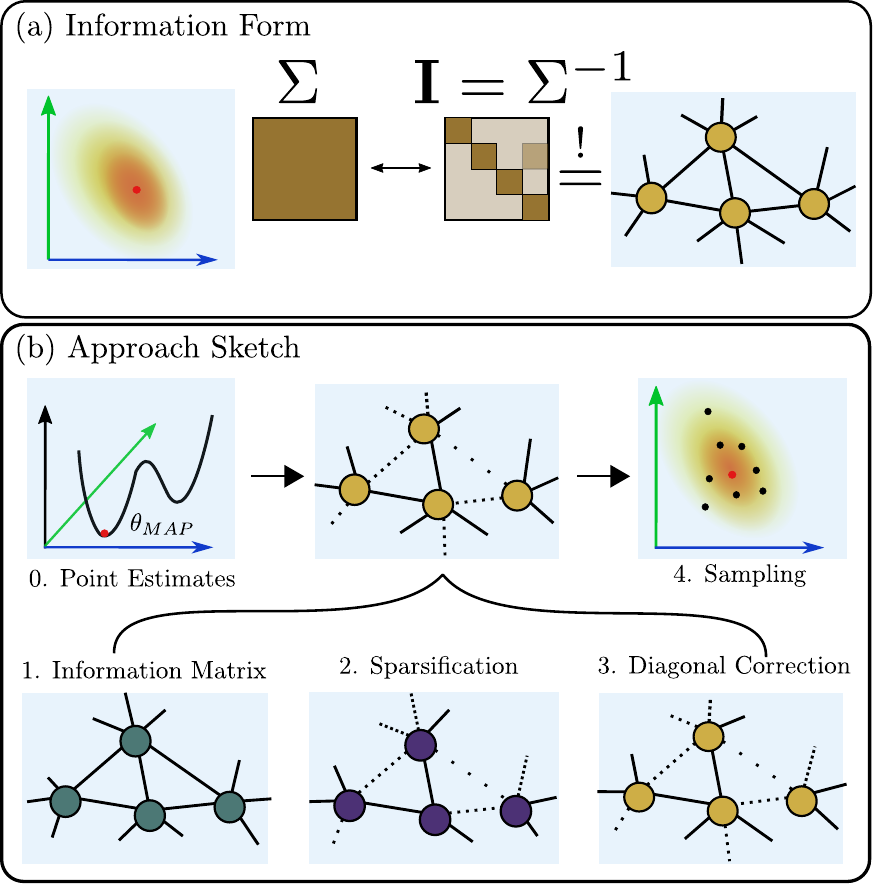}
		\caption{\textbf{Illustration of the main idea and the pipeline sketch}. We demonstrate that approximate Bayesian inference can work with the inverse covariance matrix - the information matrix. From its graphical interpretation \citep{paskin2003thin}, the diagonal elements represent the information content of the node while its off-diagonal elements represent the link between the nodes. Our approach is designed with this insight where we sparsify the weak links while keeping the information content of the node accurate.}
\label{fig:intro:concept}
\end{center}
\end{figure}

%% file: chapters/relatedworks.tex
\section{Related Works}
\textbf{Sparse Information Filters:} \citet{Thrun03d} proposed the extended sparse information filter (SEIF), which is a dual form of the extended Kalman filter. A key property of SEIF is that all update equations can be executed in constant time, which is achieved by relying on the information form and its sparsity. Our work brings the key ideas of SEIF in the context of approximate Bayesian inference for DNNs.

\textbf{Approximation of the Hessian:} The Hessian of DNNs is prohibitively too large as its size is quadratic to the number of parameters. For this problem, an efficient approximation is a layer-wise Kronecker factorization \citep{MartensG15, Botev2017} with demonstrably impressive scalability \citep{Ba2017}. In a recent extension by \citet{George2018} the eigenvalues of the Kronecker factored matrices are re-scaled so that the diagonal variance in its eigenbasis is exact. The work demonstrates a provable method of achieving improved performance. We heavily build upon these for Bayesian DNNs, as well-built software infrastructures such as \citep{dangel2020backpack} already exists. 

\textbf{Laplace Approximation:} Instead of methods rooted in variational inference \citep{Hinton:1993} and sampling \citep{Neal:1996:BLN:525544}, we utilize LA  \citep{MacKay1992} for the inference principle. Recently, diagonal \citep{Yann1989} and Kronecker-factored approximations \citep{Botev2017} to the Hessian have been applied to LA by \citet{Ritter2017ASL}. The authors have further proposed to use LA in continual learning \citep{RitterBB18}, and demonstrate competitive results by significantly outperforming its benchmarks \citep{Kirkpatrick2017OvercomingCF, Zenke2017}. Building upon \citet{Ritter2017ASL} for approximate inference, we propose to use more expressive posterior distribution than matrix normal distribution. Concurrently, \citet{kristiadi2020being} provides a formal statement on how approximate inference such as LA can mitigate overconfident behavior of DNNs with ReLU for a binary classification case.

In the context of variational inference, SLANG \citep{MishkinKNSK18} share similar spirit to ours in using a low-rank plus diagonal form of covariance where the authors show the benefits of low-rank approximation in detail. Yet, SLANG is different to ours as it does not explore Kronecker structures. SWA \citep{smith:2016:eclws}, SWAG \citep{swag} and subspace inference \citep{IzmailovMKGVW19} have also demonstrated strong results by exploring the insights on loss landscape of DNNs. We also acknowledge there exists alternatives paradigms. Some examples are the post-hoc calibrations \citep{guo17calib, wenger2020calibration}, ensembles \citep{lakshminarayanan2017simple} and combining Bayesian Neural Networks with probabilistic graphical models such as Conditional Random Fields \citep{dlr133562}.

\textbf{Dimensionality Reduction:} A vast literature is available for dimensionality reduction beyond principal component analysis \citep{wold1987principal} and singular value decomposition \citep{golub1971singular, van2009dimensionality}. To our knowledge though, dimensionality reduction in Kronecker factored eigendecomposition has not been studied.

%% file: chapters/results.tex
\section{Experimental Results}
\label{sec:result}
We perform an empirical study with regression, classification and active learning tasks. The chosen datasets are toy regression, UCI \citep{Dua_2019}, MNIST \citep{Lecun98gradient}, CIFAR10 \citep{cifar10} and ImageNet \citep{Krizhevsky2012} datasets. In total, 10 baseline with default 3 LA-based approaches (Diag, KFAC and EFB) are compared, and we also study the effects of varying LRA. 

Our main aim is to introduce the sparse information form of MND in the context of Bayesian Deep Learning, and thus, the experiments are designed to show the insight that spectral sparsification does not induce significant approximation errors while reducing the space complexity of using an expressive and structured posterior distribution. More importantly, we demonstrate that our method scales to datasets of ImageNet size and large architectures, and push the method further to compete with existing approaches. Implementation details can be found in supplementary materials.

%% file: chapters/results_toy_uci.tex
\subsection{Small Scale Experiments}

\begin{figure}[ht]
\minipage{0.24\textwidth}%
  \includegraphics[width=\linewidth]{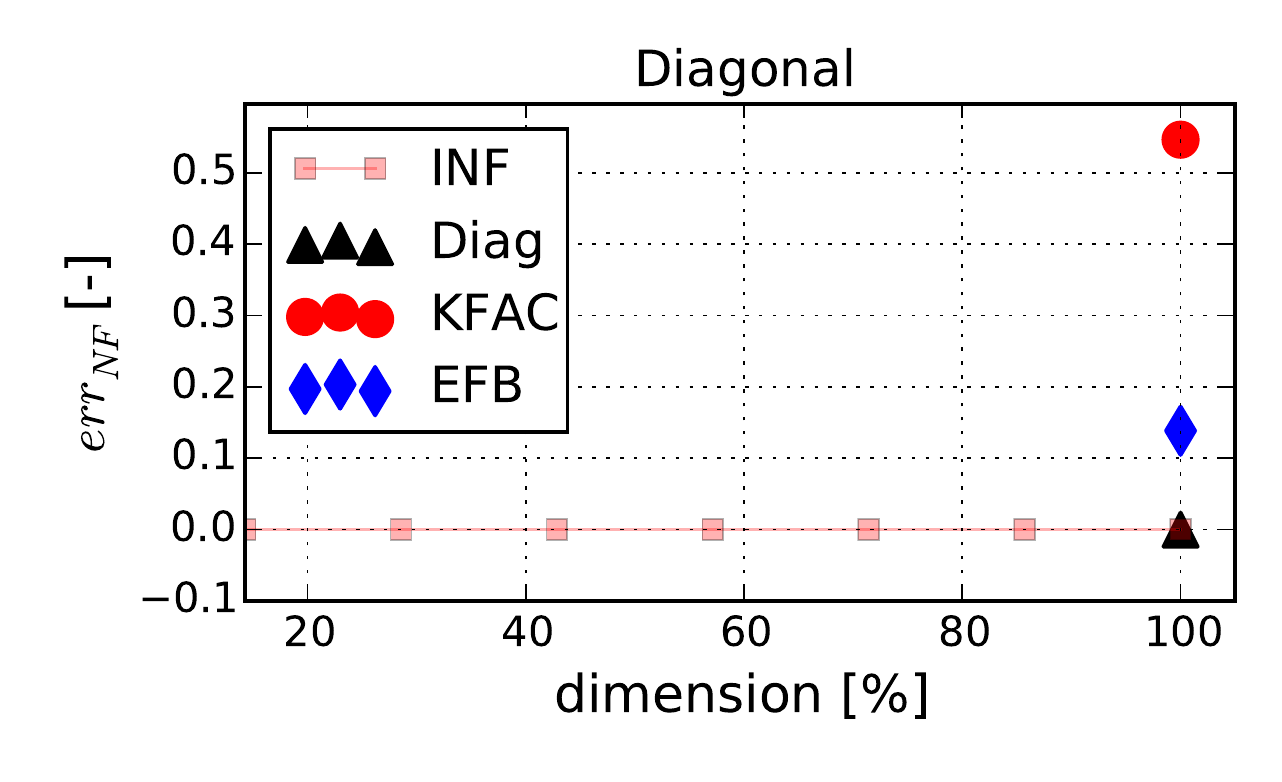}
\endminipage \hfill
\minipage{0.24\textwidth}%
  \includegraphics[width=\linewidth]{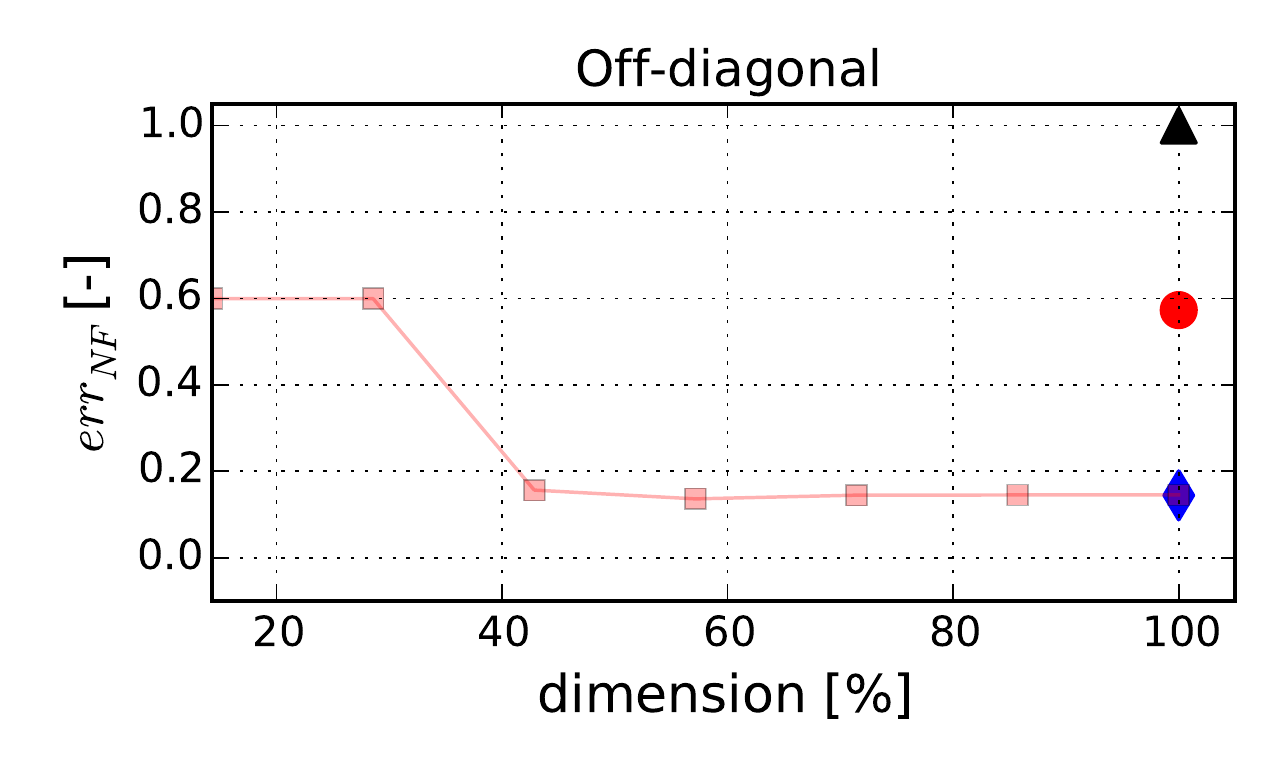}
\endminipage \hfill
\caption{\textbf{Effects of Low Rank Approximation in Frobenius norm of error.} This measure is normalized. Lower the better. EFB, Diag, KFAC and INF are compared in terms of diagonal and off-diagonal errors to exact block diagonal information matrix.}
\label{fig:diag}
\end{figure}

\begin{figure}[ht]
\centering
\includegraphics[width=0.85\linewidth]{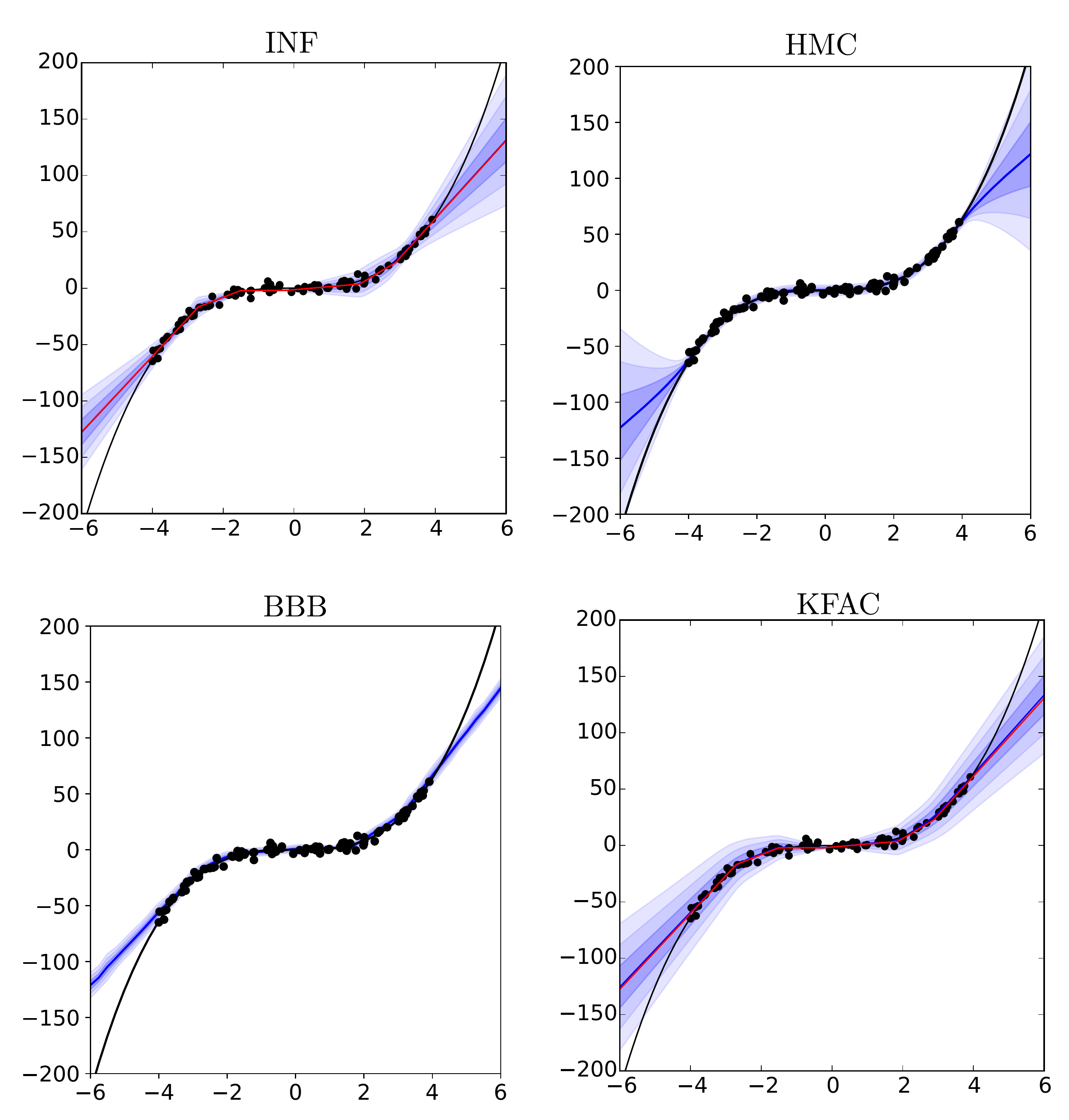}
\caption{\textbf{Predictive uncertainty.} The black dots and the black lines are data points (x, y). The red and blue lines show predictions of the deterministic network and the mean output respectively. Upto three standard deviations are shown with blue shades.}
\label{fig:1}
\end{figure}

Firstly, evaluations on toy regression and UCI datasets are presented. Due to the small scale of the set-up, these experiments have advantages that we can not only evaluate the quality of uncertainty estimation, but also directly compare various approximations to the Hessian with LRA. The later empirically validates our theoretic claims and further connects the qualities of uncertainty estimates and the approximates of the Hessian. For the toy regression problem, we consider a single-layered network with $7$ units. We have used 100 uniformly distributed points $x \sim U(-4, 4)$ and samples $y \sim N(x^3, 3^2)$. On UCI datasets we follow \citet{Hernandez2015} in which each dataset are split into 20 sets. A single layered network with $50$ units are used with an exception of protein, where we have used $100$ units.

\begin{table*}[ht]
\scriptsize
\centering
\caption{\textbf{Evaluating the accuracy of information matrix on UCI datasets.} The Frobenius norm of errors for diagonal and off-diagonal approximations w.r.t the exact block diagonal information matrix are depicted. Reported values are normalized. Here, INF (5$\%$) indicates that 95$\%$ of the ranks are thrown away, and Nr refers to the number of data points for each dataset. Lower the better.}
\label{results:tab:frobenius}
\begin{tabular}{cccccccccccc}
\midrule
Dataset &        & Diagonals    &    &  & &  Off-diagonals      &    &     &  \\
       & \textbf{Nr} & \textbf{KFAC}       & \textbf{EFB}     & \textbf{INF}   & \textbf{INF (5$\%$)} & \textbf{KFAC}       & \textbf{EFB}     & \textbf{INF}    & \textbf{INF (5$\%$)} \\
\midrule
\textit{Boston}  & 506 & 0.238$\pm$0.019      & 0.296$\pm$0.015     & \textbf{0.000$\pm$0.000}    & \textbf{0.000$\pm$0.000}  & 0.672$\pm$0.021 & \textbf{0.524$\pm$0.006} & \textbf{0.524$\pm$0.006} & \textit{0.608$\pm$0.009} \\
\textit{Concrete}  & 1030 & 0.185$\pm$0.020       & 0.253$\pm$0.008     & \textbf{0.000$\pm$0.000}  &  \textbf{0.000$\pm$0.000}  & 0.632$\pm$0.018 & \textbf{0.506$\pm$0.008} & \textbf{0.506$\pm$0.008} & 0.639$\pm$0.008 \\
\textit{Energy}  & 768 & 0.138$\pm$0.035 & 0.335$\pm$0.029 & \textbf{0.000$\pm$0.000}  &  \textbf{0.000$\pm$0.000}  & 0.646$\pm$0.012  & \textbf{0.504$\pm$0.006} & \textbf{0.504$\pm$0.006} & \textit{0.619$\pm$0.012} \\
\textit{Kin8nm} & 8192 & 0.077$\pm$0.008 & 0.256$\pm$0.020 & \textbf{0.000$\pm$0.000} &  \textbf{0.000$\pm$0.000}  & 0.594$\pm$0.005  & \textbf{0.526$\pm$0.005} & \textbf{0.526$\pm$0.005} & 0.670$\pm$0.003 \\
\textit{Naval}  & 11934 & 0.235$\pm$0.024 & 0.224$\pm$0.026 &\textbf{0.000$\pm$0.000}  & \textbf{0.000$\pm$0.000}  & 0.716$\pm$0.029  & \textbf{0.465$\pm$0.003} & \textbf{0.465$\pm$0.003} & \textit{0.480$\pm$0.003} \\
\textit{Power}  & 9568 & 0.113$\pm$0.012 & 0.252 $\pm$0.011 & \textbf{0.000$\pm$0.000}  &  \textbf{0.000$\pm$0.000}  & 0.681$\pm$0.006  & \textbf{0.492$\pm$0.008} & \textbf{0.492$\pm$0.008} & \textit{0.570$\pm$0.009} \\
\textit{Protein}  & 45730 & 0.323$\pm$0.067 & 0.332$\pm$0.043 & \textbf{0.000$\pm$0.000}  &  \textbf{0.000$\pm$0.000} & 0.779$\pm$0.040  & \textbf{0.541$\pm$0.021} & \textbf{0.541$\pm$0.021} & \textit{0.548$\pm$0.019} \\
\textit{Wine}  & 1599 & 0.221$\pm$0.021 & 0.287$\pm$0.022 & \textbf{0.000$\pm$0.000}  & \textbf{0.000$\pm$0.000} &  0.638$\pm$0.006  & \textbf{0.535$\pm$0.009} & \textbf{0.535$\pm$0.009} & 0.685$\pm$0.006 \\
\textit{Yacht}  & 308 & 0.104$\pm$0.007 & 0.201$\pm$0.019 & \textbf{0.000$\pm$0.000} & \textbf{0.000$\pm$0.000}  & 0.653$\pm$0.009  & \textbf{0.516$\pm$0.007} & \textbf{0.516$\pm$0.007} & 0.699$\pm$0.007 \\
\bottomrule
\end{tabular}
\end{table*}

We initially perform a direct evaluation of computed IM with a measure on normalized Frobenius norm of error $err_{NF}$ w.r.t the block-wise exact IM. Note that this is only possible for the small network architectures. The results are shown in figure \ref{fig:diag} and table \ref{results:tab:frobenius} for toy regression and UCI datasets respectively. Across all the experiments, we find that INF is exact on diagonal elements regardless of the ranks while KFAC and EFB induces significant errors. On off-diagonals, INF with full rank performs similar to EFB while decreasing the ranks of INF tend to increase the errors. Interestingly, INF with only $5\%$ of the ranks, often outperforms KFAC by a significant margin (e.g. Naval, Power, Protein). These observations are expected by the design of our approach and highlights the benefits of the information form. As the exact diagonals of IM are known and simple to compute (as oppose to the covariance matrix), we can design methods with theoretical guarantees on approximation quality of IM. Moreover, as the spectrum of IM tends to be sparse, LRA can be effectively exploited without inducing significant errors. For UCI experiments, we further study the varying effects of LRA in supplementary materials. 

Visualization of predictive uncertainty is shown in figure \ref{fig:1} for the toy regression. Here, HMC \citep{Neal:1996:BLN:525544} acts as ground truth while we compare our approach to KFAC and Bayes-by-backprop or BBB \citep{Blundell2015}. The hyperparameter sets for KFAC is chosen similar to \citep{Ritter2017ASL} while INF did not require the tuning of hyperparameters (after ensuring that IM is non-degenerate similar to \citet{MacKay1992}). All the methods show  higher uncertainty in the regimes far away from training data where BBB showing the most difference to HMC. Furthermore, KFAC predicts rather high uncertainty even within the regions that are covered by the training data. INF produces the most comparable fit to HMC with accurate uncertainty estimates. In supplementary materials, further comparison studies can be found. Furthermore, we also report the results of UCI experiments where we compare the reliability of uncertainty estimates using the test log-likelihood as a measure, and demonstrate competitive results as well as limitations.

%% file: chapters/results_active.tex
\subsection{Active Learning}
\begin{figure}[ht]
	\includegraphics[width=\linewidth, height=0.52\linewidth]{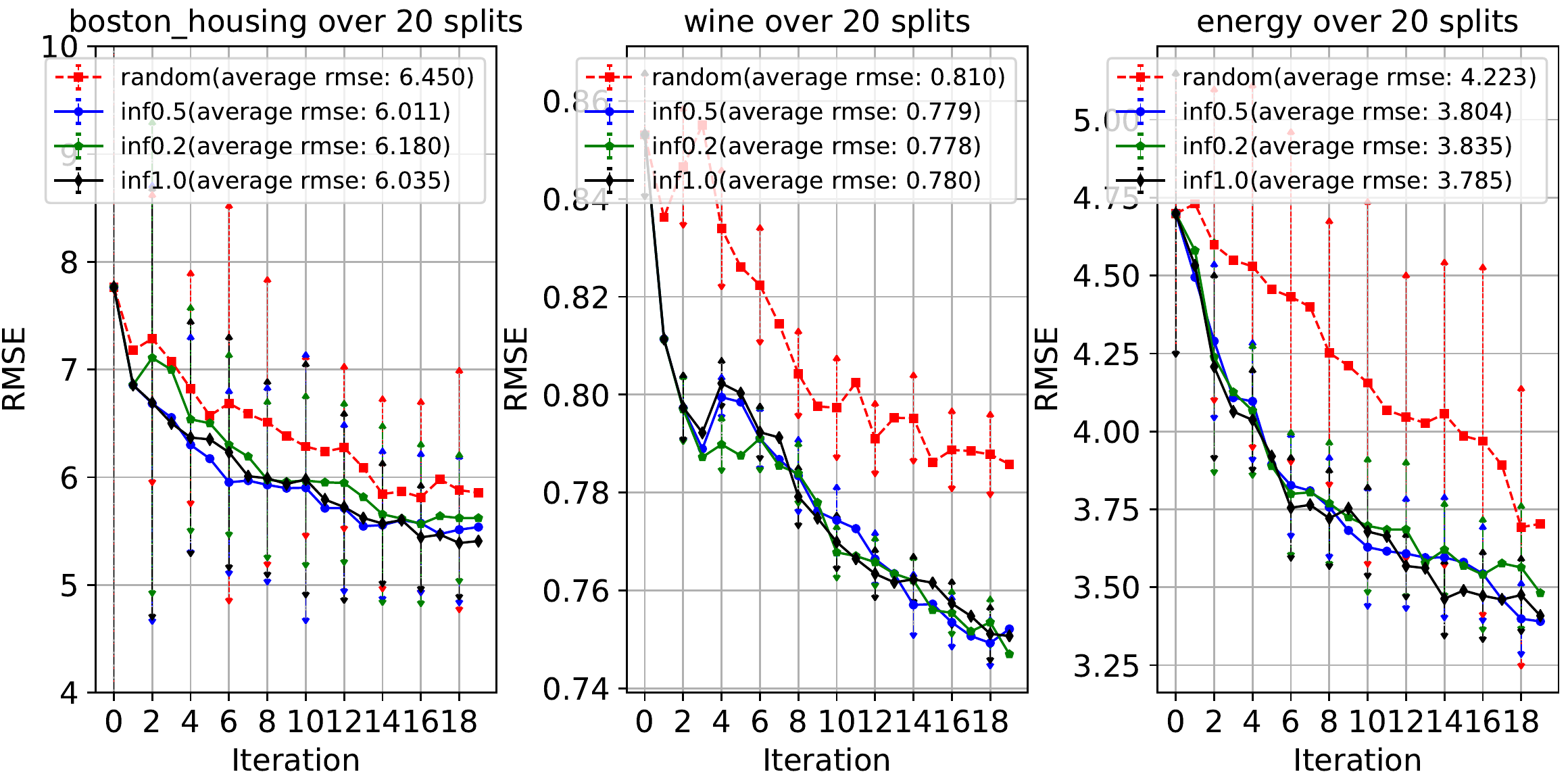}
	\caption{\textbf{Active learning with INF.} Average test RMSE over 20 splits and their standard errors in the active learning experiments with Boston Housing, Wine and Energy datasets. The average RMSE along 20 iterations of each curve is inside the bracket.}
	\label{fig:al_exper}
\end{figure}

We next show that our method can also perform a downstream task such as active learning. In this scenario, we further study the effects of LRA. For this purpose, we choose 3 UCI datasets (boston housing, wine and energy) closely following \citet{hernandez2015probabilistic}. In details the model is a neural network with a single layer of 10 hidden units. We employ the same criterion as \citet{mackay1992information} which linearizes the propagation of variance of weights to the output. Besides comparing with a baseline: random selection strategy, we also compare 3 different number of ranks (20\%, 50\%, 100\%) to verify the influence of LRA. 

As shown in figure \ref{fig:al_exper}, uncertainty estimates of INF enable the model to learn more quickly and lead to statistically significant improvements when compared to the random selection strategy. Further, even after cutting the ranks significantly, the performance can be maintained. This can be clearly seen on both Boston Housing and Energy dataset: the mean RMSE of lower percentage version decreased, while their standard deviation mostly overlap. On Wine, there is nearly no decrease in performance with lower ranks. To summarize, our experiments show that INF can perform active learning, and LRA does not jeopardize the task.

%% file: chapters/results_bdl.tex
\subsection{Classification Tasks}
\label{sec:results:class}

\begin{table*}[ht]
\scriptsize
\centering
\caption{\textbf{Results of classification experiments.} Accuracy and ECE are evaluated on in-domain distribution (MNIST and CIFAR10) whereas entropy is evaluated on out-of-distribution (notMNIST and SHVN). Lower the better for ECE. Higher the better for entropy.}
\label{results:tab:final}
\begin{tabular}{cccccccccccc|}
\toprule
Experiment & Measure & NN       & Diag     & KFAC    &  MC-dropout & Ensemble  & EFB & INF \\
\midrule
 & \textit{Accuracy}     &  0.993    & 0.9935    & 0.9929   &  0.9929  & \textbf{0.9937}        &  0.9929 &  0.9927 \\ 
MNIST vs notMNIST & \textit{ECE}    &  0.395    & 0.0075    & 0.0078   &  0.0105  & 0.0635                 &  0.012  &  \textbf{0.0069}  \\
& \textit{Entropy}     &  0.055$\pm$0.133    & 0.555 $\pm$ 0.196    & 0.599 $\pm$ 0.199 &   0.562 $\pm$ 0.19     &     0.596 $\pm$ 0.133      &  0.618 $\pm$ 0.185 &    \textbf{0.635 $\pm$ 0.19}        \\ 
\midrule
& \textit{Accuracy} &  0.8606 & \textbf{0.8659}   & 0.8572 & N/A & 0.8651  &  0.8638  &  0.8646  \\
CIFAR10 vs SHVN & \textit{ECE} &  0.0819   & 0.0358   & 0.0351  & N/A & 0.0809  &  0.0343   &  \textbf{0.0084}   \\
& \textit{Entropy}  &  0.245 $\pm$ 0.215    & 0.4129 $\pm$ 0.197  & 0.408 $\pm$ 0.197      & N/A & 0.370 $\pm$ 0.192      &  0.417 $\pm$ 0.196 &   \textbf{0.4338 $\pm$ 0.18} \\ 
\bottomrule
\end{tabular}
\end{table*}

Next, we evaluate predictive uncertainty on classification tasks where the proposed LRA is strictly necessary. To this end, we choose the classification tasks with known and unknown classes, e.g. a network is not only trained and evaluated on MNIST but also tested on notMNIST. Note that under such tests, any probabilistic methods should report their evaluations on both known and unknown classes with the same hyperparameter settings. This is because Bayesian Neural Networks can be always highly uncertain, which may seem to work well for out-of-distribution (OOD) detection tasks but overestimates the uncertainty, even for the correctly classified samples within the training data distribution. For evaluating predictive uncertainty on known classes, Expectation Calibration Error (ECE \citet{guo17calib}) has been used. Normalized entropy is reported for evaluating predictive uncertainty on unknown classes. LeNet with ReLU and a L2 coefficient of 1e-8 has been chosen for MNIST dataset, which constitutes of 2 convolution layers followed by 2 fully connected layers. The networks are intentionally trained to over-fit or over-confident, so that we can observe the effects of capturing model uncertainty. For CIFAR10, we choose VGG like architecture with 2 convolution layers followed by 3 fully connected layers. We used batch normalization instead of dropout layers.

The results are reported in table \ref{results:tab:final} where we also compared to MC-dropout \citep{Gal2016Uncertainty} and deep ensemble \citep{lakshminarayanan2017simple}, which are widely used baselines in practice. For CIFAR10, we omitted MC-dropout as additionally inserting dropout layers would result in a different network and thus, the direct comparisons would not be difficult. For LA-based methods, we have reported the best results after searching 300 hyperparameters each. We find this evaluation protocol to be crucial, as LA-based methods are sensitive to these regularizing hyperparameters. Importantly, these results demonstrate that when projected to different success criteria, no inference methods largely win uniformly. Yet these experiments also show empirical evidence that our method works in principle and compares well to the current state-of-the-art. Estimating the layer-wise parameter posterior distribution in a sparse information form of MND, and demonstrating a low rank sampling computations, we show an alternative approach of designing scalable, high performance and practical inference framework.

%% file: chapters/results_imagenet.tex
\subsection{Large Scale Experiments}

\begin{figure}[ht]
\centering
\includegraphics[width=\linewidth]{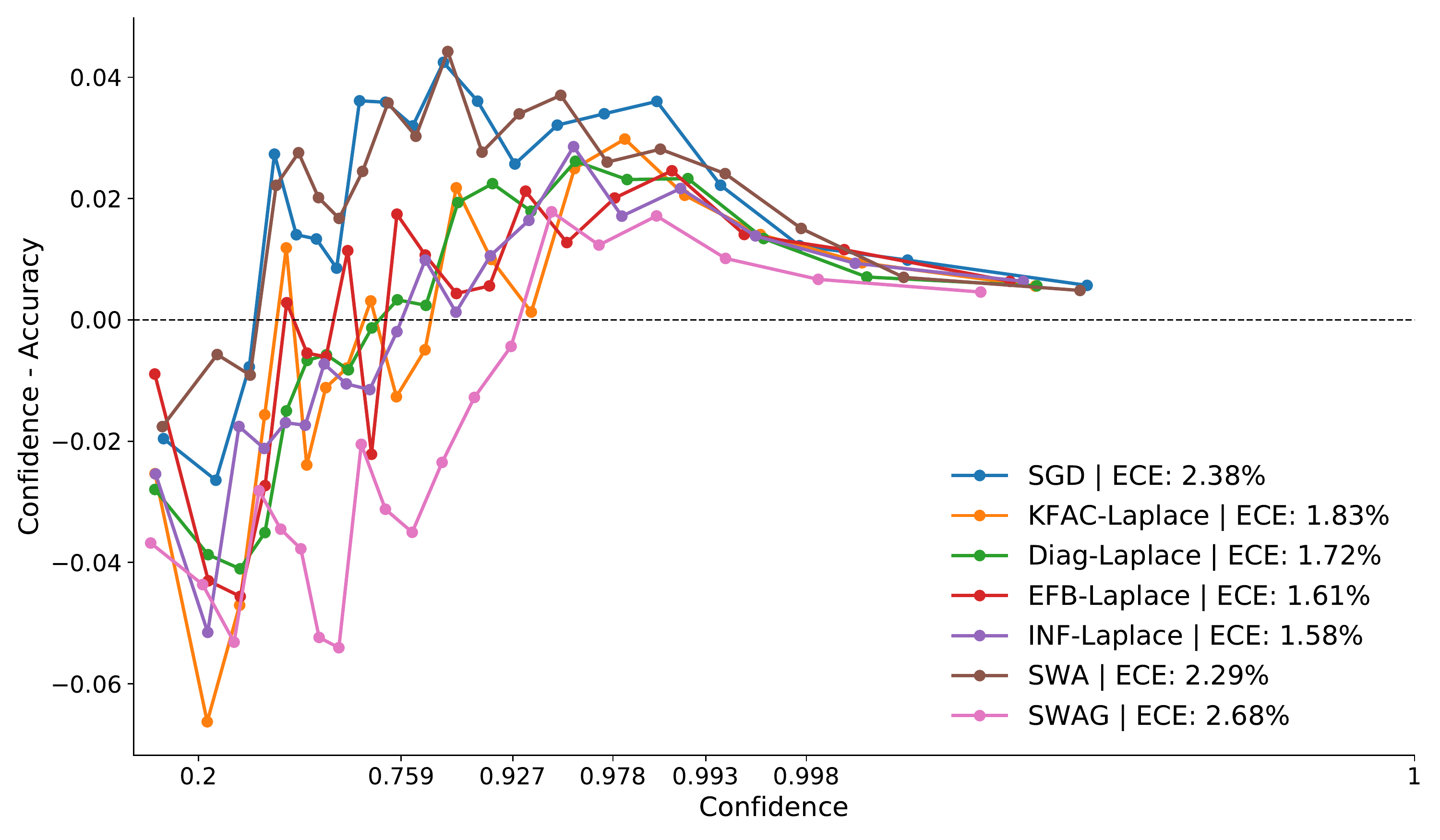}
\caption{\textbf{Reliability diagram of ResNet18 on ImageNet} Showing a calibration comparison of a deterministic forward pass (SGD) against Diag, KFAC, EKFAC, INF as well as SWA and SWAG.}
\label{fig:calib}
\end{figure}

\begin{table*}[ht]
\scriptsize
\centering
\caption{\textbf{Network Space Complexity Comparison:} The total number of information matrix parameters and its size in MB are reported for ResNet and DenseNet variants. Lower the better. Here, we also check if the methods take into account the weight correlations (corr).}
\label{results:sm:complexity:imagenet:all}
\begin{tabular}{lrrrrrrrrrrrr}
\toprule
		   & \multicolumn{3}{c}{\textbf{Diag}} & \multicolumn{3}{c}{\textbf{KFAC}} & \multicolumn{3}{c}{\textbf{EFB}} & \multicolumn{3}{c}{\textbf{INF}} \\
\midrule
 Model   &   \#Parameters &   Size & Corr &  \#Parameters &   Size & Corr &   \#Parameters &   Size & Corr &   \#Parameters &   Size & Corr \\
\midrule
 \textbf{ResNet18}  &   11,679,912 &        \textit{44.6} & $\text{\sffamily X}$ &   95,013,546 &       362.4 &   $\checkmark$ & 106,693,458 &       407.0 &   $\checkmark$ &  12,317,373 &        \textbf{47.0} &  $\checkmark$  \\
 \textbf{ResNet50}      &      25,503,912 &        \textit{97.3} &  $\text{\sffamily X}$ &   153,851,562 &       586.9 &  $\checkmark$ &   179,355,474 &       684.2 &  $\checkmark$ &    27,614,896 &       \textbf{105.3} &  $\checkmark$ \\
 \textbf{ResNet152}     &      60,041,384 &       \textit{229.0} &  $\text{\sffamily X}$ &   389,519,018 &      1485.9 &  $\checkmark$ &   449,560,402 &      1714.9 &  $\checkmark$ &    65,558,402 &       \textbf{250.1} &  $\checkmark$ \\
 \textbf{DenseNet121}   &       7,895,208 &        \textit{30.1} &  $\text{\sffamily X}$ &   103,094,954 &       393.3 &  $\checkmark$ &   110,990,162 &       423.4 & $\checkmark$ &     9,711,081 &        \textbf{37.0} &  $\checkmark$ \\
 \textbf{DenseNet161}   &      28,461,064 &       \textit{108.6} &  $\text{\sffamily X}$ &   379,105,514 &      1446.2 &  $\checkmark$ &   407,566,578 &      1554.7 &  $\checkmark$ &    32,329,191 &       \textbf{123.3} &  $\checkmark$ \\
\bottomrule
\end{tabular}
\end{table*}

\begin{figure*}[ht]
\minipage{0.25\textwidth}%
  \includegraphics[width=\linewidth]{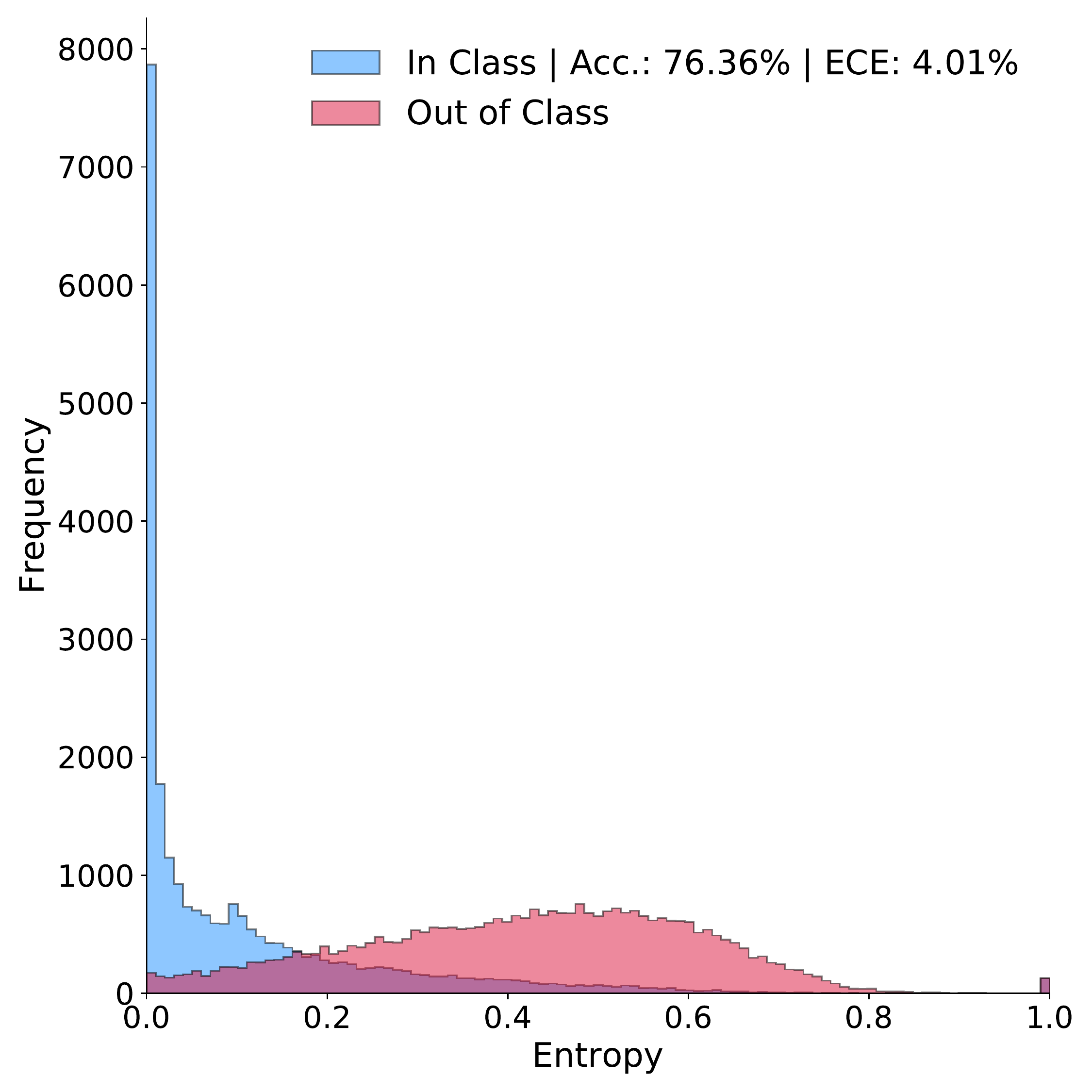}
\endminipage \hfill
\minipage{0.25\textwidth}
  \includegraphics[width=\linewidth]{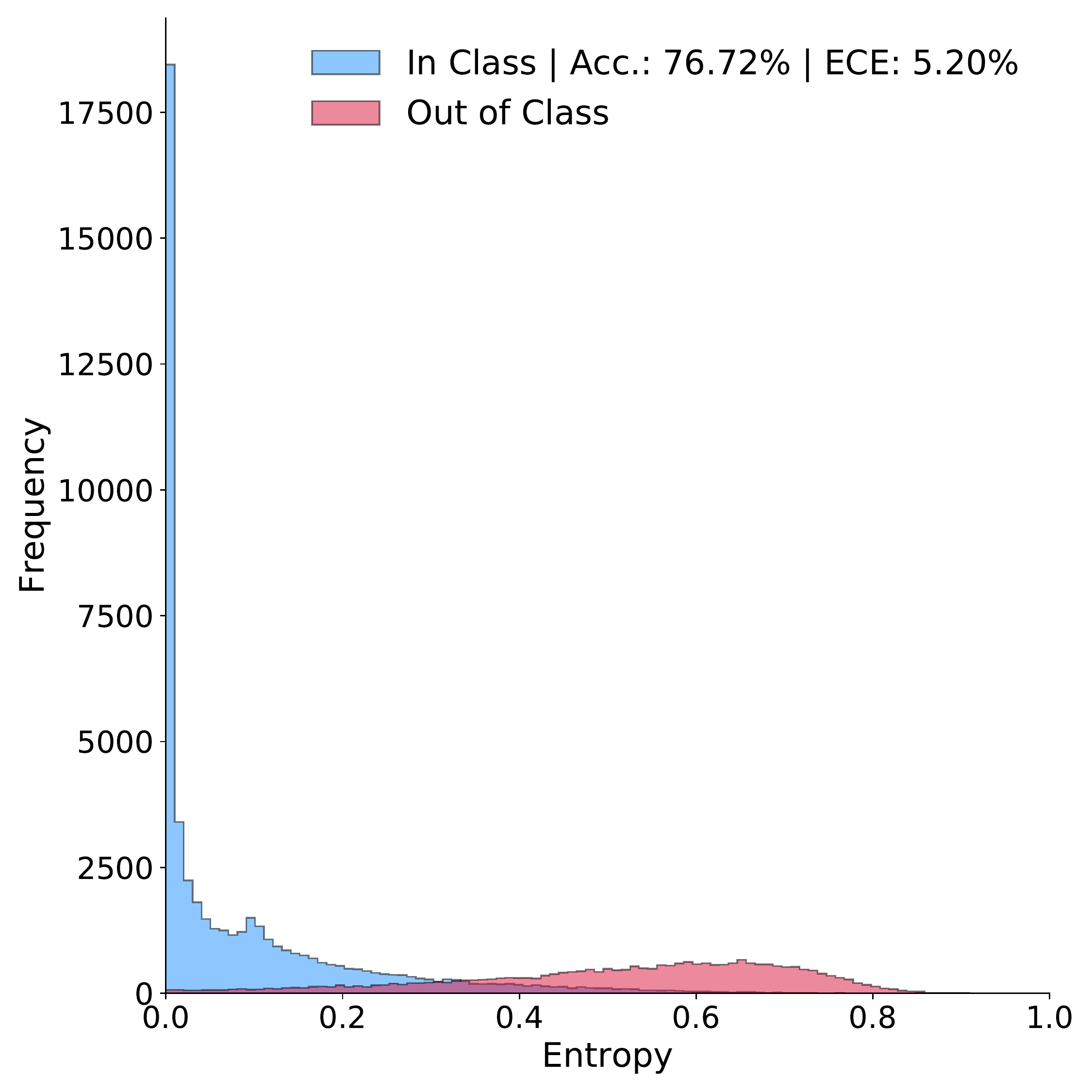}
\endminipage \hfill
\minipage{0.25\textwidth}%
  \includegraphics[width=\linewidth]{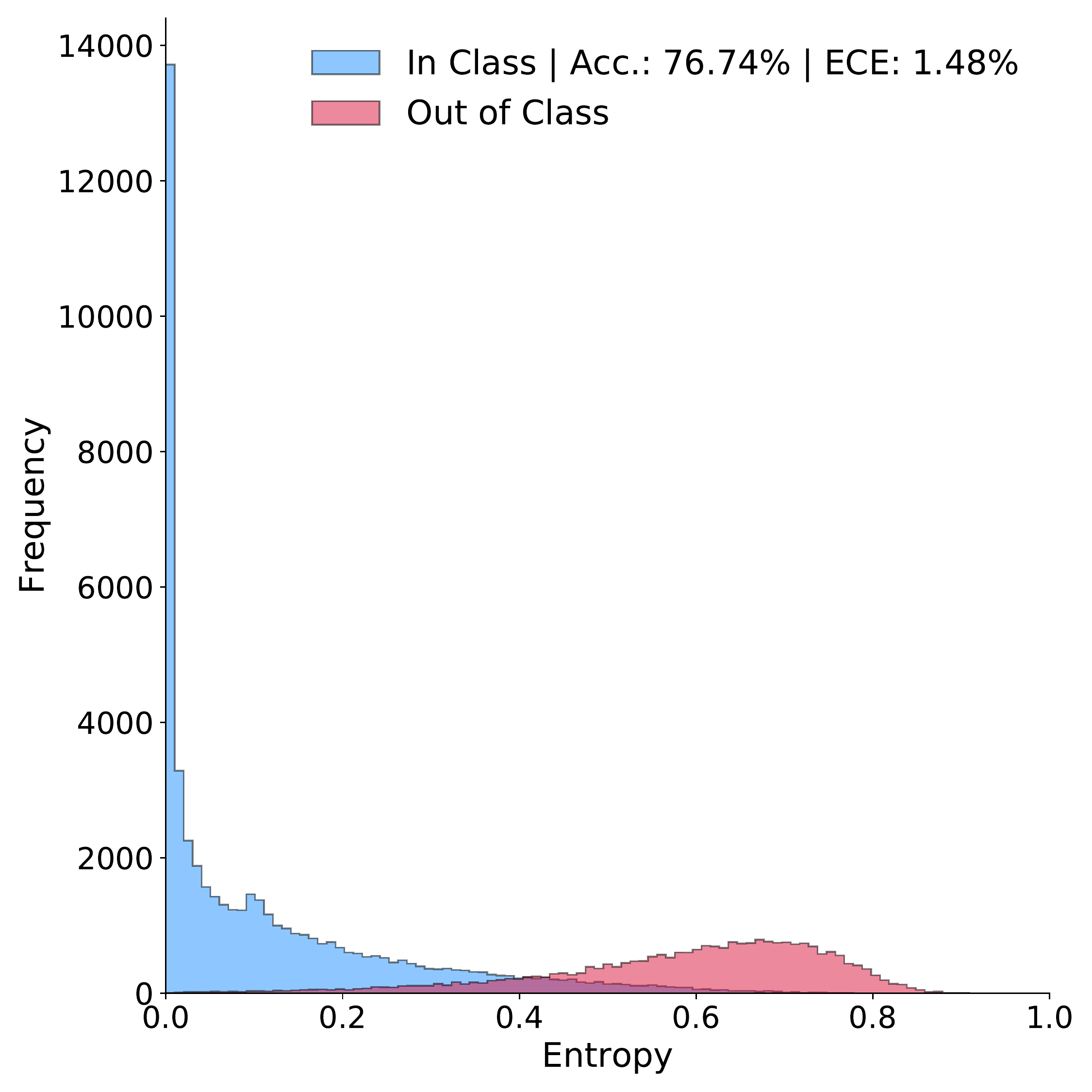}
\endminipage \hfill
\minipage{0.25\textwidth}
  \includegraphics[width=\linewidth]{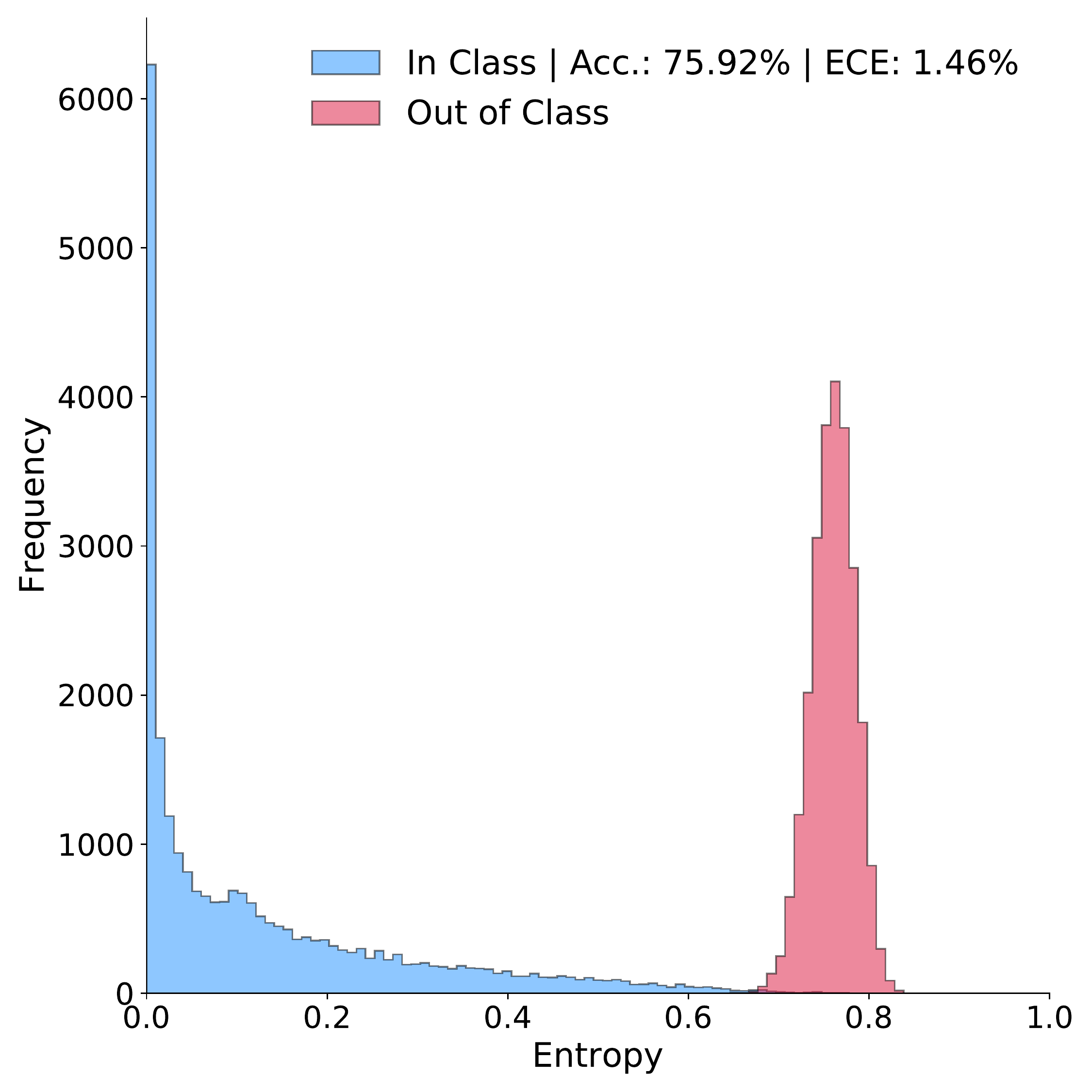}
\endminipage \hfill
\caption{\textbf{Entropy histograms ResNet50:} From left to right: SGD, SWA, SWAG and INF. While SGD, SWA and SWAG fail to separate in- and out-of-domain data, INF is able to almost completely differentiate between known and unknown data.}
\label{fig:entropy}
\end{figure*}

To show the scalability of our method, we conduct an extensive experimental evaluation on the ImageNet dataset, using 5 DNNs. This result alone is a key benefit of estimating model uncertainty of DNNs in sparse information form since approximate Bayesian inference only involves the computations of the information matrix in a training-free manner and the method boosts relaxed assumptions about the model when compared to MC-dropout (e.g. no specific regularizer is required). We also emphasize that the proposed diagonal correction requires only back-propagated gradients. EFB also uses the same gradients in their update step, and the whole chain takes around 8 hours on ImageNet with 1 NVIDIA Volta GPU. This means that one can store the exact diagonals of the Fisher during EFB computations and simply add a correction term without involving any data. Thus, the added computational overhead due to the diagonal correction is negligible in practice. Similar to section \ref{sec:results:class}, we evaluate the both calibration and OOD detection performances. SWAG and SWA are the chosen baselines, as \citet{swag} demonstrated that these methods scale to the ImageNet dataset. Moreover, to achieve comparability for large-scale settings, we scaled the other LA-based methods up so that they are applicable to ImageNet. This was not available in the original papers, and it constitutes an advance in evaluating LA-based methods for realistic scenarios. More importantly, we perform an extensive hyperparameter search over 100 randomly selected configurations for each LA-based methods. We believe that such protocols are required for fair evaluations of LA-based methods.
 
To directly compare the calibration across different methods, a variant of the reliability diagram \citep{swag} is used (figure \ref{fig:calib} for ResNet 18). We present results for all other architectures in supplementary materials. Here, we observe that all LA variants significantly outperform SGD, SWA and SWAG. Results of OOD detection tasks are reported in figure \ref{fig:entropy} where we also show the predictive uncertainty of the in-domain data for checking the under-confident behavior \citep{grimmett15introspective, mund15active, grimmett13knowing}. We use artistic impressions and paintings of landscapes and objects as OOD data. Again, SGD, SWAG, and SWA are in turn significantly outperformed by INF, which clearly separates in-distribution and OOD data. These results show the competitiveness of our approach for the real-world applications. However, we also find that all the LA-based methods almost identically yield strong results, clearly separating the in-distribution and OOD data.

While our method scales to ImageNet, we do not find that improvements in terms of Frobenius norm of error translates to performance in uncertainty estimation. This is the effects of regularizing hyperparameters ($\tau$ and N) which is a limitation of LA based approaches. We find a counter-intuitive result that Diag LA performs similar to ours and KFAC. It is therefore a strong alternative in practice where its complexity is superior than others (see table \ref{results:sm:complexity:imagenet:all}). Yet, ours, as we use rank $100$ in the experiments, are significantly superior in space complexity when compared to EFB and KFAC, while modeling the weight correlations. Therefore, we find the main use-case of INF: the applications such as aerial systems \citep{lee2018towards, JongSeok20} and medical devices \citep{petrou2018comparison, petrou2018standardized} which cannot afford much more memory due to the limited on-board computations, and requires structured form of model uncertainty, unlike Diag. Last but not least, the mathematical tools we develop and the idea of working on the inverse space of MND can also be useful in the context of variational inference as an example.

%% file: chapters/conclusion.tex
\section{Conclusion}
This work introduces the sparse information form as an alternative Gaussian posterior family for which, we presented novel mathematical tools such as a sparsification algorithm for the Kronecker factored eigendecomposition, and demonstrated how to efficiently sample from the resulting distribution. Our experiments show that our approach yields accurate estimates of the information matrix with theoretical guarantees, compares well to the current methods for the task of uncertainty estimation, scales to large scale data-sets while reducing space complexity, and can also be utilized for downstream tasks such as active learning. 

%% file: supplementary/organization.tex
\section{Organization of the document}
\label{sm:sec:1}
This document is organized as follows. Firstly, the mathematical derivations can be found in section \ref{sm:sec:2}. Then, we present the theoretical analysis (section \ref{sm:sec:3}) followed by their proofs (section \ref{sm:sec:4}). Implementation details and further results are presented in section \ref{sm:sec:5} and \ref{sm:sec:6} respectively. In particular, following additional results can be found.
\begin{itemize}
\item Empirical evidence on spectral sparsity of information matrix (section \ref{sm:sec:6:sparsity}).
\item Effects of low rank approximation on approximation quality of information matrix (section \ref{sm:sec:6:frolra}).
\item Effects of diagonal correction, data-set size, and critical review on KFAC on toy data (section \ref{sm:sec:toy}).
\item Effects of hyperparameters $N$ and $\tau$ (section \ref{sm:sec:6:hyp}).
\item Different architectures on ImageNet data-set and a time complexity analysis (section  \ref{sm:sec:ImageNet}).
\end{itemize}

%% file: supplementary/derivations.tex
\section{Mathematical Derivations}
\label{sm:sec:2}
\subsection{Problem Statement}
Let us assume that DNNs parameter posterior is estimated with MND layer-wise. Without diagonal approximation or Kronecker factorization of the covariance matrix (or similarly IM), the computational complexity of several operations namely storage, inversion and Cholesky decomposition becomes intractable. For example, if there exists a layer with 1 million parameters, the storage of covariance alone scales quadratic. If we use the simple formula for back-of-the envelope computations: total RAM for an $N \times N$ double precision matrix requires $N^2*\frac{8}{10^9}$ gigabytes, storing a covariance matrix for $N = 1000000$ requires 8000 gigabytes, which is computationally intractable for most of modern computers. Consequently, cubic in cost operations such as inversion or Cholesky decomposition is not feasible in a naive strategy. This is a reason why the current approaches use diagonal approximation or Kronecker factorization of the covariance matrix if MND is the chosen posterior family.

Following this statement, we list the memory-wise infeasible operations in detail: (i) naively extracting diagonal elements of $(U_A \otimes U_G)\Lambda (U_A \otimes U_G)^T$ and $(U_a \otimes U_g)\Lambda_{1:L} (U_a \otimes U_g)^T$, naively storing, inverting and Cholesky decomposing $(U_a \otimes U_g)\Lambda_{1:L} (U_a \otimes U_g)^T + D$ (iii) naively storing $(U_A \otimes U_G)$ and performing LRA on $(U_A \otimes U_G)\Lambda (U_A \otimes U_G)^T$. Our solution to the first two points are derived in this section while we have proposed an algorithm in the main manuscript to tackle the challenges of the last point. 

Table \ref{sm:sec2:problem_statement} shows the main result on space complexity, where the proposed sparse information form of DNNs posterior is also analyzed. Without resorting to mean-field approximations or matrix normal distribution, we show an alternative form of MND is also possible. Mathematical derivations we present below make this possible. Note that the inversion scheme considered is Gauss–Jordan elimination.

\begin{table}
\scriptsize
\centering
\caption{\textbf{Bounds on space complexity when compared to a naive strategy.} We compare the several operations with a naive strategy to our presented derivations, which is the main results of our work. Here, $L << N$ where L can be chosen with fidelity vs cost trade-off. This shows the sparse information form as a scalable Gaussian posterior family, providing alternatives to the diagonal covariance assumption or matrix normal distribution.}
\label{sm:sec2:problem_statement}
\begin{tabular}{lll}
\midrule
\textbf{Operations} & \textbf{Space Complexity} &  \\ 
          & \textit{Naive Strategy}   & \textit{Ours}   \\
\midrule 
Storage   &       $O(N^2)$           & $O(L + na + mg)$ \\
Inversion &       $O(N^3)$           &   $O(L^3)$    \\ 
Cholesky &   $O(N^3)$               &    $O(L^3)$   \\
Decomposition &                  &       \\ 
\bottomrule
\end{tabular}
\end{table}

\subsection{Diagonal Correction without Full Evaluation}
\label{sm:iter}
Directly evaluating $U_A \otimes U_G$ may not be computationally feasible for modern DNNs. Therefore, we derive the analytical form of the diagonal elements for $(U_A \otimes U_G) \Lambda (U_A \otimes U_G)^T$ without having to fully evaluate the Kronecker product. Let $U_A \in \mathbb{R}^{n \times n}$ and $U_G \in \mathbb{R}^{m \times m}$ be the square matrices. $\Lambda \in \mathbb{R}^{mn \times mn}$ is a diagonal matrix by construction. $V = U_A \otimes U_G \in \mathbb{R}^{mn \times mn}$ is a Kronecker product with elements $v_{i,j}$ with $i = m(\alpha-1) + \gamma$ and $j = m(\beta-1) + \zeta$ (from definition of Kronecker product). Then, the diagonal entries of $(U_A \otimes U_G)\Lambda (U_A \otimes U_G)^T$ can be computed as follows:
\begin{equation}
\label{sm:sec3:eq:1}
    \left [ (U_A \otimes U_G)\Lambda (U_A \otimes U_G)^T \right ]_{ii} = \sum_{j=1}^{nm} (v_{i, j} \sqrt{\Lambda_j})^2
\end{equation}

\textbf{Derivation:} As a first step of the derivation, we express $(A \otimes B)\Lambda (A \otimes B)^T$ in the following form:
\begin{equation*}
\begin{aligned}
   (U_A \otimes U_G)\Lambda (U_A \otimes U_G)^T &= (U_A \otimes U_G)\Lambda^{\frac{1}{2}} \Lambda^{\frac{1}{2}}(U_A \otimes U_G)^T \\
    &= \left [(U_A \otimes U_G)\Lambda^{\frac{1}{2}}\right ] \left [(U_A \otimes U_G)\Lambda^{\frac{1}{2}}\right ]^T \\
    &= UU^T
\end{aligned}
\end{equation*}

Then, $\text{diag}(UU^T)_i = \left [ UU^T \right ]_{ii} = \sum_{j=1}^{nm}u_{ij}^2$ by definition. Now, we let $(U_A \otimes U_G)\Lambda^{\frac{1}{2}} = V\Lambda^{\frac{1}{2}}$ with $\Lambda^{\frac{1}{2}}$ being again a diagonal matrix. Therefore, $u_{ij} = v_{i, j} \sqrt{\Lambda_j}$ due to the multiplication with a diagonal matrix from a right hand side. Substituting back these results in $\left [ (U_A \otimes U_G)\Lambda (U_A \otimes U_G)^T \right ]_{ii} = \sum_{j=1}^{nm} (v_{i, j} \sqrt{\Lambda_j})^2$ which completes the derivation. Formulating equation \ref{sm:sec3:eq:1} for the non-square matrices (which results after a low rank approximation) such as $U_a \in \mathbb{R}^{n \times a}$ and $U_g \in \mathbb{R}^{m \times g}$ and paralleling this operation are rather trivial and hence, we omit this part of the derivation.

\subsection{Low Rank Sampler - Analytical Form}
\label{section:sampler}
For a full Bayesian analysis which is approximated by a Monte Carlo integration, sampling is a crucial operation for predicting uncertainty. We start by stating the problem.

\textbf{Problem statement:} Consider drawing samples $\theta_t^s =$ vec($W_t^{s}$) $\in \mathbb{R}^{nm}$ from the proposed sparse information form:
\begin{equation}
\label{eq:sampling:3}
    \theta_t^s \sim \mathcal{N}^{-1}(\theta_{\text{MAP}}^{IV}, (U_a \otimes U_g)\Lambda_{1:L}(U_a \otimes U_g)^T + D)
\end{equation}

Drawing such samples from a covariance form of MND requires finding a symmetrical factor of the covariance matrix (e.g. Chloesky decomposition) which is cubic in cost $O(N^3)$. Even worse, when represented in an information form as in \eqref{eq:sampling:3}, it requires first an inversion of information matrix and then the computation of a symmetrical factor which overall constitutes two operations of cost $O(N^3)$. Clearly, if $N$ lies in a high dimension such as 1 million, even storing is obvious not feasible, let alone the sampling computations. Therefore, we need a sampling computation that (a) keeps the Kronecker structure while sampling so that first, the storage is memory-wise feasible, and then (b) the operations that require cubic cost such as inversion, must be performed in the dimensions of low rank L instead of full parameter dimensions N. We provide the solution below.

\textbf{Analytical solution:} Let us define $X^l \in \mathbb{R}^{mn}$ and $X^s \in \mathbb{R}^{m \times n}$ as the samples from a standard Multivariate Normal Distribution in \eqref{eq:sampling:5} where we denote the followings: $0_{nm} \in \mathbb{R}^{nm}$, $I_{mn} \in \mathbb{R}^{mn \times mn}$, $0_{n \times m} \in \mathbb{R}^{n \times m}$, $I_{n} \in \mathbb{R}^{n \times n}$ and $I_{m} \in \mathbb{R}^{m \times m}$. Note that these sampling operations are cheap.

\begin{equation}
\label{eq:sampling:5}
    X^l \sim N(0_{nm}, I_{nm}) \ \text{ or } \ X^s \sim \mathcal{MN}(0_{n \times m}, I_{n}, I_{m}).
\end{equation}

Furthermore, we denote $\theta_t^s = vec(W_t^s) \in \mathbb{R}^{mn}$, $\theta_{\text{MAP}} = vec(W_{\text{MAP}}) \in \mathbb{R}^{mn}$ as a sample from equation \ref{eq:sampling:3} and its mean as a vector respectively. We also note that $\Lambda_{1:L} \in \mathbb{R}^{L \times L}$ and $D \in \mathbb{R}^{mn \times mn}$ are the low ranked form of the re-scaled eigen-values and the diagonal correction term as previously defined. $U_a \in \mathbb{R}^{n \times a}$ and $U_g \in \mathbb{R}^{m \times g}$ are the eigenvectors of low ranked eigen-basis so that $n \geq a$, $m \geq g$ and $L=ag$. Then, the samples of \ref{eq:sampling:3} can be computed analytically as\footnote{We show how the Kronecker structure of $F^c$ can be exploited to compute $F^cX^l$ in the derivation only.}:

\begin{equation}
\label{eq:sampling:sm:new}
\begin{aligned}
    \theta_t^s &= \theta_{\text{MAP}} + F^cX^l \ \text{where, }\\
    F^c & = D^{-\frac{1}{2}}\bigg(I_{nm} - D^{-\frac{1}{2}} (U_a \otimes U_g)\Lambda_{1:L}^{\frac{1}{2}} \\
    & (C^{-1}+V_s^TV_s)^{-1}\Lambda_{1:L}^{\frac{1}{2}}(U_a \otimes U_g)^TD^{-\frac{1}{2}} \bigg). 
\end{aligned}
\end{equation}

Firstly, the symmetrical factor $F^c \in \mathbb{R}^{mn \times mn}$ in \eqref{eq:sampling:sm:new} is a function of matrices that are feasible to store as they involve diagonal matrices or small matrices in a Kronecker structure. Furthermore,

\begin{equation}
\begin{aligned}
    V_s & = D^{-\frac{1}{2}} (U_a \otimes U_g)\Lambda_{1:L}^{\frac{1}{2}} \\
    C & = A_c^{-T}(B_c-I_{L})A_c^{-1} \ \text{with} \ A_c \ \text{and} \ B_c
\end{aligned}
\end{equation}

being the Cholesky decomposed matrices of $V_s^TV_s \in \mathbb{R}^{L \times L}$ and $V_s^TV_s+I_L \in \mathbb{R}^{L \times L}$ such that:

\begin{equation}
\begin{aligned}
    A_cA_c^T &= V_s^TV_s \ \text{and} \\
    B_cB_c^T &= V_s^TV_s+I_L.
\end{aligned}
\end{equation}

Consequently, the matrices in \eqref{eq:sampling:sm:new} are defined as $C \in \mathbb{R}^{L \times L}$, $(C^{-1}+V_s^TV_s) \in \mathbb{R}^{L \times L}$ and $I_L \in \mathbb{R}^{L \times L}$. In this way, the two operations namely Cholesky decomposition and inversion that are cubic in cost $O(N^3)$ are reduced to the low rank dimension L with complexity $O(L^3)$.

\textbf{Derivation:} Firstly, note that sampling from a standard multivariate Gaussian for $X^l$ or $X^s$ is computationally cheap (see equation \ref{eq:sampling:5}). Given a symmetrical factor for the covariance $\Sigma = F^cF^{c^T}$ (e.g. by Cholesky decomposition), samples can be drawn via $\theta_{\text{MAP}} + F^cX^l$ as depicted in \eqref{eq:sampling:sm:new}. Our derivation involves finding such symmetrical factor for the given form of covariance matrix while exploring the Kronecker structure for the sampling computations so that the space complexity is bounded to $O(L^3)$ instead of $O(N^3)$.

Let us first reformulate the covariance (inverse of information matrix) as follows.

\begin{equation}
\begin{aligned}
   \Sigma &= \bigg((U_a \otimes U_g)\Lambda_{1:L}(U_a \otimes U_g)^T + D \bigg)^{-1} \\
    &= \bigg [ D^{\frac{1}{2}}\big (D^{-\frac{1}{2}} (U_a \otimes U_g)\Lambda_{1:L}^{\frac{1}{2}} \\
    & \Lambda_{1:L}^{\frac{1}{2}}(U_a \otimes U_g)^T D^{-\frac{1}{2}} + I_{nm} \big )D^{\frac{1}{2}} \bigg ]^{-1}\\
    &=D^{-\frac{1}{2}}\bigg [ \big ((D^{-\frac{1}{2}} (U_a \otimes U_g)\Lambda_{1:L}^{\frac{1}{2}} ) \\
    & (D^{-\frac{1}{2}} (U_a \otimes U_g)\Lambda_{1:L}^{\frac{1}{2}})^T + I_{nm}  \big ) \bigg ]^{-1}D^{-\frac{1}{2}}\\
    &= D^{-\frac{1}{2}}\left [ V_sV_s^T + I_{nm} \right ]^{-1}D^{-\frac{1}{2}}.
\end{aligned}
\end{equation}

Here, we define: $V_s=D^{-\frac{1}{2}} (U_a \otimes U_g)\Lambda_{1:L}^{\frac{1}{2}}$. Now, a symmetrical factor for $\Sigma=F^cF^{c^T}$ can be found by exploiting the above structure. We let $W^c$ be a symmetrical factor for $V_sV_s^T + I_{nm}$ so that $F^c=D^{-\frac{1}{2}}W^{c^{-1}}$ is the symmetrical factor of $\Sigma$. Following the work of \citet{Ambikasaran2014FastSF} the symmetrical factor $W^c$ can be found using equations:

\begin{equation}
\begin{aligned}
    W^c &= I_{nm} + V_sCV_s^T \\
    C &= A_c^{-T}(B_c-I_{L})A_c^{-1}.
\end{aligned}
\end{equation}

Note that A and B are Cholesky decomposed matrices of $V_s^TV_s \in \mathbb{R}^{L \times L}$ and $V_s^TV_s+I_L \in \mathbb{R}^{L \times L}$ respectively. As a first result, this operation is bounded by complexity $O(L^3)$ instead of the full parameter dimension $N$. Calculations of $V_s^TV_s$ can also be performed in iterations similar to derivations shown in section \ref{sm:iter}. Now the symmetrical factor for $\Sigma$ can be expressed as follows.

\begin{equation}
\begin{aligned}
    F^c &= D^{-\frac{1}{2}}W^{-1} = D^{-\frac{1}{2}}(I_{nm} + V_sCV_s^T)^{-1}  \\
      &= D^{-\frac{1}{2}}\bigg(I_{nm} - V_s(C^{-1}+V_s^TV_s)^{-1}V_s^T \bigg).
\end{aligned}
\end{equation}

Woodbury's Identity is used here. Now, by substitution:

\begin{equation}
\begin{aligned}
    \theta_t^s &= \theta_{\text{MAP}} + F^cX^l \ \text{where,} \\
    F^c & = D^{-\frac{1}{2}}\bigg(I_{nm} - V_s(C^{-1}+V_s^TV_s)^{-1}V_s^T \bigg) \\
     & = D^{-\frac{1}{2}}\bigg(I_{nm} - D^{-\frac{1}{2}} (U_a \otimes U_g)\Lambda_{1:L}^{\frac{1}{2}} \\
     & (C^{-1}+V_s^TV_s)^{-1}\Lambda_{1:L}^{\frac{1}{2}}(U_a \otimes U_g)^TD^{-\frac{1}{2}} \bigg). 
\end{aligned}
\end{equation}

This completes the derivation of \eqref{eq:sampling:sm:new}. As a result, the inversion operation is bounded by complexity $O(L^3)$. Furthermore, the derivation constitutes smaller matrices $U_a$ and $U_g$ or diagonal matrices $D$ and $I_{mn}$ which can be stored as vectors. In short the complexity has significantly reduced.

Now we further derive computations that exploits rules of Kronecker products. Consider:

\begin{equation}
\begin{aligned}
    F^cX^l &= D^{-\frac{1}{2}}\bigg(I_{nm} - D^{-\frac{1}{2}} (U_a \otimes U_g)\Lambda_{1:L}^{\frac{1}{2}} \\
    & (C^{-1}+V_s^TV_s)^{-1}\Lambda_{1:L}^{\frac{1}{2}}(U_a \otimes U_g)^TD^{-\frac{1}{2}} \bigg)X^l.
\end{aligned}
\end{equation}

Then, it follows by defining inverted matrix $L^c = (C^{-1}+V_s^TV_s)^{-1} \in \mathbb{R}^{L \times L}$ with a cost $O(L^3)$:

\begin{equation}
\begin{aligned}
    F^cX^l &= D^{-\frac{1}{2}}\bigg(I_{nm} - D^{-\frac{1}{2}} (U_a \otimes U_g)\Lambda_{1:L}^{\frac{1}{2}}L^c \\
    & \Lambda_{1:L}^{\frac{1}{2}}(U_a \otimes U_g)^TD^{-\frac{1}{2}} \bigg)X^l.
\end{aligned}
\end{equation}

We further reduce this by evaluating $D^{-\frac{1}{2}}$ and defining $X_D^l = D^{-\frac{1}{2}}X^l \in \mathbb{R}^{mn}$ and $P^c = \Lambda_{1:L}^{\frac{1}{2}}L^c\Lambda_{1:L}^{\frac{1}{2}} \in \mathbb{R}^{L \times L}$. We note that this multiplication operation is memory-wise feasible.

\begin{equation}
    F^cX^l = X_D^l - \bigg(D^{-1} (U_a \otimes U_g)P^c(U_a \otimes U_g)^TX_D^l \bigg).
\end{equation}

Now, we map $X_D^l$ to matrix normal distribution by an unvec($\cdot$) operation so that $X_D^s =$ unvec($X_D^l$) $\in \mathbb{R}^{m \times n}$ or equivalently $X_D^l =$ vec($X_D^s$). Using a widely known relation for Kronecker product that is - $(U_a \otimes U_g)^T$ vec($X_D^s$) $=$ vec($U_g^T X_D^s U_a$), it follows:

\begin{equation}
\begin{aligned}
    F^cX^l  &= X_D^l - \bigg(D^{-1} (U_a \otimes U_g)P^c \text{vec}(U_g^T X_D^s U_a )\bigg).
\end{aligned}
\end{equation}

Note that matrix multiplication is performed with small matrices. Repeating a similar procedure as above we obtain the equation below for $X_P^s = P^c$ $(U_a \otimes U_g)^TX_D^l$,

\begin{equation}
\begin{aligned}
    F^cX^l  &= X_D^l - \bigg(D^{-1} (U_a \otimes U_g)X_P^s \bigg) \\
    &= X_D^l - \bigg(D^{-1} \text{vec}(U_g X_P^s U_a^T) \bigg).
\end{aligned}
\end{equation}

This completes the derivation. Lastly, we provide a remark below to summarize the main points.

\textbf{Remark:} We presented a new derivation to sample from \eqref{eq:sampling:3}, a low-rank and information formulation of MND. This analytical solution ensures (a) $O(N^3) >> O(L^3)$ for Cholesky decomposition, (b) $O(N^3) >> O(L^3)$ for a matrix inversion, (c) storage of small matrices $U_g$, $U_a$, a diagonal matrix $D$ and identity matrices and finally (d) matrix multiplications that only involve these matrices. This is a direct benefit of our proposed LRA that preserves Kronecker structure in eigenvectors. Furthermore, this result shows the sparse information form as a new scalable Gaussian posterior family for approximate Bayesian inference.

%% file: supplementary/proofs.tex
\section{Theoretical Analysis and Results}
\label{sm:sec:3}
Some of the interesting theoretical properties are as follows with proofs provided in section \ref{sm:sec:4}.

\subsection{More Accurate Information Matrix}
Theoretical results of adding a diagonal correction term to Kronecker factored eigenbasis are captured below.

\textbf{Lemma 1: } \textit{Let $\boldsymbol{I}$ be the real information matrix, and let $\boldsymbol{I}_{\text{\text{inf}}}$ and $\boldsymbol{I}_{\text{efb}}$ be the INF and EFB estimates of it respectively. It is guaranteed to have $\left \| \boldsymbol{I} - \boldsymbol{I}_{\text{efb}} \right \|_F \geq \left \| \boldsymbol{I} - \boldsymbol{I}_{\text{\text{inf}}} \right \|_F$.}

\textbf{Corollary 1:} \textit{Let $\boldsymbol{I}_{\text{kfac}}$ and $\boldsymbol{I}_{\text{\text{inf}}}$ be KFAC and our estimates of real information matrix $\boldsymbol{I}$ respectively. Then, it is guaranteed to have $\left \| \boldsymbol{I} - \boldsymbol{I}_{\text{kfac}} \right \|_F \geq \left \| \boldsymbol{I} - \boldsymbol{I}_{\text{\text{inf}}} \right \|_F$.}

For interested readers, find the proof $\left \| \boldsymbol{I} - \boldsymbol{I}_{kfac} \right \|_F \geq \left \| \boldsymbol{I} - \boldsymbol{I}_{\text{efb}} \right \|_F$ in \citet{George2018}. Note that $\left \| \boldsymbol{I} - \boldsymbol{I}_{kfac} \right \|_F \geq \left \| \boldsymbol{I} - \boldsymbol{I}_{\text{efb}} \right \|_F$ may not mean that $\left \| \boldsymbol{I}^{-1} - \boldsymbol{I}_{kfac}^{-1} \right \|_F \geq \left \| \boldsymbol{I}^{-1} - \boldsymbol{I}_{\text{efb}}^{-1} \right \|_F$ or vice versa. Yet, our proposed approximation yields better estimates than KFAC in the information form of MND.

\subsection{Properties of Low-Rank Information Matrix}

To our knowledge, the proposed sparse IM have not been studied before. Therefore, we theoretically motivate its design and validity for better insights.

\textbf{Lemma 2: } \textit{Let $\boldsymbol{I}$ be the real Fisher information matrix, and let $\boldsymbol{\hat{I}}_{\text{\text{inf}}}$, $\boldsymbol{I}_{\text{efb}}$ and $\boldsymbol{I}_{\text{kfac}}$ be the low rank INF, EFB and KFAC estimates of it respectively. Then, it is guaranteed to have $\left \| \text{ diag}(\boldsymbol{I}) - \text{ diag}(\boldsymbol{I}_{\text{efb}}) \right \|_F \geq \left \| \text{ diag}(\boldsymbol{I}) - \text{ diag}(\boldsymbol{\hat{I}}_{\text{inf}}) \right \|_F = 0$ and $\left \| \text{ diag}(\boldsymbol{I}) - \text{diag}(\boldsymbol{I}_{\text{kfac}}) \right \|_F \geq \left \| \text{  diag}(\boldsymbol{I}) - \text{ diag}(\boldsymbol{\hat{I}}_{\text{inf}}) \right \|_F = 0$. Furthermore, if the eigenvalues of $\boldsymbol{\hat{I}}_{\text{\text{inf}}}$ contains all non-zero eigenvalues of $\boldsymbol{I}_{\text{\text{inf}}}$, it follows: $\left \| \boldsymbol{I} - \boldsymbol{I}_{\text{efb}} \right \|_F \geq \left \| \boldsymbol{I} - \boldsymbol{\hat{I}}_{\text{inf}} \right \|_F$.}

Lemma 2 shows the optimally in capturing the diagonal variance while indicating that our approach also becomes effective in estimating off-diagonal entries if IM contains many close to zero eigenvalues. Validity of this assumption has been studied by \citet{SagunEGDB18} where it is shown that the Hessian of overparameterized DNNs tend to have many close-to-zero eigenvalues. Intuitively, from a graphical interpretation of IM, diagonal entries indicate information present in each nodes and off-diagonal entries are links of these nodes. Our sparsification scheme reduces the strength of the weak links while keeping the diagonal variance exact. This is a result of the diagonal correction after LRA which exploits spectrum sparsity of IM.

\textbf{Lemma 3: } \textit{The low rank matrix $\hat{\Sigma} = \left ( (U_a \otimes U_g)\Lambda_{1:L} (U_a \otimes U_g)^T + D \right )^{-1} \in \mathbb{R}^{N \times N}$ is a non-degenerate covariance matrix if the diagonal correction matrix D and LRA $(U_a \otimes U_g)\Lambda_{1:L} (U_a \otimes U_g)^T$ are both symmetric and positive definite. This condition is satisfied if $(U_a \otimes U_g)\Lambda_{1:L} (U_a \otimes U_g)_{ii}^T < \mathbb{E} \left [ \delta \theta_i^2 \right ]$ for all i $\in \left \{ 1, 2, \cdots , d \right \}$ and with $\Lambda_{1:L} \nsubseteq {0}$.}

This comments on validity of the resulting posterior (a sufficient condition only) and proves that sparsifying the matrix can lead to a valid non-degenerate covariance if two conditions are met. As non-degenerate covariance can have a uniquely defined inverse, it is important to check these two conditions. We note that searching the rank can be automated with off-line computations that does not involve any data. Thus, it does not introduce significant overhead. In case D does not turn out to be, there are still several techniques that can deal with it. We recommend eigen-value clipping \citep{Chen2018BDAPCHBA} or finding nearest positive semi-definite matrices \citep{HIGHAM1988103}. For a side note, above Lemma provides a sufficient condition and even if D is not positive definite, there is no indication that the given representation is an invalid form of covariance. These conditions have been a conservative guideline to make the likelihood term non-degenerate which we found to work well in practice. Lastly, $D^{-1}$ is more numerically stable when we add a prior precision term and a scaling factor $(ND+\tau I)^{-1}$. 

Before introducing the next theoretical property we define,

\begin{equation}
    \boldsymbol{\hat{I}}_{1:K}^{\text{top}} = (U_A \otimes U_G)_{1:K}\Lambda_{1:K} (U_A \otimes U_G)_{1:K}^T
\end{equation}

as a low rank EFB estimates of true Fisher that preserves top K eigenvalues. Similarly, $\boldsymbol{\hat{I}}_{1:L}^{\text{top}}$ can be defined which preserves top L eigenvalues. In contract, our proposal to preserve Kronecker structure in eigenvectors $\boldsymbol{\hat{I}}_{1:L}$ is denoted as shown below. Now, we start our analysis with Lemma 2.

\begin{equation}
    \boldsymbol{\hat{I}}_{1:L} = (U_a \otimes U_g)\Lambda_{1:L} (U_a \otimes U_g)^T.
\end{equation}

\textbf{Lemma 4: } \textit{Let $\boldsymbol{I} \in \mathbb{R}^{N \times N}$ be the real Fisher information matrix, and let $\boldsymbol{\hat{I}}_{1:K}^{\text{top}} \in \mathbb{R}^{N \times N}$, $\boldsymbol{\hat{I}}_{1:L}^{\text{top}} \in \mathbb{R}^{N \times N}$ and $\boldsymbol{\hat{I}}_{1:L} \in \mathbb{R}^{N \times N}$ be the low rank estimates of $\boldsymbol{I}$ of EFB obtained by preserving top K, L and top K plus additional J resulting in L eigenvalues. Here, we define $K < L$. Then, the approximation error of $\boldsymbol{\hat{I}}_{1:L}$ is as follows: $\left \| \boldsymbol{I} - \boldsymbol{\hat{I}}_{1:L}^{\text{top}} \right \|_F \geq \left \| \boldsymbol{I} -\boldsymbol{\hat{I}}_{1:L} \right \|_F \geq \left \| \boldsymbol{I} - \boldsymbol{\hat{I}}_{1:K}^{\text{top}} \right \|_F$.}

This bound provides an insight that if preserving top L eigenvalues result in prohibitively too large covariance matrix, our LRA provides an alternative to preserving top K eigenvalues given that K $<$ L. In practice, note that $\boldsymbol{\hat{I}}_{1:L}$ is a memory-wise feasible option as we formulate $\boldsymbol{\hat{I}}_{1:L} = (U_a \otimes U_g)\Lambda_{1:L} (U_a \otimes U_g)^T$ which preserves the Kronecker structure in eigenvectors. This can be a case where evaluating $(U_a \otimes U_g)$ or $(U_a \otimes U_g)_{1:K}$ is not feasible to store.

\section{Proofs}
\label{sm:sec:4}

\subsection{More Accurate of Information Matrix}
\textbf{Proposition 1:} \textit{Let $\boldsymbol{I} \in R^{N \times N}$ be the real information matrix, and let $\boldsymbol{I}_{\text{inf}} \in R^{N \times N}$ and $\hat{\boldsymbol{I}}_{\text{inf}} \in R^{N \times N}$ be our estimates of it with rank d and k such that $k < d$. Their diagonal entries are equal that is $\boldsymbol{I}_{ii}$ = $\boldsymbol{I}_{\text{inf}_{ii}}$ = $\hat{\boldsymbol{I}}_{\text{inf}_{ii}}$ for all i = 1, $\dots$, N.}

\textit{proof:} 
The proof trivially follows from the definitions of $\boldsymbol{I} \in R^{N \times N}$, $\boldsymbol{I}_{\text{inf}} \in R^{N \times N}$ and $\hat{\boldsymbol{I}}_{\text{inf}} \in R^{N \times N}$. As the exact Fisher is an expectation on outer products of back-propagated gradients, its diagonal entries equal $\boldsymbol{I}_{ii} = \mathbb{E} \left [ \delta \theta_i^2 \right ]$ for all i = 1, 2, $\dots$, N.

In the case of full ranked $\boldsymbol{I}_{inf}$, substituting $D_{ii} = \mathbb{E} \left [ \delta \theta_i^2 \right ] - \sum_{j=1}^{nm} (v_{i, j} \sqrt{\Lambda_j})^2$ with $\sum_{j=1}^{nm} (v_{i, j} \sqrt{\Lambda_j})^2 = (U_A \otimes U_G)\Lambda (U_A \otimes U_G)_{ii}^T$ results in equation \ref{sm:eq:4:1} for all i = 1, 2, $\dots$, N. 

\begin{equation}
\label{sm:eq:4:1}
\begin{aligned}
    \boldsymbol{I}_{inf_{ii}} &= (U_A \otimes U_G)\Lambda (U_A \otimes U_G)_{ii}^T + D_{ii} \\
    &= (U_A \otimes U_G)\Lambda (U_A \otimes U_G)_{ii}^T + \mathbb{E} \left [ \delta \theta_i^2 \right ] \\
    & - (U_A \otimes U_G)\Lambda (U_A \otimes U_G)_{ii}^T \\
    &= \mathbb{E} \left [ \delta \theta_i^2 \right ]
\end{aligned}
\end{equation}

Similarly, we substitute $\hat{D}_{ii} = \mathbb{E} \left [ \delta \theta_i^2 \right ] - \sum_{j=1}^{L} (\hat{v}_{i, j} \sqrt{\Lambda_{1:L}})^2$ with $\sum_{j=1}^{L} (\hat{v}_{i, j} \sqrt{\Lambda_{1:L}})^2 = (U_a \otimes U_g)\Lambda_{1:L} (U_a \otimes U_g)_{ii}^T$ which results in equation \ref{sm:eq:4:2} for all i = 1, 2, $\dots$, N.

\begin{equation}
\label{sm:eq:4:2}
\begin{aligned}
    \hat{\boldsymbol{I}}_{inf_{ii}} &= (U_a \otimes U_g)\Lambda_{1:L} (U_a \otimes U_g)_{ii}^T + D_{ii} \\
    &= (U_a \otimes U_g)\Lambda_{1:L} (U_a \otimes U_g)_{ii}^T + \mathbb{E} \left [ \delta \theta_i^2 \right ] \\
    & - (U_a \otimes U_g)\Lambda_{1:L} (U_a \otimes U_g)_{ii}^T \\
    &= \mathbb{E} \left [ \delta \theta_i^2 \right ]
\end{aligned}
\end{equation}

Therefore, we have $\boldsymbol{I}_{ii}$ = $\boldsymbol{I}_{\text{inf}_{ii}}$ = $\hat{\boldsymbol{I}}_{\text{inf}_{ii}}$ for all i = 1, 2, $\dots$, N.

\textbf{Lemma 1: } \textit{Let $\boldsymbol{I}$ be the real information matrix, and let $\boldsymbol{I}_{\text{\text{inf}}}$ and $\boldsymbol{I}_{\text{efb}}$ be the INF and EFB estimates of it respectively. It is guaranteed to have $\left \| \boldsymbol{I} - \boldsymbol{I}_{\text{efb}} \right \|_F \geq \left \| \boldsymbol{I} - \boldsymbol{I}_{\text{\text{inf}}} \right \|_F$.}

\textit{proof:} 
Let $e^2=\left \| A - B \right \|^2_F$ define a squared Frobenius norm of error between the two matrices $A \in \mathbb{R}^{N \times N}$  and $B \in \mathbb{R}^{N \times N}$. Now, $e^2$ can be formulated as,

\begin{equation}
\label{sm:eq:4:3}
\begin{aligned}
e_b^2 &= \left \| A - B \right \|^2_F \\
    &= \sum_{i}(A-B)^2_{ii} + \sum_i \sum_{j \neq i}(A-B)^2_{ij} 
\end{aligned}
\end{equation}

The first term of equation \ref{sm:eq:4:3} belongs to errors of diagonal entries in B w.r.t A whilst the second term is due to the off-diagonal entries. 

Now, it follows that, 

\label{sm:eq:4:4}
\begin{center}
$\left \| \boldsymbol{I} - \boldsymbol{I}_{\text{efb}} \right \|_F \geq \left \| \boldsymbol{I} - \boldsymbol{I}_{\text{inf}} \right \|_F$ \\
$e_{efb}^2 \geq e_{inf}^2$ \\
$\sum_{i}(\boldsymbol{I}-\boldsymbol{I}_{\text{efb}})^2_{ii} + \sum_i \sum_{j \neq i}(\boldsymbol{I}-\boldsymbol{I}_{\text{efb}})^2_{ij}  \geq \sum_{i}(\boldsymbol{I} - \boldsymbol{I}_{\text{inf}})^2_{ii} + \sum_i \sum_{j \neq i}(\boldsymbol{I} - \boldsymbol{I}_{\text{inf}})^2_{ij}$  \\
$\sum_{i}(\boldsymbol{I}-\boldsymbol{I}_{\text{efb}})^2_{ii} + \sum_i \sum_{j \neq i}(\boldsymbol{I}-\boldsymbol{I}_{\text{efb}})^2_{ij}  \geq \sum_i \sum_{j \neq i}(\boldsymbol{I} - \boldsymbol{I}_{\text{inf}})^2_{ij}$ \\
$\sum_{i}(\boldsymbol{I}-\boldsymbol{I}_{\text{efb}})^2_{ii} + \sum_i \sum_{j \neq i}(\boldsymbol{I}-\boldsymbol{I}_{\text{efb}})^2_{ij}  \geq \sum_i \sum_{j \neq i}(\boldsymbol{I}-\boldsymbol{I}_{\text{efb}})^2_{ij}$
\end{center}

Note that $\sum_{i}(\boldsymbol{I} - \boldsymbol{I}_{\text{inf}})^2_{ii}=0$ using proposition 1. Furthermore, $\sum_i \sum_{j \neq i}(\boldsymbol{I} - \boldsymbol{I}_{\text{inf}})^2_{ij} = \sum_i \sum_{j \neq i}(\boldsymbol{I}-\boldsymbol{I}_{\text{efb}})^2_{ij}$ since by definition, $\boldsymbol{I}_{\text{efb}}$ and $\boldsymbol{I}_{\text{inf}}$ have the same off-diagonal terms.

\textbf{Corollary 1:} \textit{Let $\boldsymbol{I}_{\text{kfac}}$ and $\boldsymbol{I}_{\text{\text{inf}}}$ be KFAC and our estimates of real information matrix $\boldsymbol{I}$ respectively. Then, it is guaranteed to have $\left \| \boldsymbol{I} - \boldsymbol{I}_{\text{kfac}} \right \|_F \geq \left \| \boldsymbol{I} - \boldsymbol{I}_{\text{\text{inf}}} \right \|_F$.}

For interested readers, find the proof $\left \| \boldsymbol{I} - \boldsymbol{I}_{kfac} \right \|_F \geq \left \| \boldsymbol{I} - \boldsymbol{I}_{\text{efb}} \right \|_F$ in \citet{George2018}.

\subsection{Properties of Low-Rank Information Matrix}

\textbf{Lemma 2: } \textit{Let $\boldsymbol{I}$ be the real Fisher information matrix, and let $\boldsymbol{\hat{I}}_{\text{\text{inf}}}$, $\boldsymbol{I}_{\text{efb}}$ and $\boldsymbol{I}_{\text{kfac}}$ be the low rank INF, EFB and KFAC estimates of it respectively. Then, it is guaranteed to have $\left \| \text{ diag}(\boldsymbol{I}) - \text{ diag}(\boldsymbol{I}_{\text{efb}}) \right \|_F \geq \left \| \text{ diag}(\boldsymbol{I}) - \text{ diag}(\boldsymbol{\hat{I}}_{\text{inf}}) \right \|_F = 0$ and $\left \| \text{ diag}(\boldsymbol{I}) - \text{diag}(\boldsymbol{I}_{\text{kfac}}) \right \|_F \geq \left \| \text{  diag}(\boldsymbol{I}) - \text{ diag}(\boldsymbol{\hat{I}}_{\text{inf}}) \right \|_F = 0$. Furthermore, if the eigenvalues of $\boldsymbol{\hat{I}}_{\text{\text{inf}}}$ contains all non-zero eigenvalues of $\boldsymbol{I}_{\text{\text{inf}}}$, it follows: $\left \| \boldsymbol{I} - \boldsymbol{I}_{\text{efb}} \right \|_F \geq \left \| \boldsymbol{I} - \boldsymbol{\hat{I}}_{\text{inf}} \right \|_F$.}

\textit{proof:} The first part follows from proposition 1 which states that for all the elements i, $\boldsymbol{I}_{ii} = \hat{\boldsymbol{I}}_{\text{inf}}$, $\left \| \text{ diag}(\boldsymbol{I}) - \text{diag}(\boldsymbol{I}_{\text{efb}}) \right \|_F \geq \left \| \text{ diag}(\boldsymbol{I}) - \text{diag}(\boldsymbol{\hat{I}}_{\text{inf}}) \right \|_F = 0$ and $\left \| \text{ diag}(\boldsymbol{I}) - \text{diag}(\boldsymbol{I}_{\text{kfac}}) \right \|_F \geq \left \| \text{ diag}(\boldsymbol{I}) - \text{diag}(\boldsymbol{\hat{I}}_{\text{inf}}) \right \|_F = 0$. This results by the design of the method, in which, we correct the diagonals in parameter space after LRA.

For the second part of the proof, lets recap that Lemma 2 (Wely's idea on eigenvalue perturbation) that removing zero eigenvalues does not affect the approximation error in terms of Frobenius norm. This then implies that off-diagonal elements of $\boldsymbol{\hat{I}}_{\text{inf}}$ and $\boldsymbol{I}_{\text{efb}}$ are equivalent. Then,:

\begin{center}
$\left \| \boldsymbol{I} - \boldsymbol{I}_{\text{efb}} \right \|_F \geq \left \| \boldsymbol{I} - \boldsymbol{\hat{I}}_{\text{inf}} \right \|_F$ \\
$e_{efb}^2 \geq e_{inf}^2$ \\
$\sum_{i}(\boldsymbol{I}-\boldsymbol{I}_{\text{efb}})^2_{ii} + \sum_i \sum_{j \neq i}(\boldsymbol{I}-\boldsymbol{I}_{\text{efb}})^2_{ij}  \geq \sum_{i}(\boldsymbol{I} - \boldsymbol{\hat{I}}_{\text{inf}})^2_{ii} + \sum_i \sum_{j \neq i}(\boldsymbol{I} - \boldsymbol{\hat{I}}_{\text{inf}})^2_{ij}$  \\
$\sum_{i}(\boldsymbol{I}-\boldsymbol{I}_{\text{efb}})^2_{ii} + \sum_i \sum_{j \neq i}(\boldsymbol{I}-\boldsymbol{I}_{\text{efb}})^2_{ij}  \geq \sum_i \sum_{j \neq i}(\boldsymbol{I} - \boldsymbol{\hat{I}}_{\text{inf}})^2_{ij}$ \\
$\sum_{i}(\boldsymbol{I}-\boldsymbol{I}_{\text{efb}})^2_{ii} + \sum_i \sum_{j \neq i}(\boldsymbol{I}-\boldsymbol{I}_{\text{efb}})^2_{ij}  \geq \sum_i \sum_{j \neq i}(\boldsymbol{I}-\boldsymbol{I}_{\text{efb}})^2_{ij}$
\end{center}

Again, $\sum_{i}(\boldsymbol{I} - \boldsymbol{\hat{I}}_{\text{inf}})^2_{ii} = 0$ according to proposition 1 for all the elements i, which completes the proof.

\textbf{Lemma 3: } \textit{The low rank matrix $\hat{\Sigma} = \left ( (U_a \otimes U_g)\Lambda_{1:L} (U_a \otimes U_g)^T + D \right )^{-1} \in \mathbb{R}^{N \times N}$ is a non-degenerate covariance matrix if the diagonal correction matrix D and LRA $(U_a \otimes U_g)\Lambda_{1:L} (U_a \otimes U_g)^T$ are both symmetric and positive definite. This condition is satisfied if $(U_a \otimes U_g)\Lambda_{1:L} (U_a \otimes U_g)_{ii}^T < \mathbb{E} \left [ \delta \theta_i^2 \right ]$ for all i $\in \left \{ 1, 2, \cdots , N \right \}$ and with $\Lambda_{1:L} \nsubseteq {0}$.}

\textit{proof:} 
Let us first rewrite $\hat{\boldsymbol{I}}_{\text{inf}} = (U_a \otimes U_g)\Lambda_{1:L} (U_a \otimes U_g)^T + D$ in the following form.

\begin{equation}
\begin{aligned}
   & (U_a \otimes U_g)\Lambda_{1:L} (U_a \otimes U_g)^T + D = \\
   & (U_a \otimes U_g)\Lambda_{1:L}^{\frac{1}{2}} \Lambda_{1:L}^{\frac{1}{2}}(U_a \otimes U_g)^T+ D \\
    &= \left [(U_a \otimes U_g)\Lambda_{1:L}^{\frac{1}{2}}\right ] \left [(U_a \otimes U_g)\Lambda_{1:L}^{\frac{1}{2}}\right ]^T + D\\
    &= UU^T + D
\end{aligned}
\end{equation}

Now, if $D$ and $(U_a \otimes U_g)\Lambda_{1:L} (U_a \otimes U_g)^T$ is both symmetric and positive definite, it follows that for an arbitrary vector $x \in \mathbb{R}^{d}$, $x^TUU^Tx > 0$ as eigenvalues $R_i > 0$ by construction. Furthermore, $x^TDx > 0$ also holds by the definition of positive definiteness. Therefore, we have $x^T(UU^T + D)x = x^TUU^Tx + x^TDx > 0$ which leads to the proof that $\boldsymbol{I}_{\text{inf}}$ is positive definite if $D$ and $(U_a \otimes U_g)\Lambda_{1:L} (U_a \otimes U_g)^T$ is both symmetric and positive definite. As this results in non-degenerate IM, the covariance $\Sigma$ is non-degenerate as well.

Trivially following the definition of $D_{ii} =  \mathbb{E} \left [ \delta \theta_i^2 \right ] - (U_A \otimes U_G)\Lambda (U_A \otimes U_G)_{ii}^T$, $D_{ii} > 0$ for all i when $(U_a \otimes U_g)\Lambda_{1:L} (U_a \otimes U_g)_{ii}^T < \mathbb{E} \left [ \delta \theta_i^2 \right ]$. Again, by the definition of $\Lambda_{ii} = \mathbb{E}\left [ (V^T \delta \theta)_{i}^2 \right ] \geq 0$, $\Lambda_{1:L}$ containing no zero eigenvalues result in the positive definite matrix $(U_a \otimes U_g)\Lambda_{1:L} (U_a \otimes U_g)^T$, which completes the proof.

\textbf{Lemma 4: } \textit{Let $\boldsymbol{I} \in \mathbb{R}^{N \times N}$ be the real Fisher information matrix, and let $\boldsymbol{\hat{I}}_{1:K}^{\text{top}} \in \mathbb{R}^{N \times N}$, $\boldsymbol{\hat{I}}_{1:L}^{\text{top}} \in \mathbb{R}^{N \times N}$ and $\boldsymbol{\hat{I}}_{1:L} \in \mathbb{R}^{N \times N}$ be the low rank estimates of $\boldsymbol{I}$ of EFB obtained by preserving top K, L and top K plus additional J resulting in L eigenvalues. Here, we define $K < L$. Then, the approximation error of $\boldsymbol{\hat{I}}_{1:L}$ is as follows: $\left \| \boldsymbol{I} - \boldsymbol{\hat{I}}_{1:L}^{\text{top}} \right \|_F \geq \left \| \boldsymbol{I} -\boldsymbol{\hat{I}}_{1:L} \right \|_F \geq \left \| \boldsymbol{I} - \boldsymbol{\hat{I}}_{1:K}^{\text{top}} \right \|_F$.}

\textit{proof:} From the definition, $(U_A \otimes U_G)\Lambda (U_A \otimes U_G)^T = V\Lambda V^T$ is PSD as $\Lambda_{ii} = \mathbb{E}\left [ (V^T \delta \theta)_{i}^2 \right ] \geq 0$ for all elements i and $VV^T = I$ with $I$ as an identity matrix (orthogonality). Naturally, low rank approximations $(U_A \otimes U_G)_{1:L^{\text{top}}}\Lambda_{1:L^{\text{top}}} (U_A \otimes U_G)_{1:L^{\text{top}}}^T$, $(U_A \otimes U_G)_{1:K^{\text{top}}}\Lambda_{1:K^{\text{top}}}(U_A \otimes U_G)_{1:K^{\text{top}}}^T$ and  $(U_a \otimes U_g)\Lambda_{1:L} (U_a \otimes U_g)^T = (U_A \otimes U_G)_{1:L}\Lambda_{1:L} (U_A \otimes U_G)_{1:L}^T$ are again PSD by the fact that low rank approximation does not introduce negative eigenvalues. 

Now, a well known fact from dimensional reduction literature is that low rank approximation preserving the top eigenvalues result in best approximation errors in terms of Frobenius norm for the given rank. Informally stating Wely's ideas on eigenvalue perturbation:

Let $B \in \mathbb{R}^{m \times n}$ with rank smaller or equal to p (one can also use complex space $\mathbb{C}$ instead of $\mathbb{R}$) and let $E = A-B$ with $A\in \mathbb{R}^{m \times n}$. Then, it follows that,

\begin{equation}
\begin{aligned}
    \left \| A - B \right \|_F^2 &= \sigma_1(A-B)^2 + \cdots + \sigma_\mu(A-B)^2 \\ 
    & \geq  \sigma_{p+1}(A-B)^2 + \cdots + \sigma_\mu(A-B)^2 \\
    & = \left \| A - B_{1:p} \right \|_F^2,
\end{aligned}
\end{equation}

where $\sigma_1, \cdots \sigma_\mu$ are the singular values of A with $\mu = \text{min}(n,m)$. The convention here is that $\sigma_i(A)$ is the ith largest singular value and $\sigma_i(A)=0$ for $i>\text{rank}$(A). Using this insight, and the fact that in the given settings, squared singular values are variances in new space lead to:

\begin{center}
$ \left \| \boldsymbol{I} - \boldsymbol{\hat{I}}_{1:K}^{\text{top}} \right \|_F  \geq \left \| \boldsymbol{I} - \boldsymbol{\hat{I}}_{1:L} \right \|_F \geq \left \| \boldsymbol{I} - \boldsymbol{\hat{I}}_{1:L}^{\text{top}} \right \|_F$ \\
\end{center}

This completes the proof of Lemma 4.

%% file: supplementary/implementation_details.tex
\section{Implementation Details}
\label{sm:sec:5}
\begin{table*}
\scriptsize
\centering
\caption{\textbf{UCI benchmark.} Root mean squared error (RMSE) is reported for the used set-up of UCI benchmark experiments. Note that as test log-likelihood depends on accuracy (in addition to uncertainty estimation), we have used linearized LA on the output space so that the test accuracy or RMSE is kept the same amongst the compared LA-based approaches. Our model overfits in some datasets so that effectiveness of having Bayesian Neural Network can be seen clearly. This also does not affect our results as all LA-based methods are built on top of the same model.}
\label{results:uci:rmse}
\begin{tabular}{cccccccccccc}
\toprule
Datasets & Boston & Concrete       & Energy    & Kin8nm   & Naval & Power &  Protein     & Wine   &   Yacht \\
\midrule
RMSE & 3.361$\pm$0.929 & 6.181$\pm$0.727     & 0.573$\pm$0.070    & 0.164$\pm$0.005 & 0.010$\pm$0.0001  & 4.322$\pm$0.153 & 4.516$\pm$0.123 & 0.637$\pm$0.034 & 9.568$\pm$1.132 \\
\bottomrule
\end{tabular}
\end{table*}

\begin{figure*}
\minipage{0.1\textwidth}
  \includegraphics[width=\linewidth]{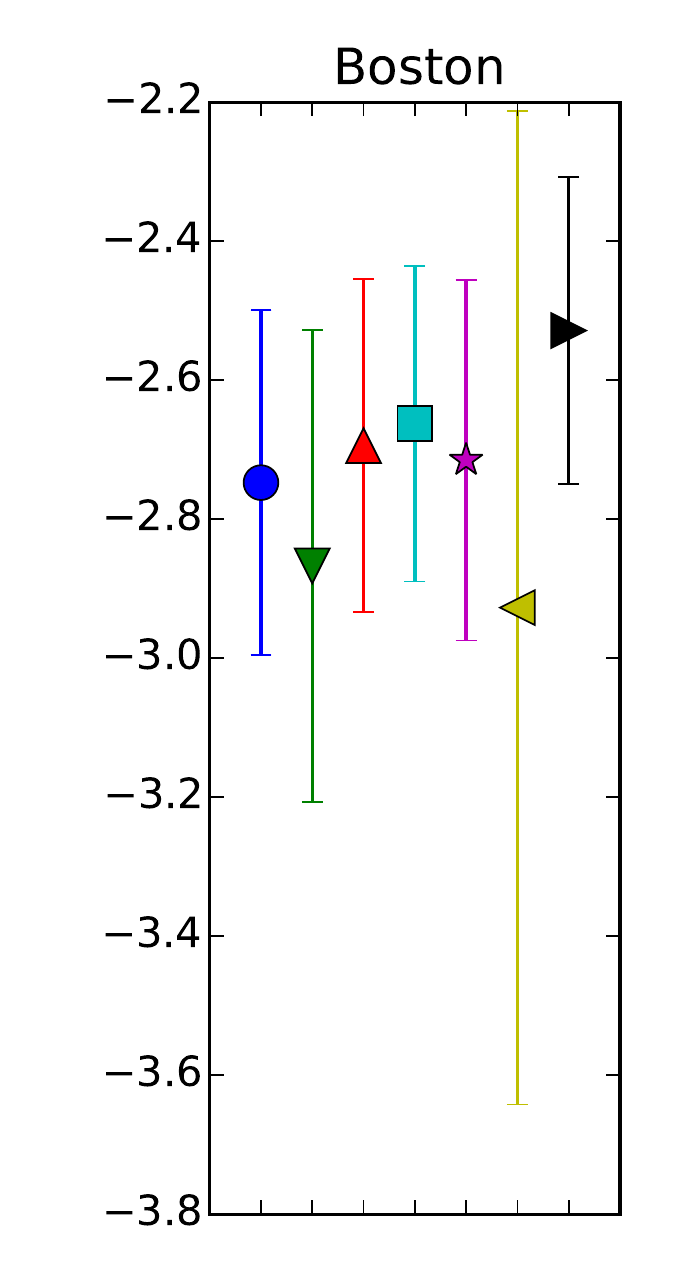}
\endminipage \hfill
\minipage{0.1\textwidth}
  \includegraphics[width=\linewidth]{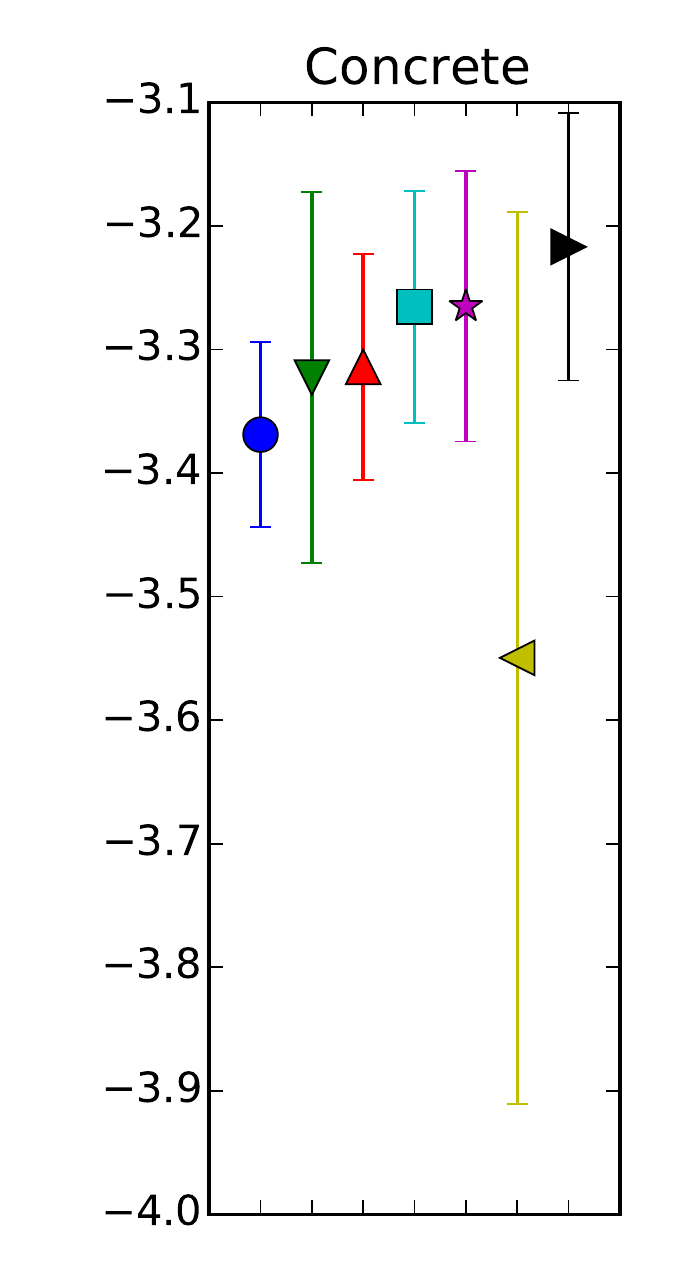}
\endminipage \hfill
\minipage{0.1\textwidth}
  \includegraphics[width=\linewidth]{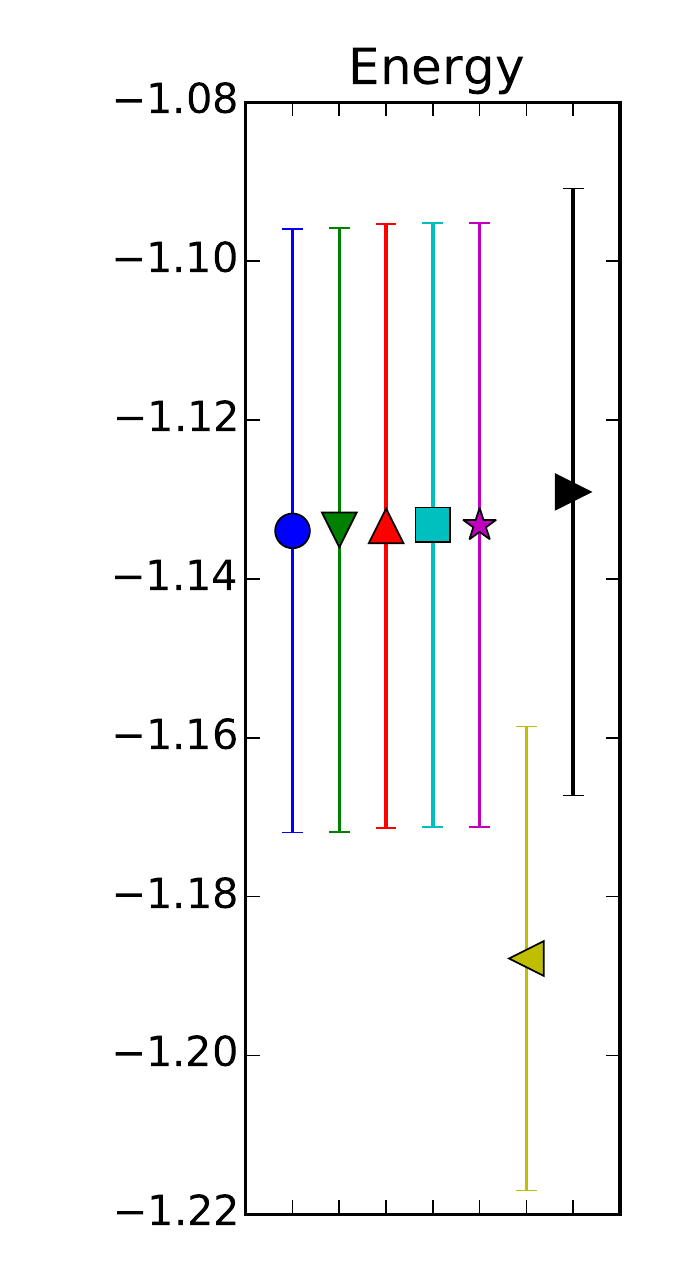}
\endminipage \hfill
\minipage{0.1\textwidth}
  \includegraphics[width=\linewidth]{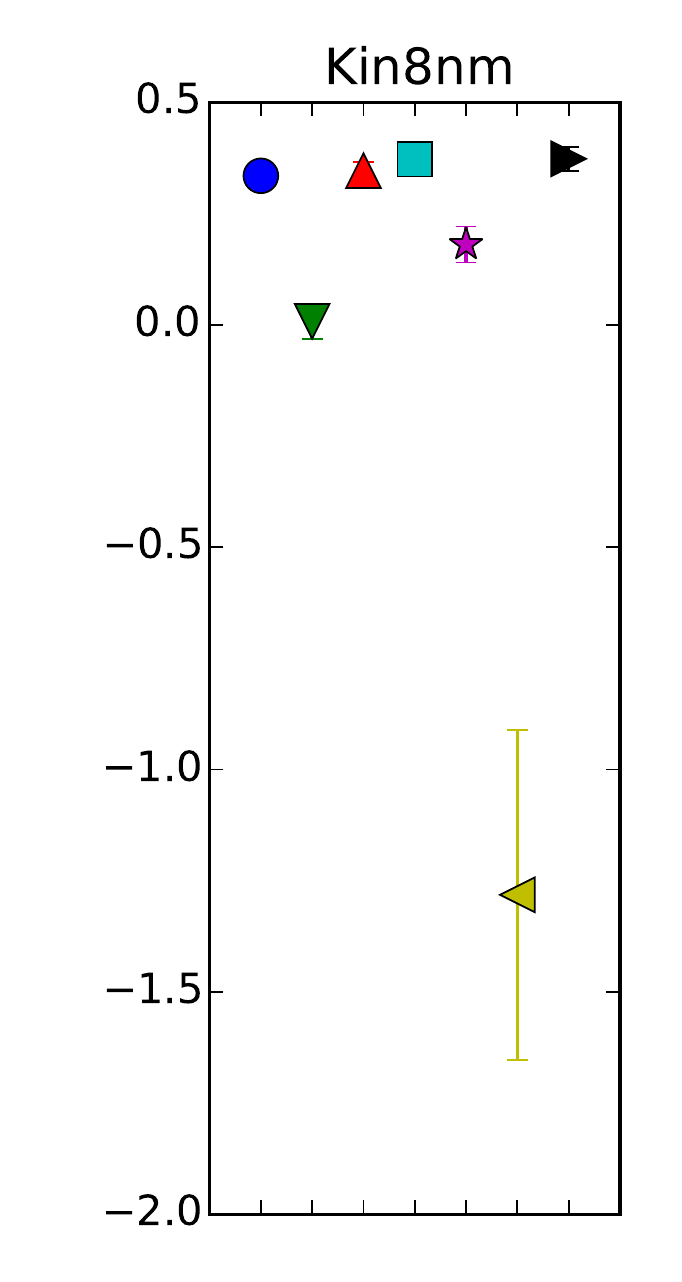}
\endminipage \hfill
\minipage{0.1\textwidth}
  \includegraphics[width=\linewidth]{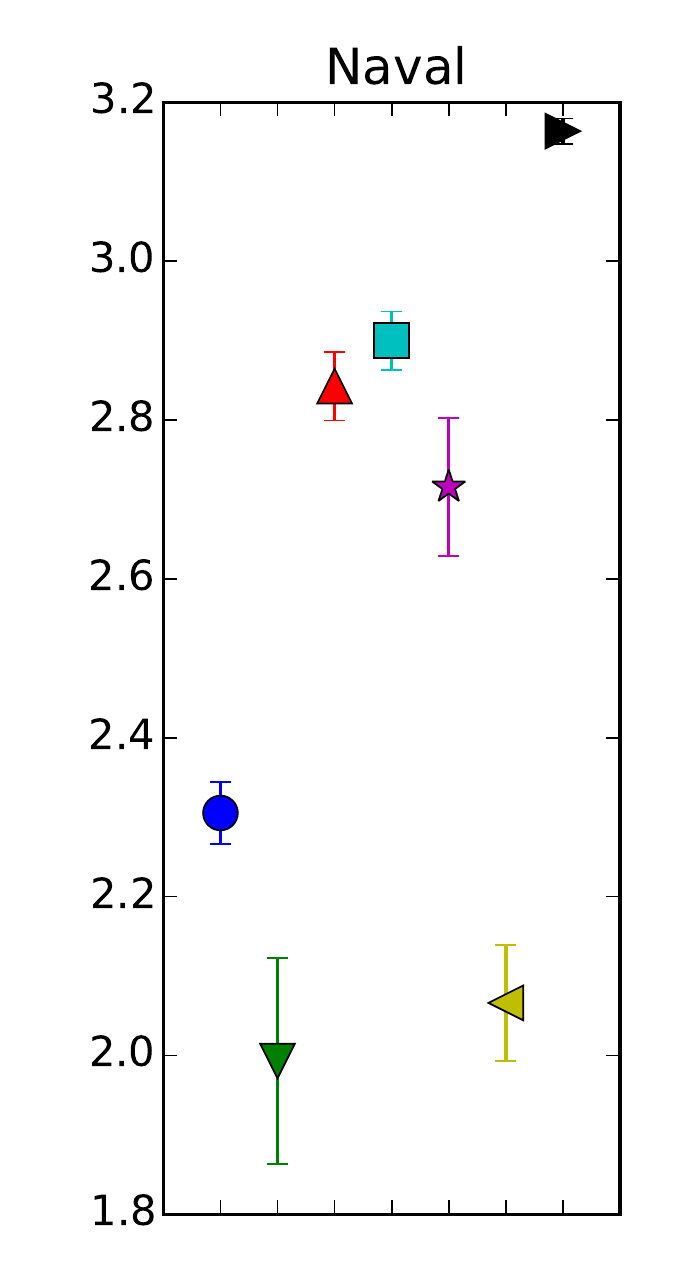}
\endminipage \hfill
\minipage{0.1\textwidth}
  \includegraphics[width=\linewidth]{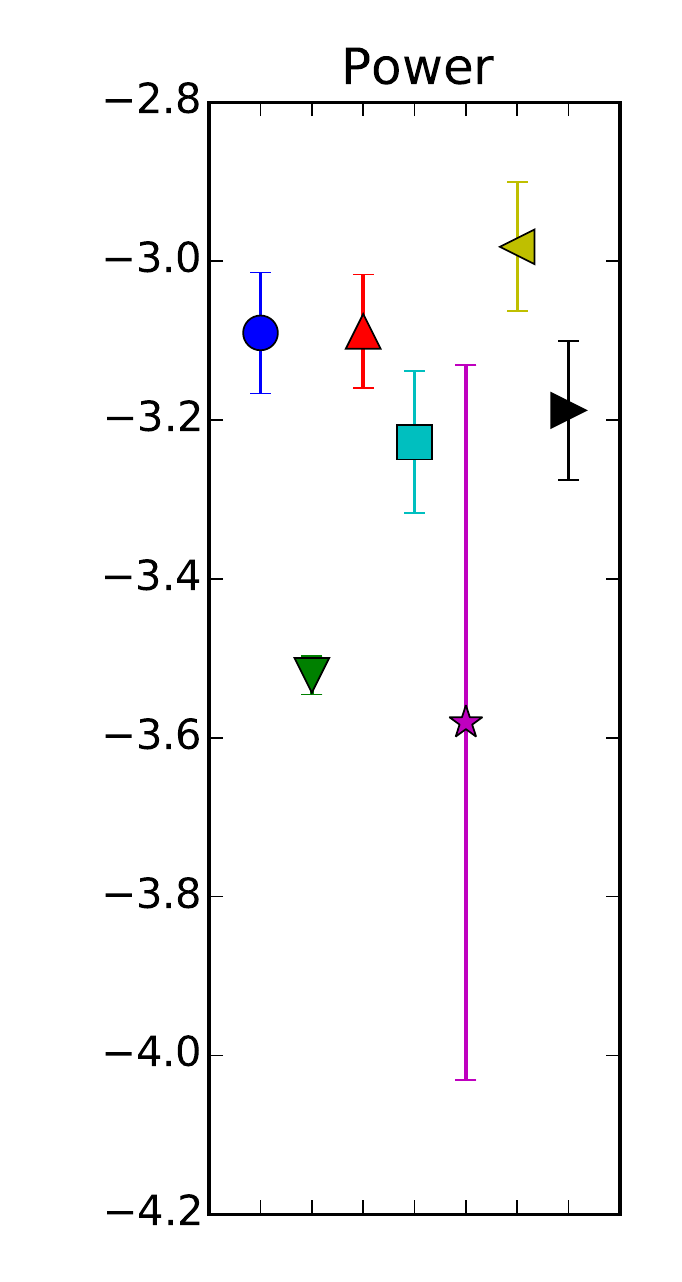}
\endminipage \hfill
\minipage{0.1\textwidth}
  \includegraphics[width=\linewidth]{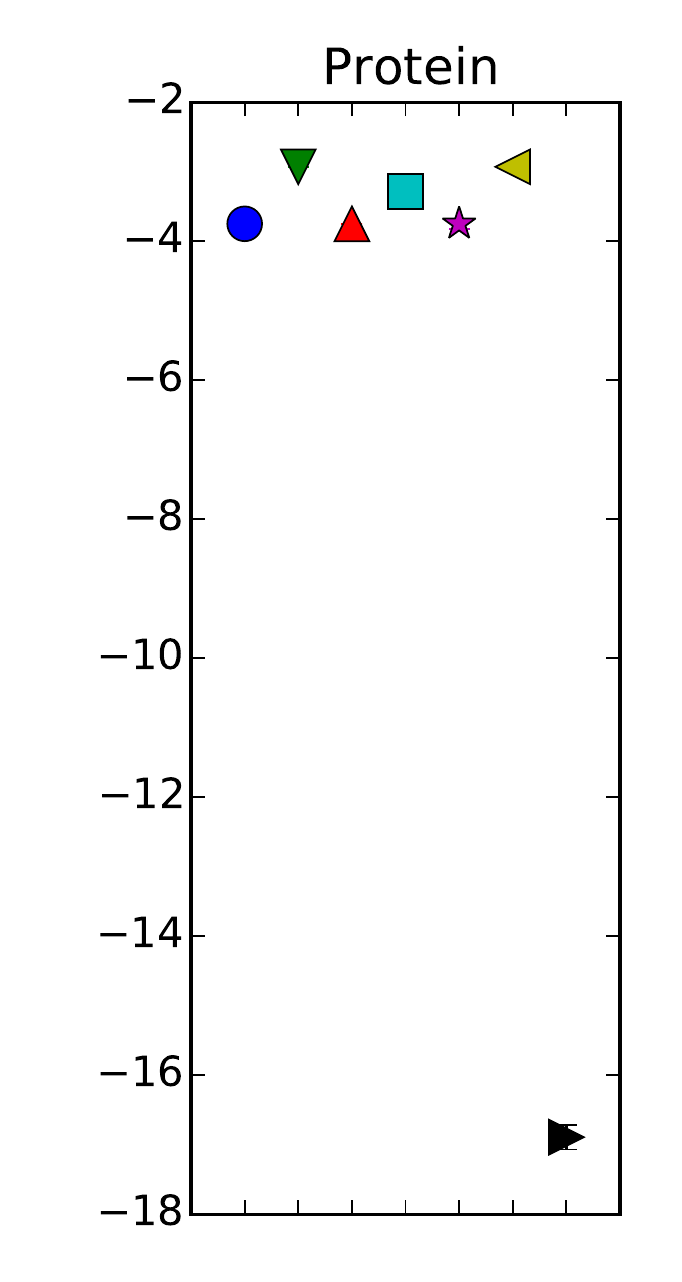}
\endminipage \hfill
\minipage{0.1\textwidth}
  \includegraphics[width=\linewidth]{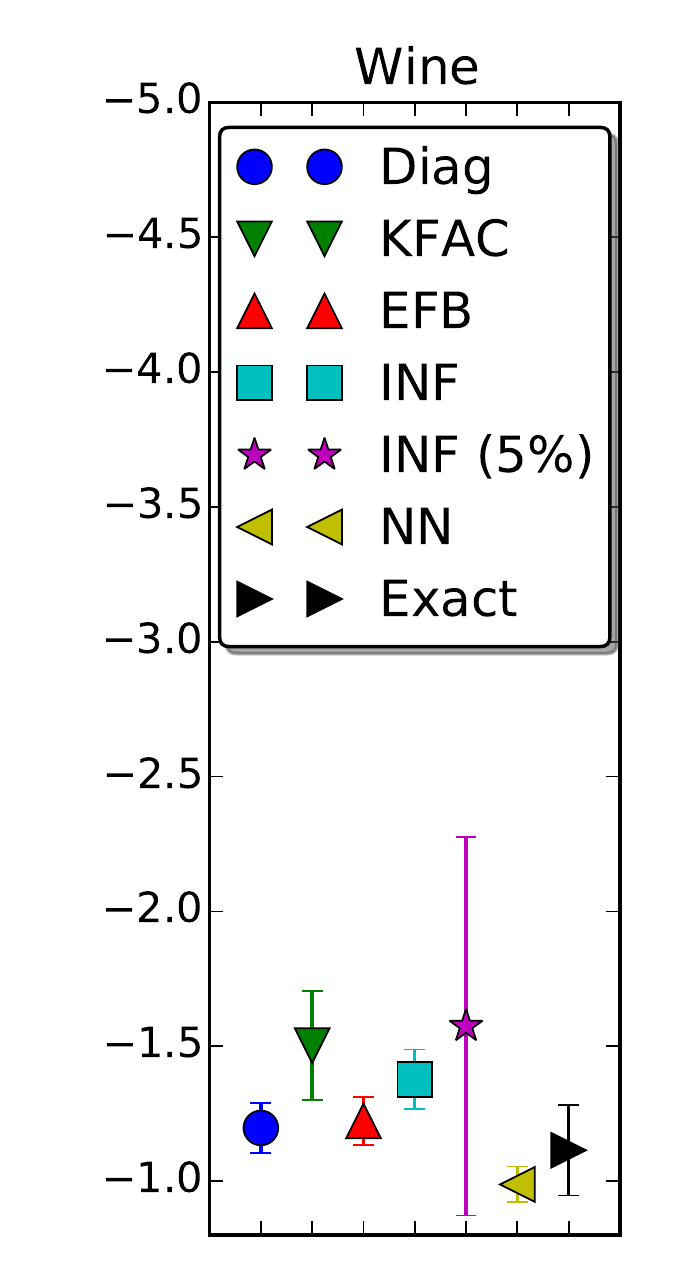}
\endminipage \hfill
\minipage{0.1\textwidth}
  \includegraphics[width=\linewidth]{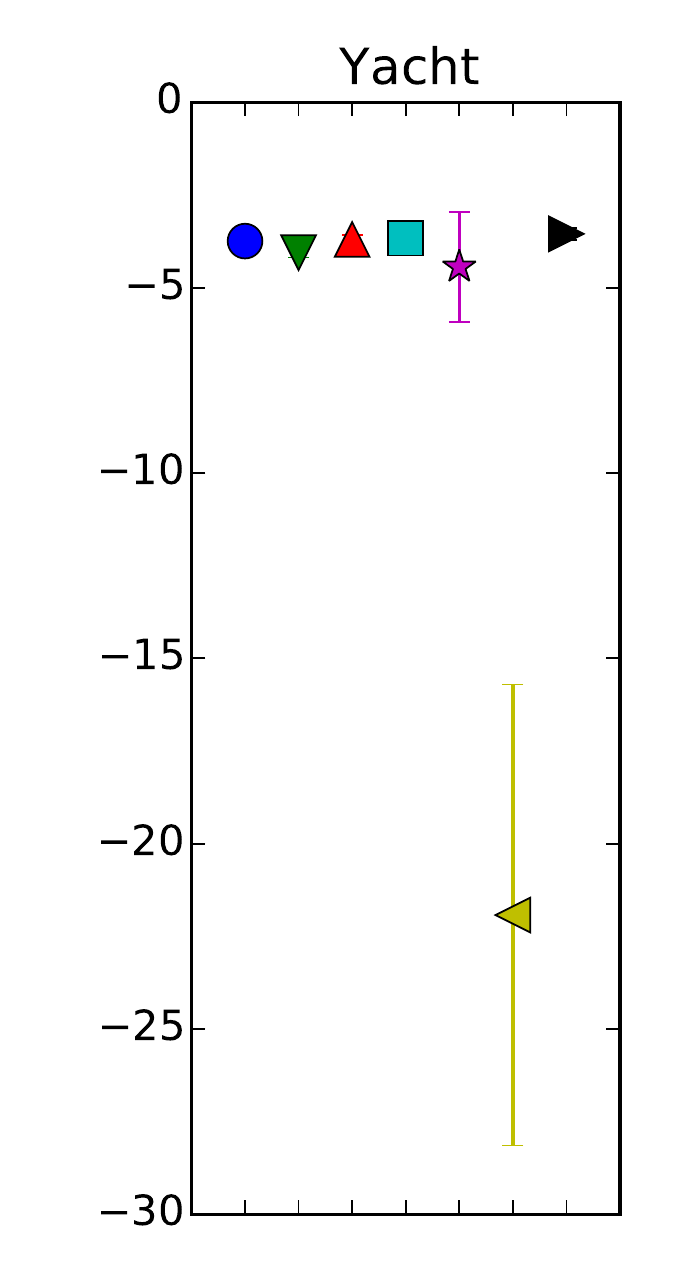}
\endminipage \hfill
\caption{\textbf{Evaluating predictive uncertainty on UCI datasets.} We report test log likelihood on the y-axis and compare Diag, KFAC, EFB, INF variants, NN and Exact. Here, exact refers Laplace Approximation using the block-wise exact information matrix while NN denotes a deterministic neural network. Using linear approximation to the predictive uncertainty \citep{mackay1992information}, accuracy between each methods are kept the same, which ensures fair comparisons using test log likelihood. Higher the better.}
\label{fig:uci}
\end{figure*}

The following experiments are implemented using Tensorflow \citep{tf2016}: (i) toy regression, (ii) UCI benchmark, (iii) active learning and (iv) classification on MNIST and CIFAR10. The KFAC library from Tensorflow \footnote{Available at \url{https://github.com/tensorflow/kfac}} was used to implement the Fisher estimator \citep{MartensG15} for our methods and the works of \citet{Ritter2017ASL}. On the other hand, Pytorch \citep{paszke2019pytorch} has been used for ImageNet and adversarial defense experiments \footnote{Available at \url{TBD}}. The plug-and-play code is made available for Pytorch.

Note that empirical Fisher usually is not a good estimate of the Hessian, as it is typically biased \citep{MartensG15, kunstner2019limitations}. Instead, KFAC library offers several estimation modes. We have used the gradients mode for KFAC whereas the exact mode was used for Diag. NVIDIA Tesla and 1080Ti are used for all the experiments. 

\subsection{Small scale experiments}
The training details are as follows: Adam has been used with a learning rate of 0.001 with zero prior precision or L2 regularization coefficient ($\tau = 0.2$ for KFAC, $\tau = 0.45$ for Diag, $N=1$ and $\tau = 0$ for both FB and INF have been used). Mean squared error (MSE) loss is used. The exact block-wise Hessian and their approximations for the given setup contained zero values on its diagonals. This can be interpreted as zero variance in the IM, meaning no information, resulting in an IM being degenerate for the likelihood term. In such cases, the covariance may not be uniquely defined \citep{Thrun03d}. Therefore, we treated these variances as deterministic, making the IM non-degenerate (similar findings reported by \citet{MacKay1992}). We have used Numpyro \citep{phan2019composable} for the implementations of HMC, with 50000 samples. For BBB, we have used an open-source implementation \footnote{Available at \url{https://github.com/ThirstyScholar/bayes-by-backprop}} where the Gaussian noise is sampled in a batch initially, and a symmetric sampling is deployed. Lastly, $K_{mc}=100$ samples were used.

Experiments on UCI benchmark \footnote{Dataset and splits have been taken from \url{https://github.com/yaringal/DropoutUncertaintyExps}.} have been conducted to evaluate various IM estimates and their effects on predictive uncertainty estimation. We evaluate LA-based approaches only which has an advantage that the only differences between each approaches are approximations of IM. To explain, inference principle, network architectures, and training convergence can be kept the same. We note that, this is difficult for approaches based on variational inference. Due to this, meaningful comparisons can be often difficult as specific details such as number of epochs can have significant effects on the results \citep{Mukhoti2018}. In this line of argument, we have further used so-called linearized LA \citep{mackay1992information, foong2019between} instead of sampling based evaluation \citep{Gal2016Uncertainty, Ritter2017ASL}:

\begin{equation*}
p(y^*|x^*,x,y) \approx \mathcal{N}(f_{\theta_{\text{MAP}}}(x^*), \Sigma_{\text{alea}} + \delta \theta(x^*)^T \Sigma_{\text{epis}}\delta \theta(x^*)).
\end{equation*}

As shown, the mean of prediction $f_{\theta_{\text{MAP}}}(x^*)$ depends only on the new test data $x^*$. As test log-likelihood depends on the accuracy of the predictor, we can keep the accuracy of the predictors the same among-st various LA-based approaches (reported in table \ref{results:uci:rmse}). Furthermore, as the covariance matrix for predictive uncertainty only depends on \textit{aleatoric} uncertainty $\Sigma_{\text{alea}}$, gradients on the new test input $\delta \theta(x^*)$ and episdemic uncertainty $\Sigma_{\text{epis}}$, comparisons between LA-based approaches are simpler as the experiments can be implemented in a way that the difference lies only in various approximations of $\Sigma_{\text{epis}}$. Closely following \citet{foong2019between}, layer-wise exact IM has been established and the same hyperparameter settings are applied across other LA-based approaches \footnote{We also present the results of sampling based evaluations in section \ref{sm:sec:6:hyp} where we extensively search hyperparameters for each LA methods separately.}. Figure \ref{fig:uci} reports the results of UCI experiments where we compare the reliability of uncertainty estimates using the test log-likelihood as a measure. As shown, we find that our approach compares well to the others. Note that in Power and Protein, LA approaches were under-performing even when compared to a deterministic DNN. This is a known limitation of LA: the approximated posterior may cover areas of low probability mass, the approaches perform similar to a deterministic DNN or become unstable. Here, our experiments also indicate that improvements in terms of Frobenius norm of error may not directly translate to performance in uncertainty estimation, atleast for LA, which requires in-depth treatment for future works. 

\subsection{Active Learning Experiments}

Details of experiment settings are as follows: we split each one into training, validation, and test set with 20, 100 (which is reasonable when compared with the size of test set) and 100 data points randomly for $20$ times, respectively. The remaining points serve as pool set, in which we are not allowed to obtain their labels. So we have $20$ splits in this experiment. Once the model chooses the data point in the pool set and move it into the training set, its label becomes available, which simulates the scenario where humans annotate it. The experiments progress as follows: firstly, we trained the model with the initial training set and select one point from pool set which will be put into the training set with its label. Then we train the model again and proceed into the next iteration. In each iteration, we evaluate our model on the test set and report the root mean square error (RMSE). We select the model during training and the corresponding hyperparameter ($\tau$) based on its performance on the validation set.  While the range of $N$ is $[0.5*N, 1.0*N]$, where N is the size of the training set in current iteration. The range of $\tau$ is $[1, 200, 400]$. 
For other hyperparameters, we use learning rate of $0.01$ for boston housing and energy, $0.001$ for wine and L2 regularization of $0.0$ for boston housing and wine, $1e-5$ for energy. The mini-batch size is set to the initial size of training set, $20$. Only $1$ point is selected in each iteration and the number of iteration is set to $20$. Regarding model selection, we use early stopping based on the RMSE on the validation set and the maximum epoch is set to $40$. 

\begin{figure*}[ht]
\minipage{0.25\textwidth}%
  \includegraphics[width=\linewidth]{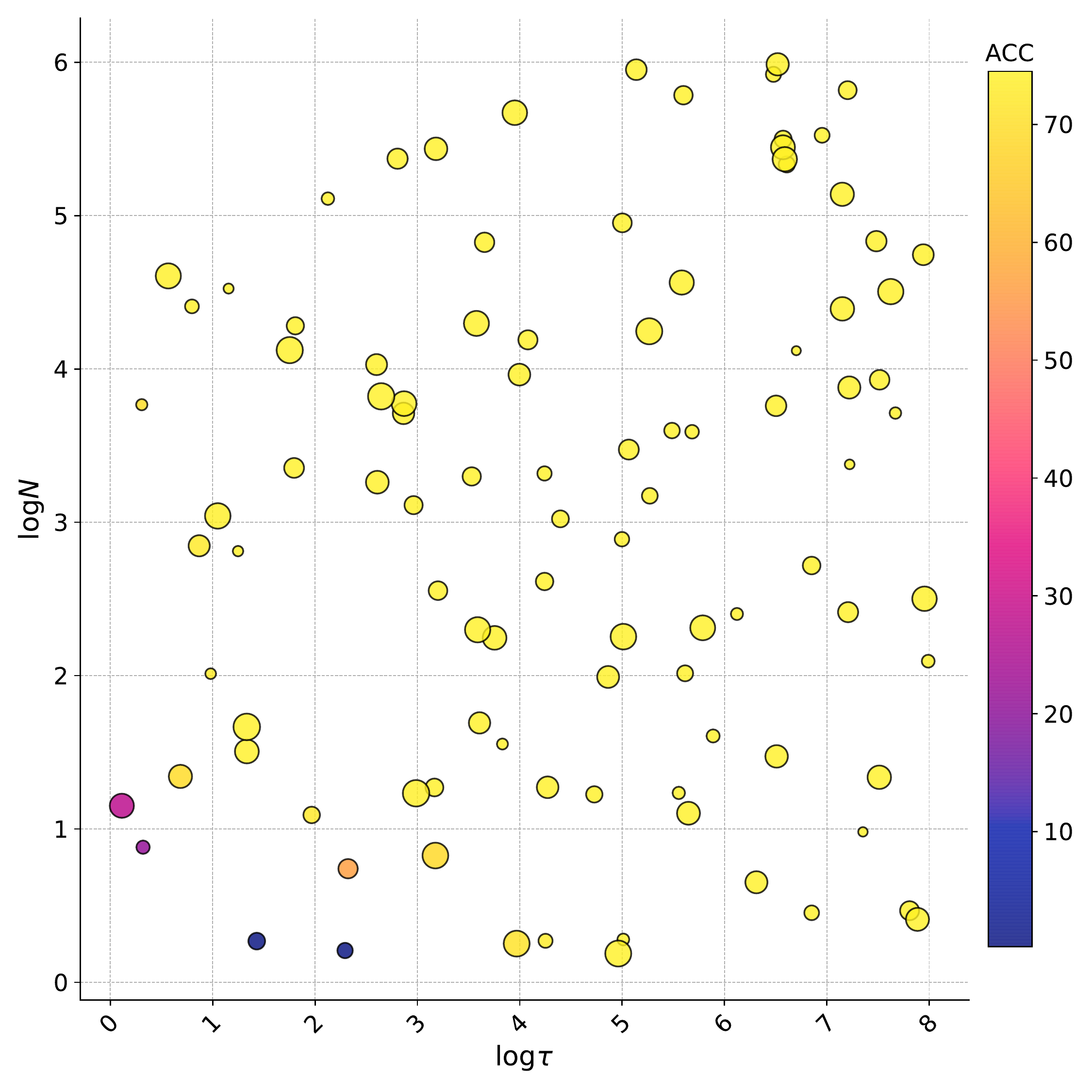}
\endminipage \hfill
\minipage{0.25\textwidth}
  \includegraphics[width=\linewidth]{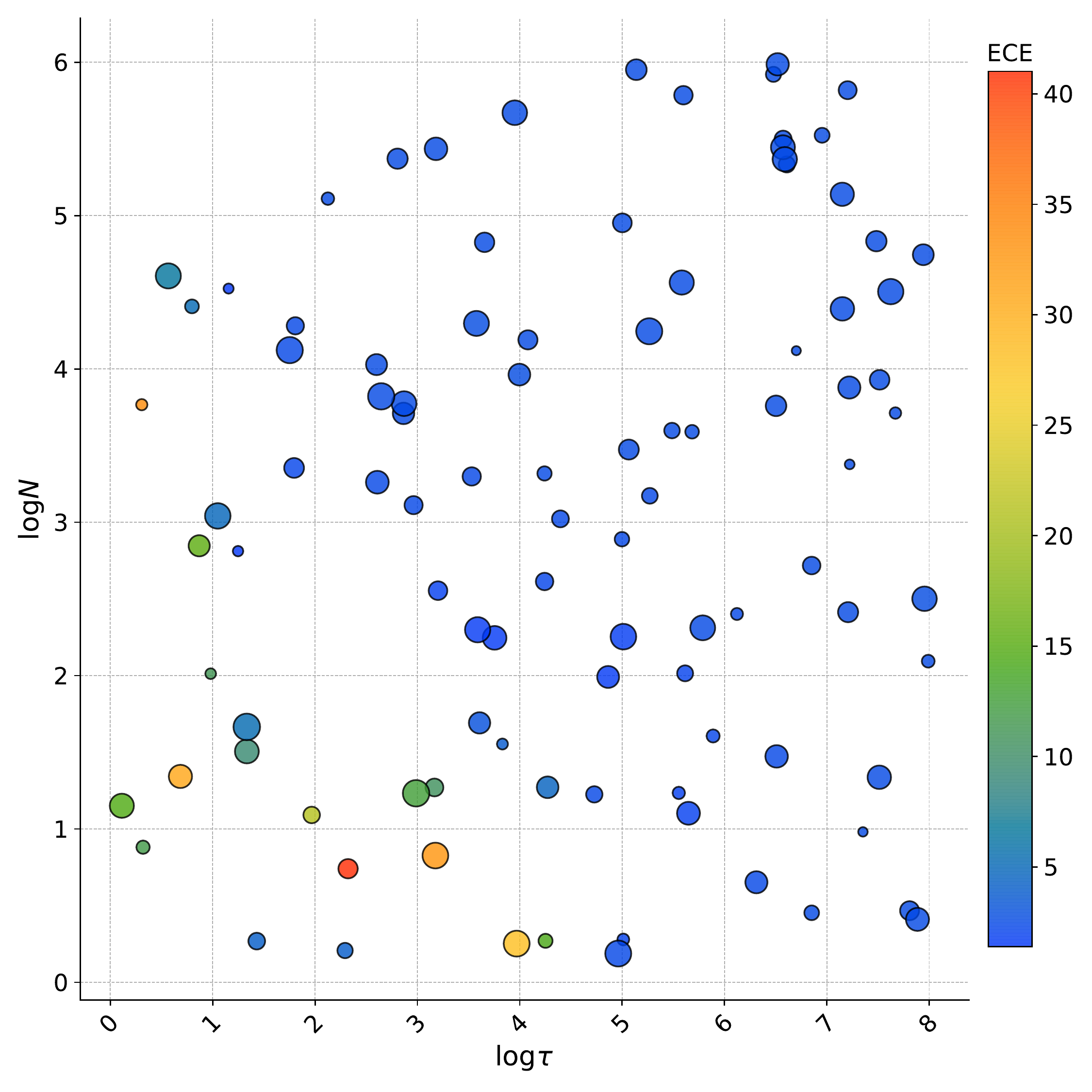}
\endminipage \hfill
\minipage{0.25\textwidth}%
  \includegraphics[width=\linewidth]{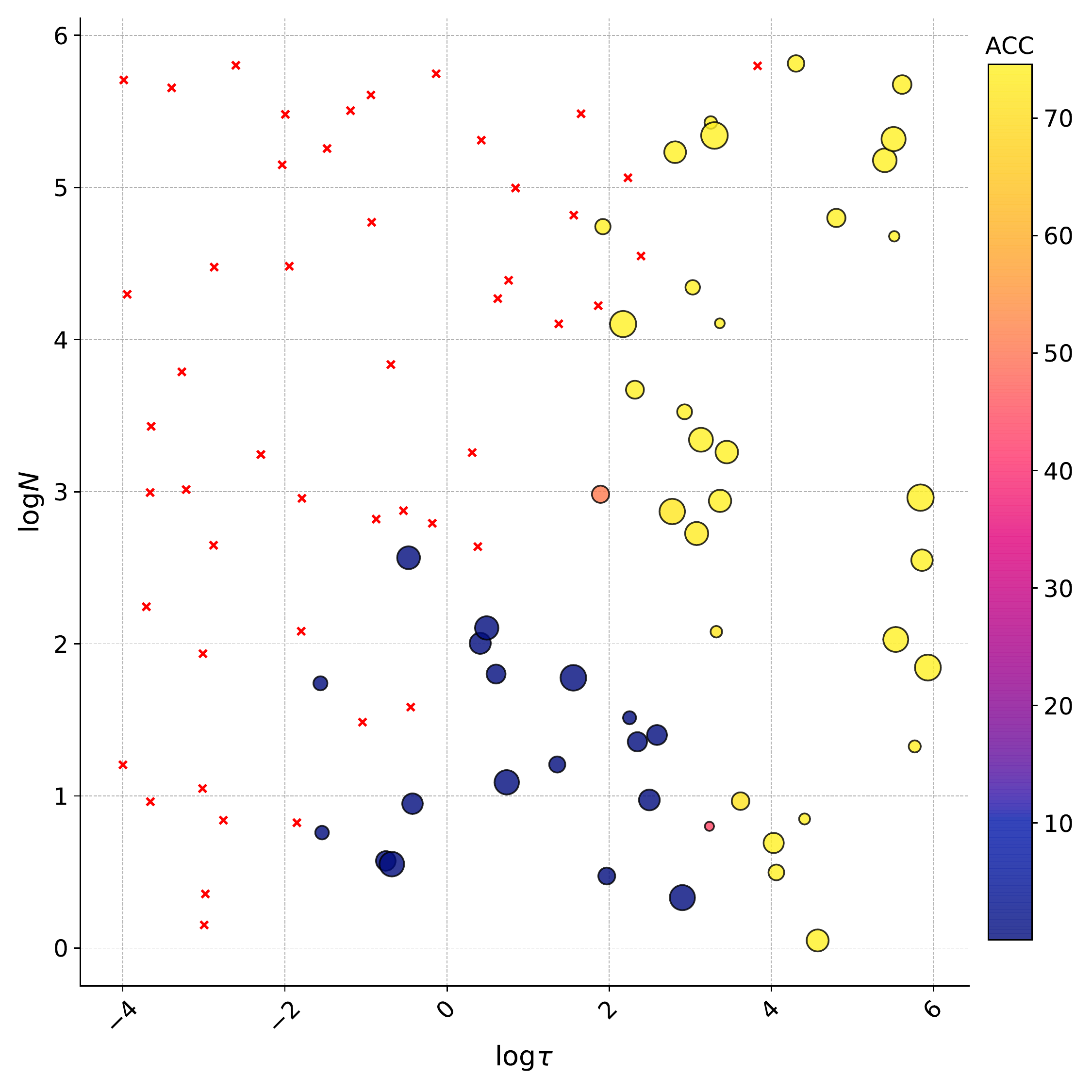}
\endminipage \hfill
\minipage{0.25\textwidth}
  \includegraphics[width=\linewidth]{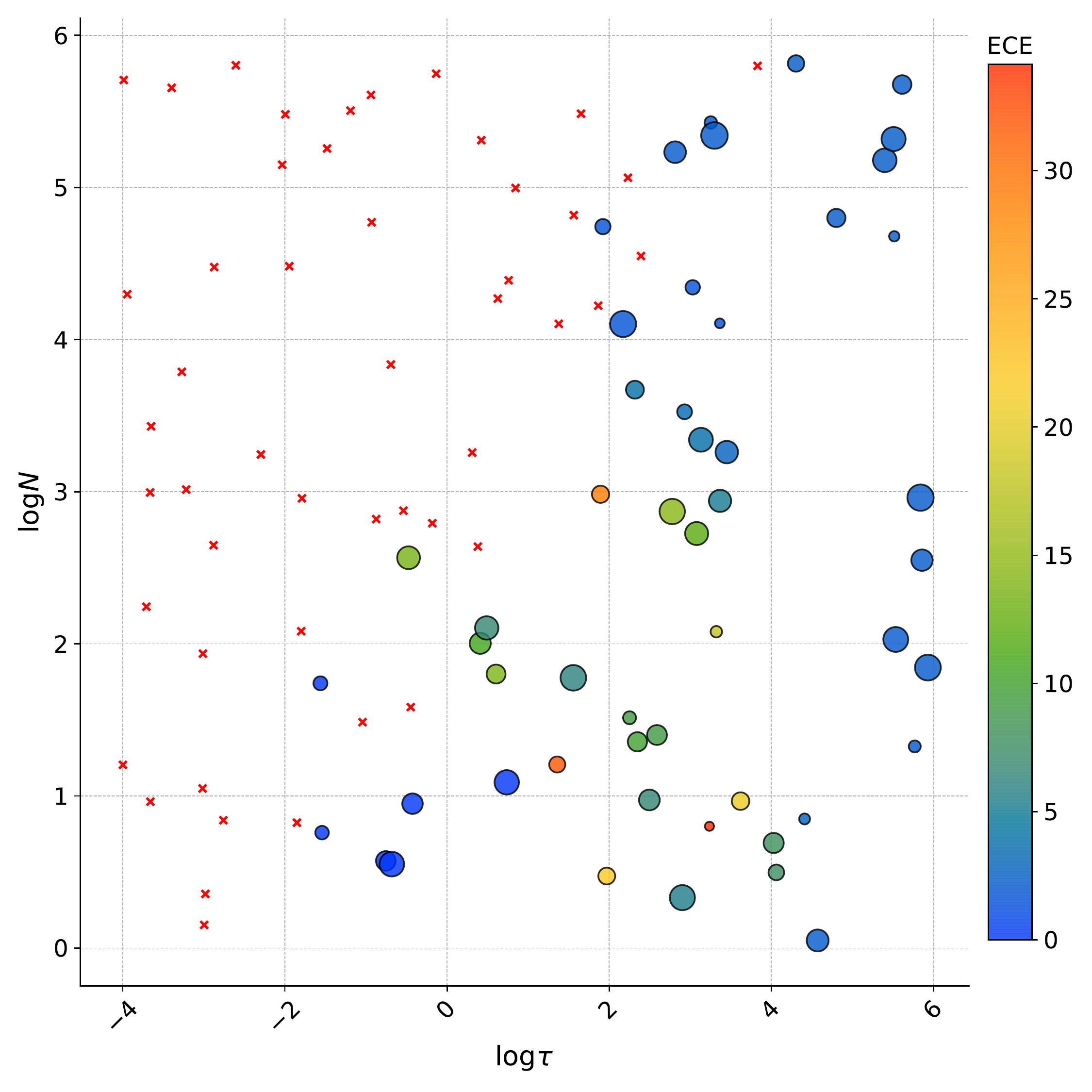}
\endminipage \hfill
\caption{\textbf{Hyperparameter search} Results of random hyperparameter search for DenseNet121. From left to right: Diag Acc., Diag ECE, KFAC Acc. and KFAC ECE. Red crosses indicate configurations that where not invertible due to degeneracy or numerical instability. For accuracy, higher is better while for ECE lower is better.}
\label{fig:hyper}
\end{figure*}

\begin{figure*}[ht]
\minipage{0.25\textwidth}%
  \includegraphics[width=\linewidth]{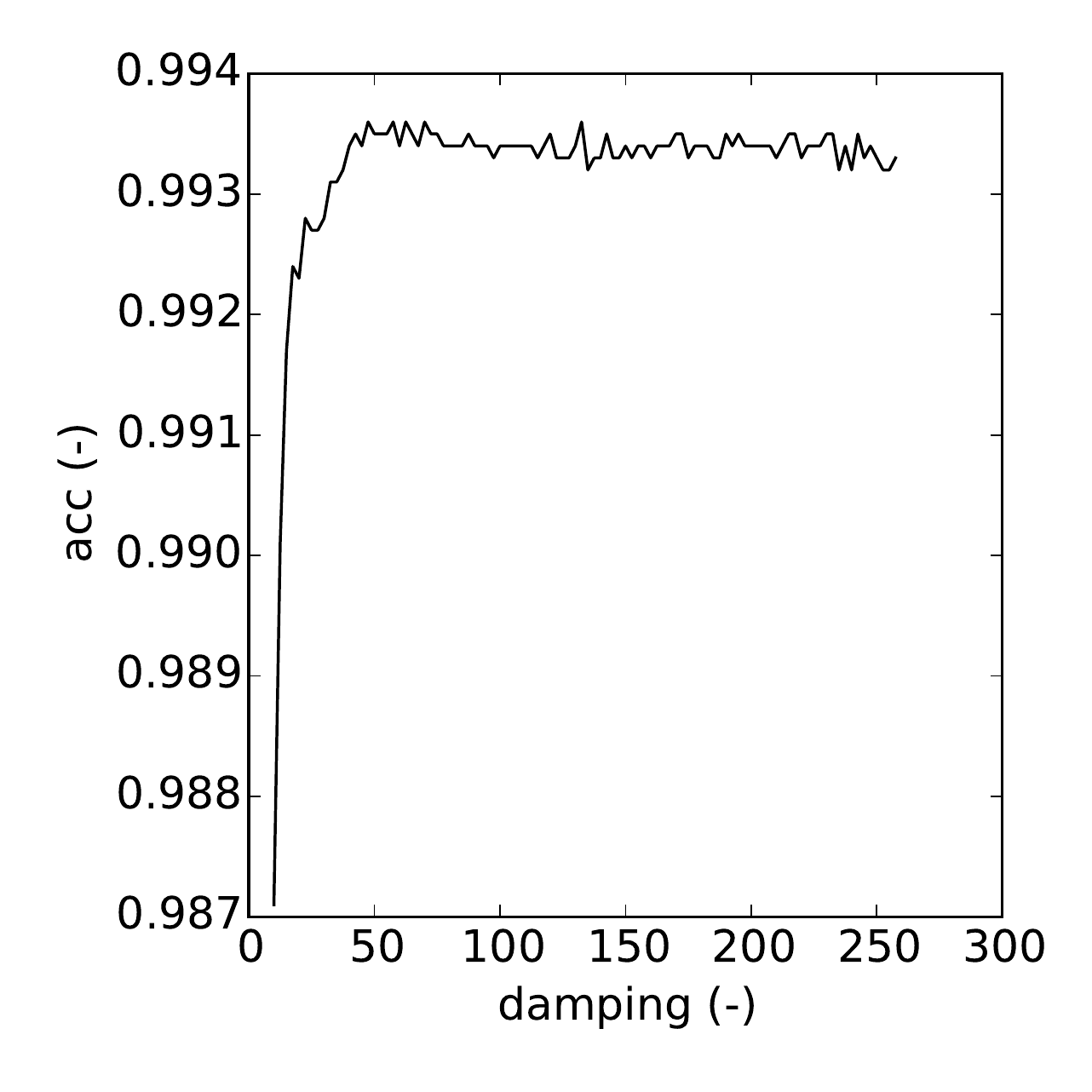}
\endminipage \hfill
\minipage{0.25\textwidth}
  \includegraphics[width=\linewidth]{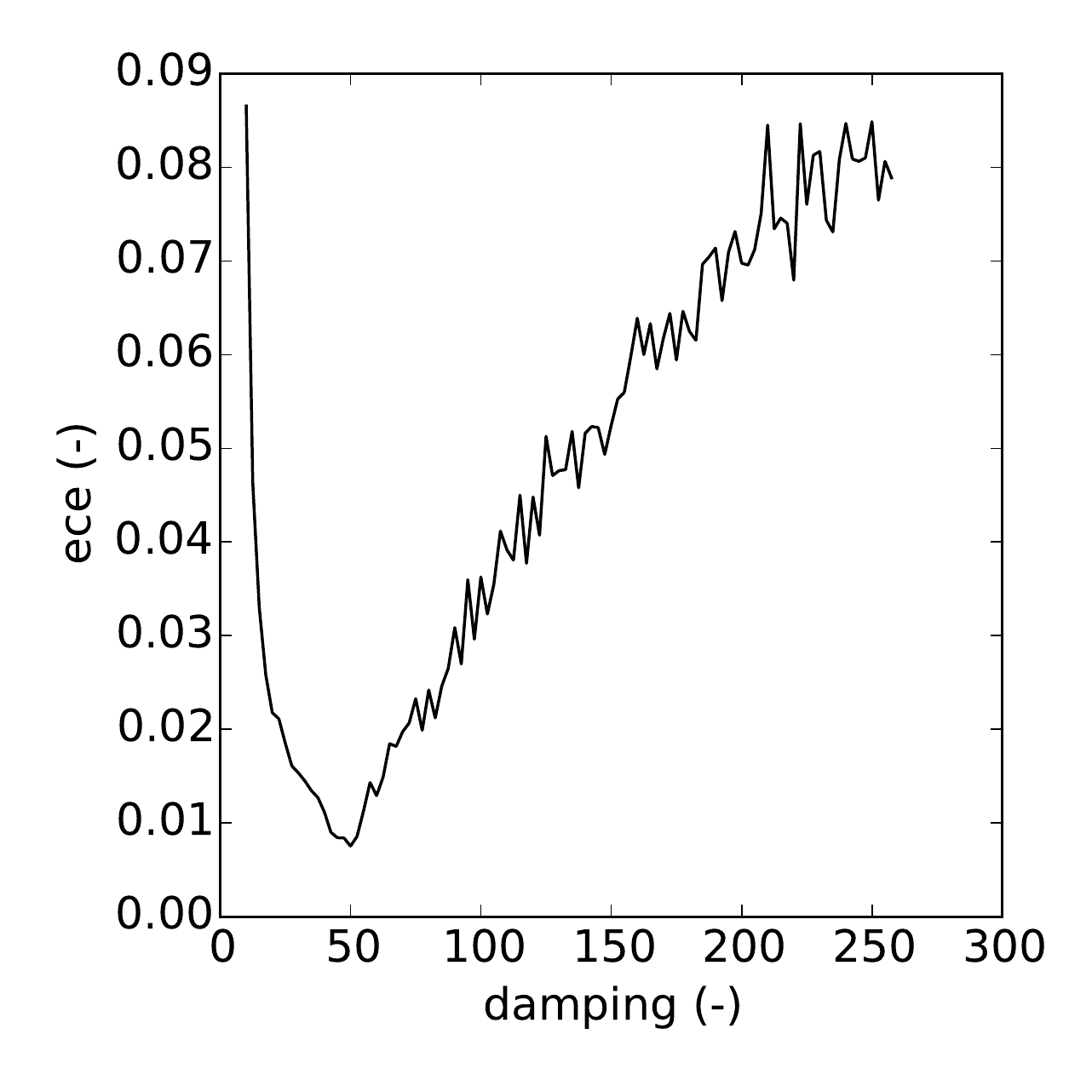}
\endminipage \hfill
\minipage{0.25\textwidth}
  \includegraphics[width=\linewidth]{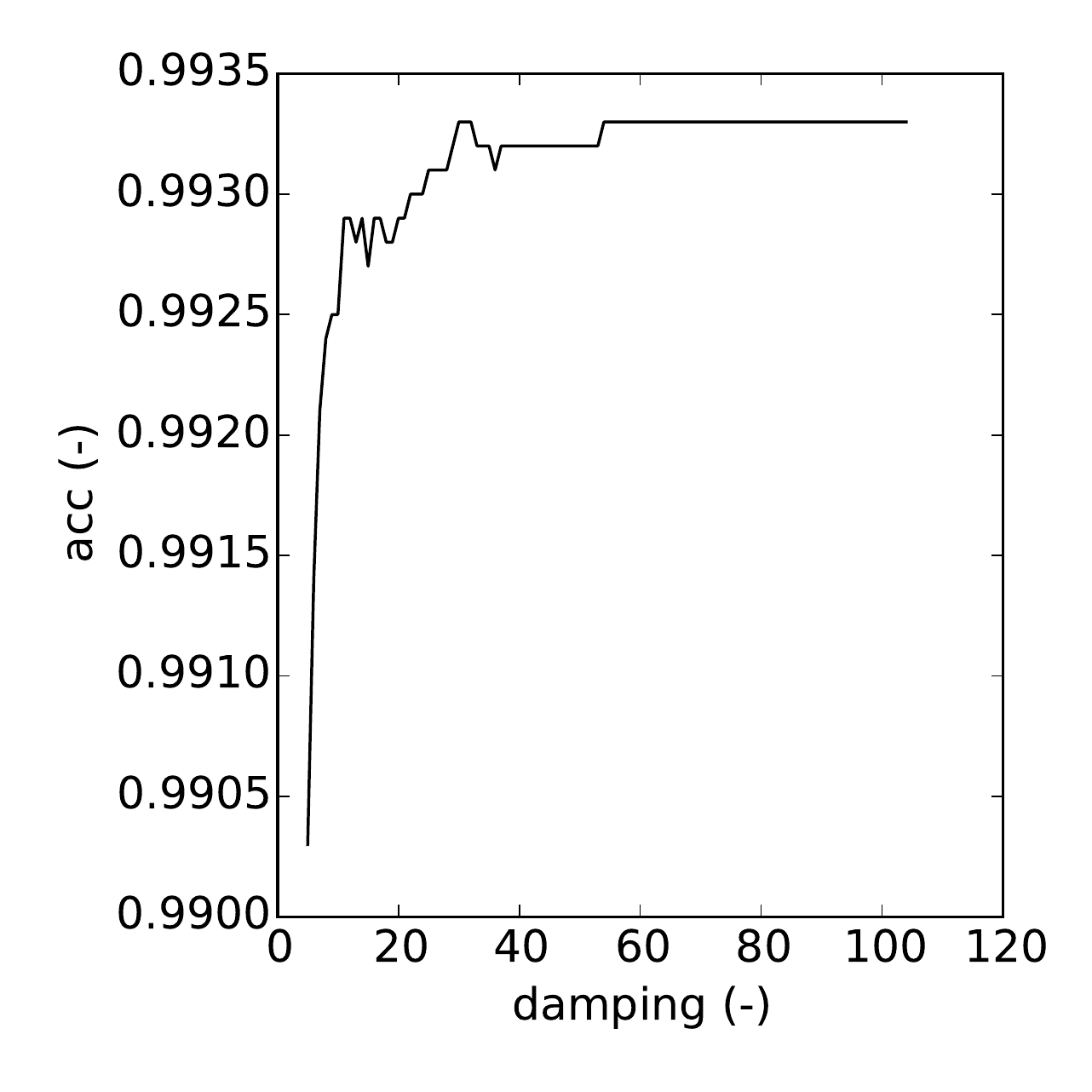}
\endminipage \hfill
\minipage{0.25\textwidth}%
  \includegraphics[width=\linewidth]{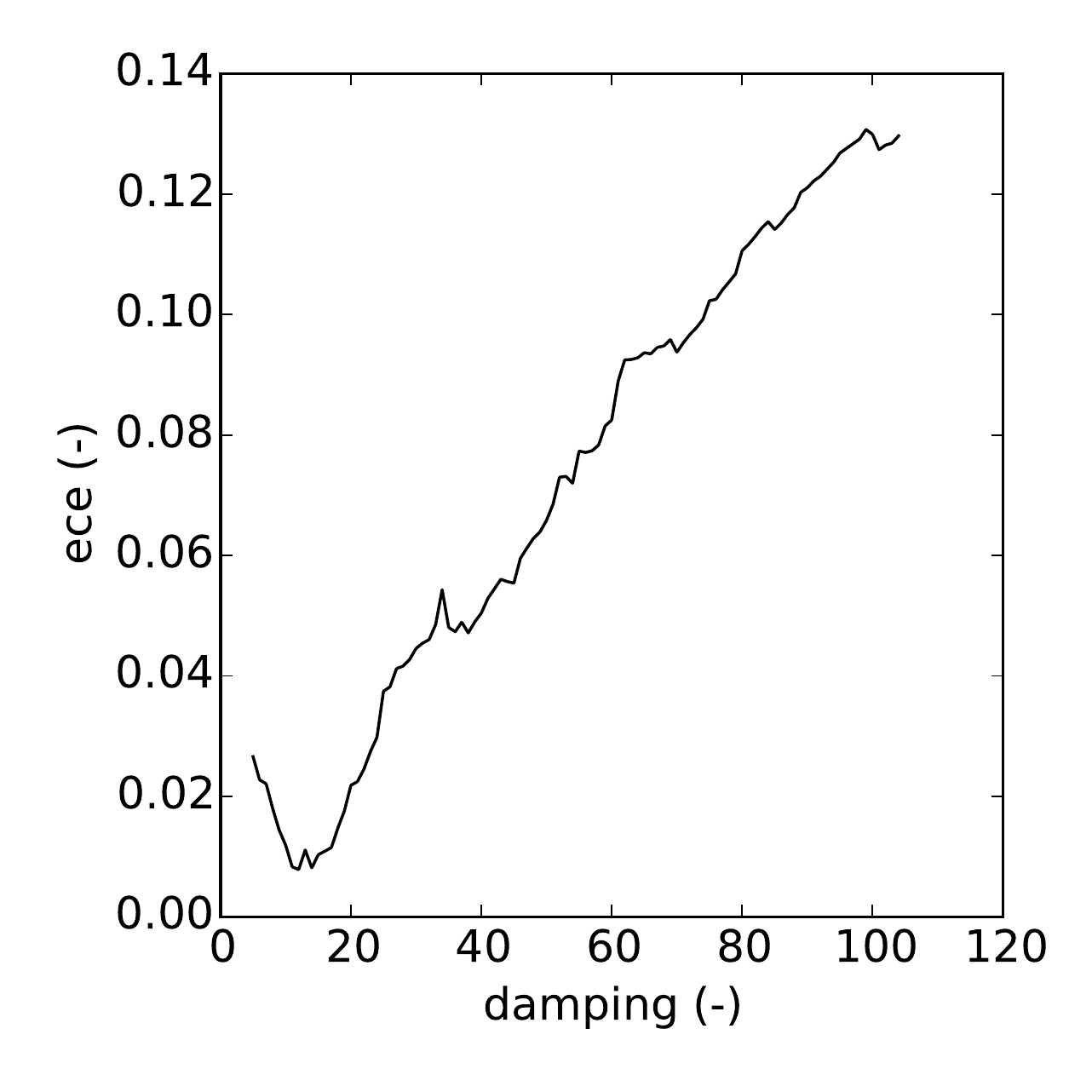}
\endminipage \hfill
\caption{\textbf{Grid search results.} For Diag (left two figures) and KFAC (right two) an extensive grid search has been conducted to ensure fair comparison. Here, we report the results with pseudo observation term of 50000 on MNIST. This ensures that a main difference to DEF Laplace is the expression for model uncertainty as the inference and network architectures are kept the same.}
\label{fig:grid:mnist}
\end{figure*}

\subsection{MNIST and CIFAR10 Experiments}
\begin{table}[ht]
\scriptsize
\centering
\caption{\textbf{Necessity of low rank approximation and reduction in complexity.} Reduced dimensions from N to the chosen rank L per layer are reported for both MNIST and CIFAR10 experiments. CNN stand for convolution while FC is for fully connected layers. The complexity of sampling $O(N^3)$ are reduced to $O(L^3)$. Employed strategy here was to keep the maximum for the rank K, which results in a seemingly arbitrary rank L.}
\label{tab:complexity}
\begin{tabular}{cccc}
\toprule
MNIST & Dim N [-]       & Dim L [-]    & Percent [$\%$]   \\
\midrule
\textit{CNN-1}     &  800   & 450    &  \textbf{56.25} \\ 
\textit{CNN-2}    &  51200    & 5185    &  \textbf{10.12}     \\
\textit{FC-1}     &  3211264    & 5625    & \textbf{0.18} \\ 
\textit{FC-2}     &  10240   & 4775    & \textbf{46.63}  \\ 
\toprule
CIFAR & Dim N [-]  & Dim L [-] & Percent [$\%$] \\ \hline
\textit{CNN-1}  &   4800     &  4800 & \textbf{100}  \\
\textit{CNN-2}  &      102400           &  2112  & \textbf{2.06} \\
\textit{FC-1}     &    884736     &  3980 &  \textbf{0.45} \\
\textit{FC-2}    &    73728   &  5499 &  \textbf{7.45}     \\
\textit{FC-3}     &   1920     &  1920 &    \textbf{100} \\ 
\bottomrule
\end{tabular}
\end{table}

For MNIST, the dropout layer is used to the FC layers with a rate of 0.6. An important information is the size of each layers. The first layer constitutes 32 filters with 5 by 5 kernel, followed by the second layer with 64 filters and 5 by 5 kernel. The first fully connected layer then constitutes 1024 units and the last one ends with 10 units. We note that, this validates our method on memory efficiency as the third layer has a large number of parameters, and its covariance, being quadratic in its size, cannot be stored in our utilized GPUs. For CIFAR10 experiments, the most relevant settings are: the first layer constitutes 5 by 5 kernel with 64 filters. This is then again followed by the same. Units of 384, 192, and 10 have been used for the fully connected layers. Lastly, random cropping, flipping, brightness changes and contrast are the applied data augmentations.

Implementation of deep ensemble \citep{lakshminarayanan2017simple} is kept rather simple by not using the adversarial training, but we combined 15 networks that were trained with different initialization. The same architecture and training procedure were used for all. For dropout, we have tried a grid search of dropout probabilities of 0.5 and 0.8, and have reported the best results. For the methods based on LA, we have performed grid search on hyperparameters N of (1, 50000, 100000) and 100 values of $\tau$ were tried using known class validation set. Note that for every method, and different datasets, each method required different values of $\tau I$ to give a reasonable accuracy. Figure \ref{fig:grid:mnist} depicts examples on MNIST where minimum ECE were selected.  
The LRA is imposed as a way to tackle the challenges of computational intractability. To empirically access the reduction in complexity, we depict the parameter and low rank dimensions N and L respectively in table \ref{tab:complexity}. As shown, LRA based sampling computations reduce the computational complexity significantly. Furthermore, this explains the necessity of LRA - certain layers (e.g. FC-1 of both MNIST and CIFAR experiments) are computationally intractable to store, infer and sample. 

\subsection{ImageNet Experiments}

To demonstrate that our method does not require changes in the training procedure, we used pre-trained weights from Pytorch \footnote{Available at \url{https://pytorch.org/docs/stable/torchvision/models.html}}. Results are discussed in section \ref{sm:sec:ImageNet}. SWAG and SWA are the available baselines which have been evaluated on the ImageNet dataset and implementations are officially open-sourced \footnote{Available at \url{https://github.com/wjmaddox/swa_gaussian}}. We closely followed the described experimental procedure. For the out-of-domain data we have used artistic impressions and paintings of landscapes and objects \footnote{Available at \url{https://www.kaggle.com/c/painter-by-numbers/data}}. Lastly, performing multiple forward passes for the entire validation set is still a computationally expensive task. We therefore chose $K_{mc} = 30$ during inference, which we empirically found sufficient for the convergence, similar to \citet{swag}.



We performed an extensive hyperparamter search for all LA methods using 100 randomly sampled pairs of $N$ and $\tau$ selected from a log-scale between 0 and 10. We resorted to random search instead of grid search, as it tends to yield stronger results with a smaller number of samples \citep{bergstra2012random}. The results for the accuracy (Acc.) and expected calibration error (ECE) are shown in figure \ref{fig:hyper}. The ECE can be extremely low in insufficiently regularized areas, because the accuracy is also very low there, which is why we show the results for both metrics.

%% file: supplementary/further_results.tex
\section{Further Results and Critical Analysis}
\label{sm:sec:6}

\subsection{Spectral Sparsity of Information Matrix}
\label{sm:sec:6:sparsity}
\begin{figure}
\minipage{0.24\textwidth}%
  \includegraphics[width=\linewidth]{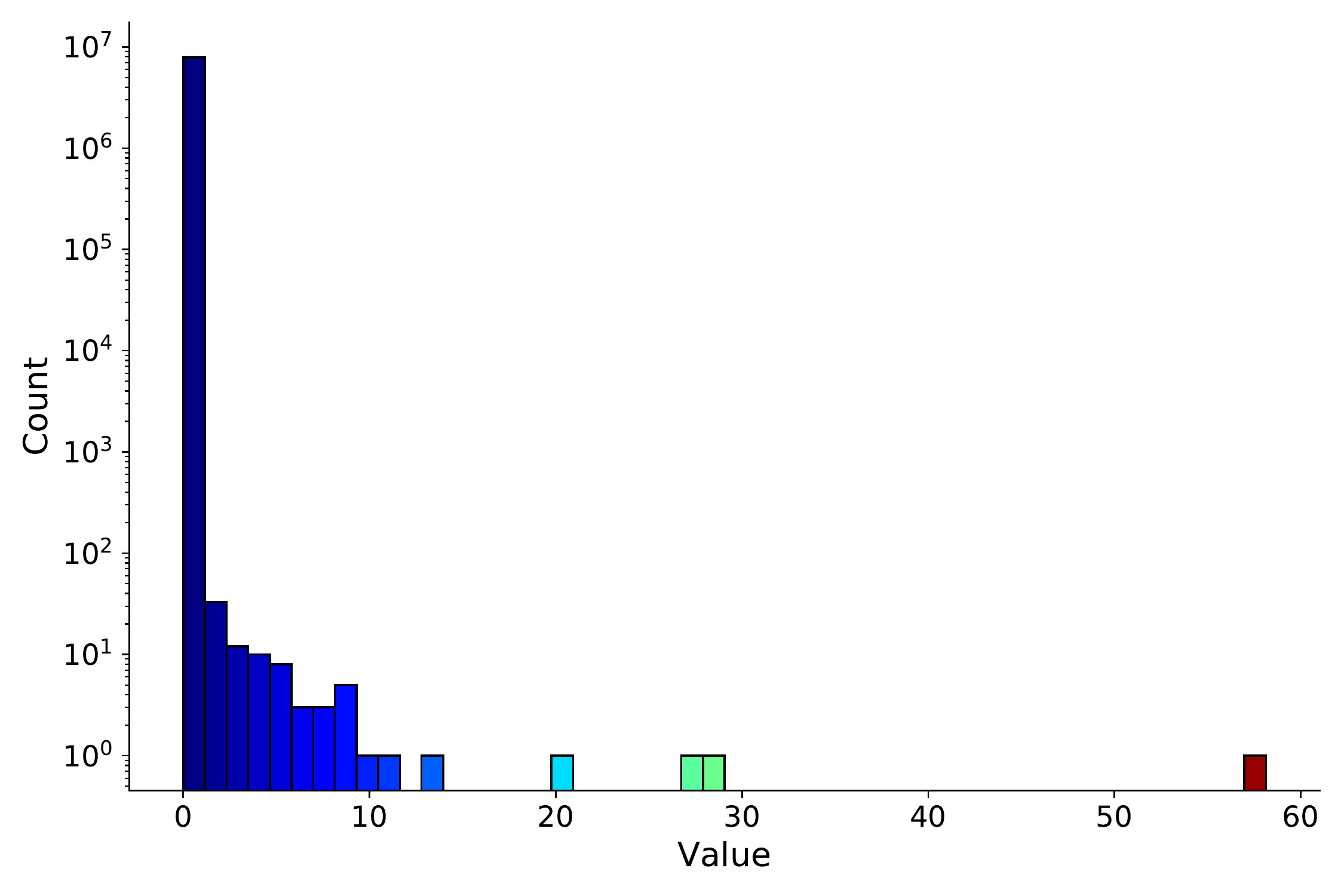}
  \caption{\small{Densenet121}}
  \label{eig:fig1}
\endminipage \hfill
\minipage{0.24\textwidth}%
  \includegraphics[width=\linewidth]{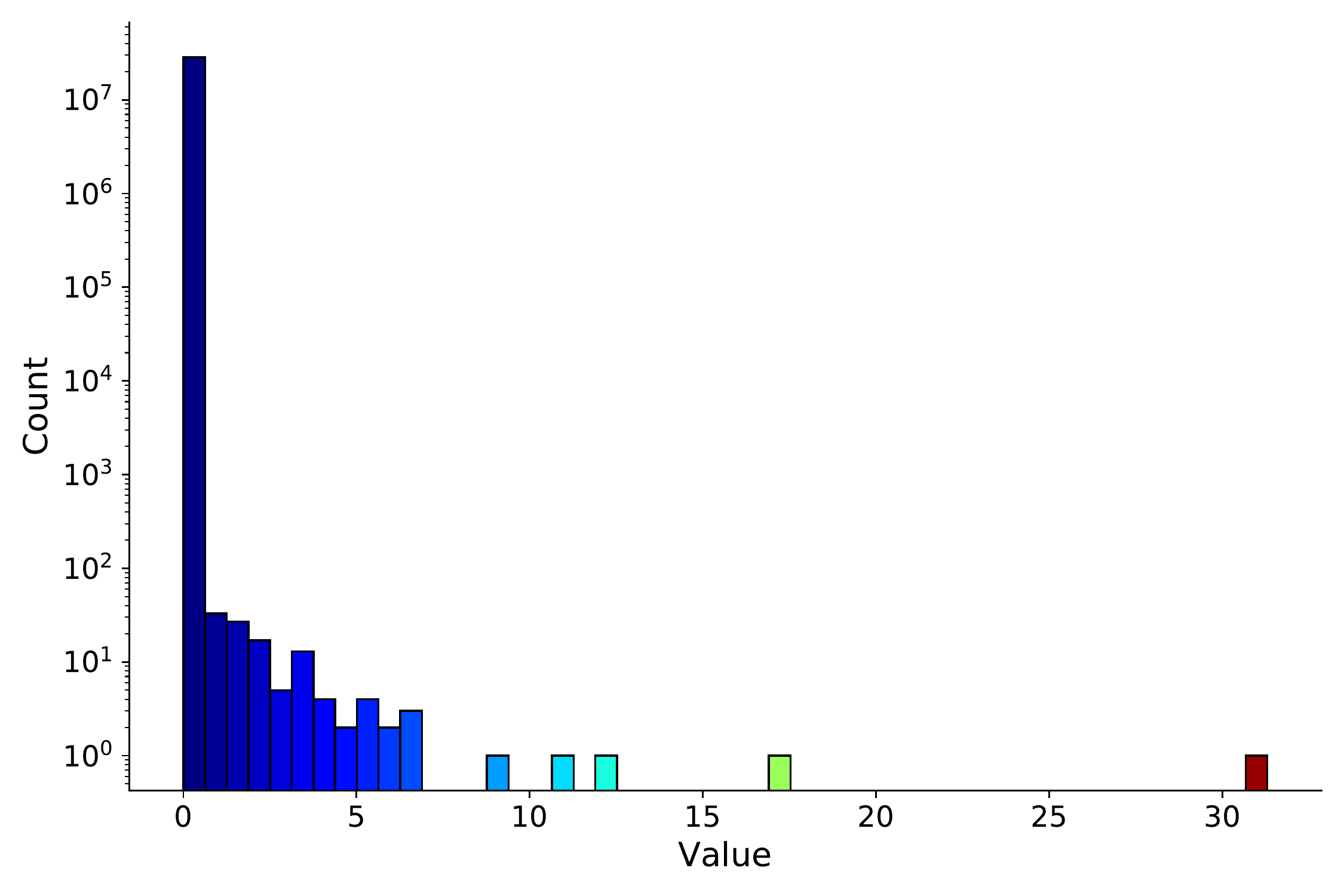}
  \caption{\small{Densenet161}}
  \label{eig:fig2}
\endminipage \hfill
\minipage{0.24\textwidth}%
  \includegraphics[width=\linewidth]{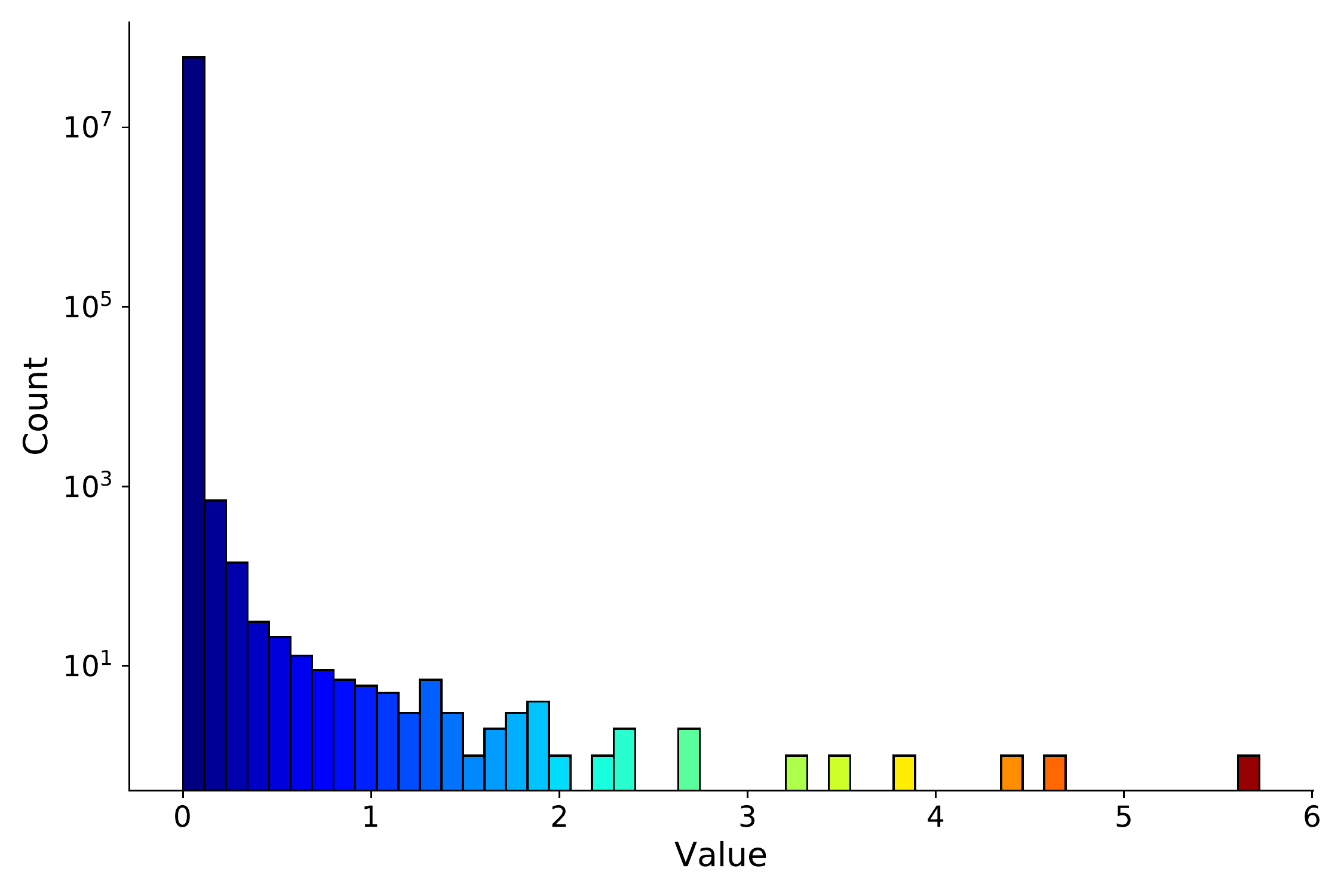}
  \caption{\small{Resnet152}}
  \label{eig:fig3}
\endminipage \hfill
\minipage{0.24\textwidth}%
  \includegraphics[width=\linewidth]{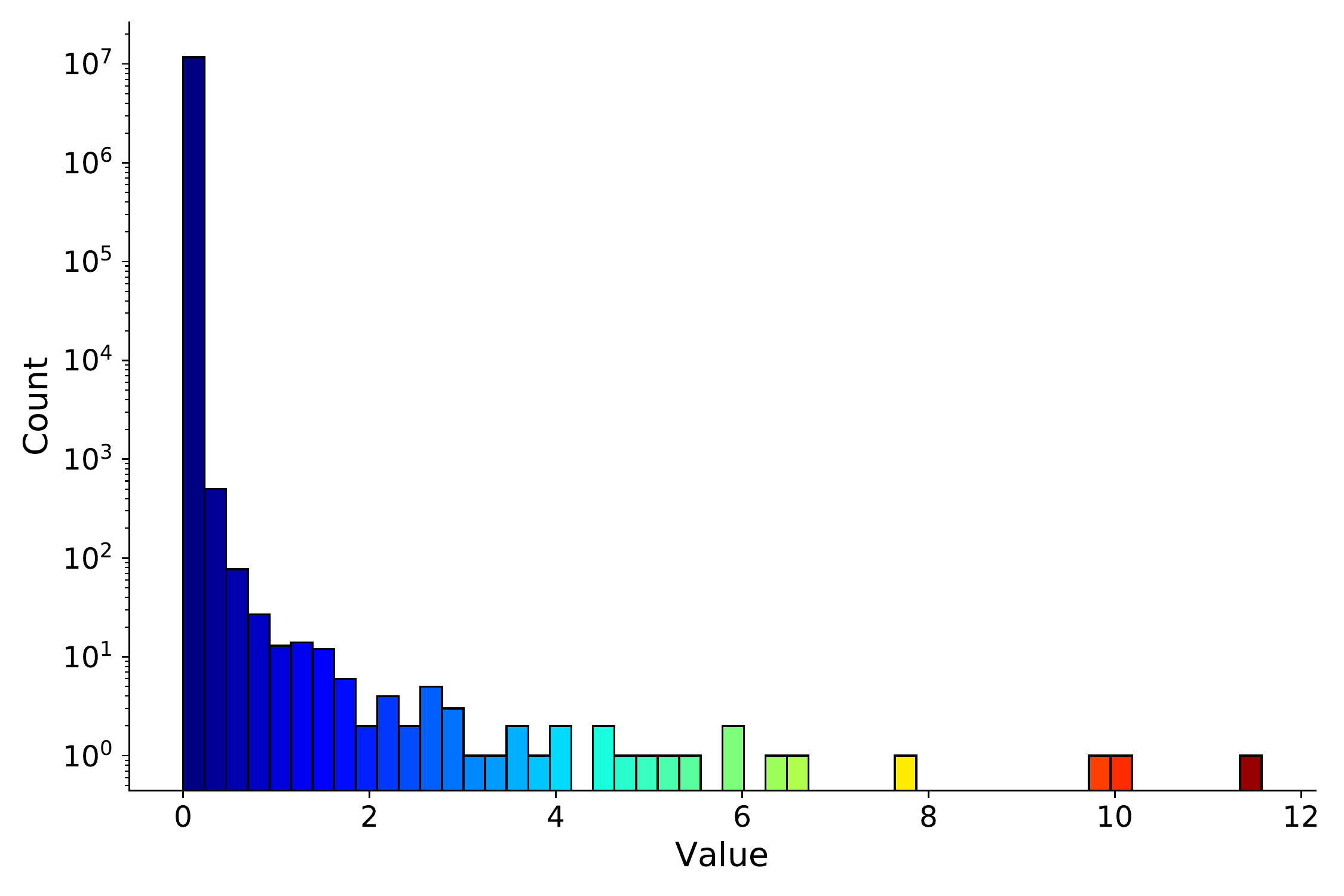}
  \caption{\small{Resnet18}}
  \label{eig:fig4}
\endminipage \hfill
\minipage{0.24\textwidth}%
  \includegraphics[width=\linewidth]{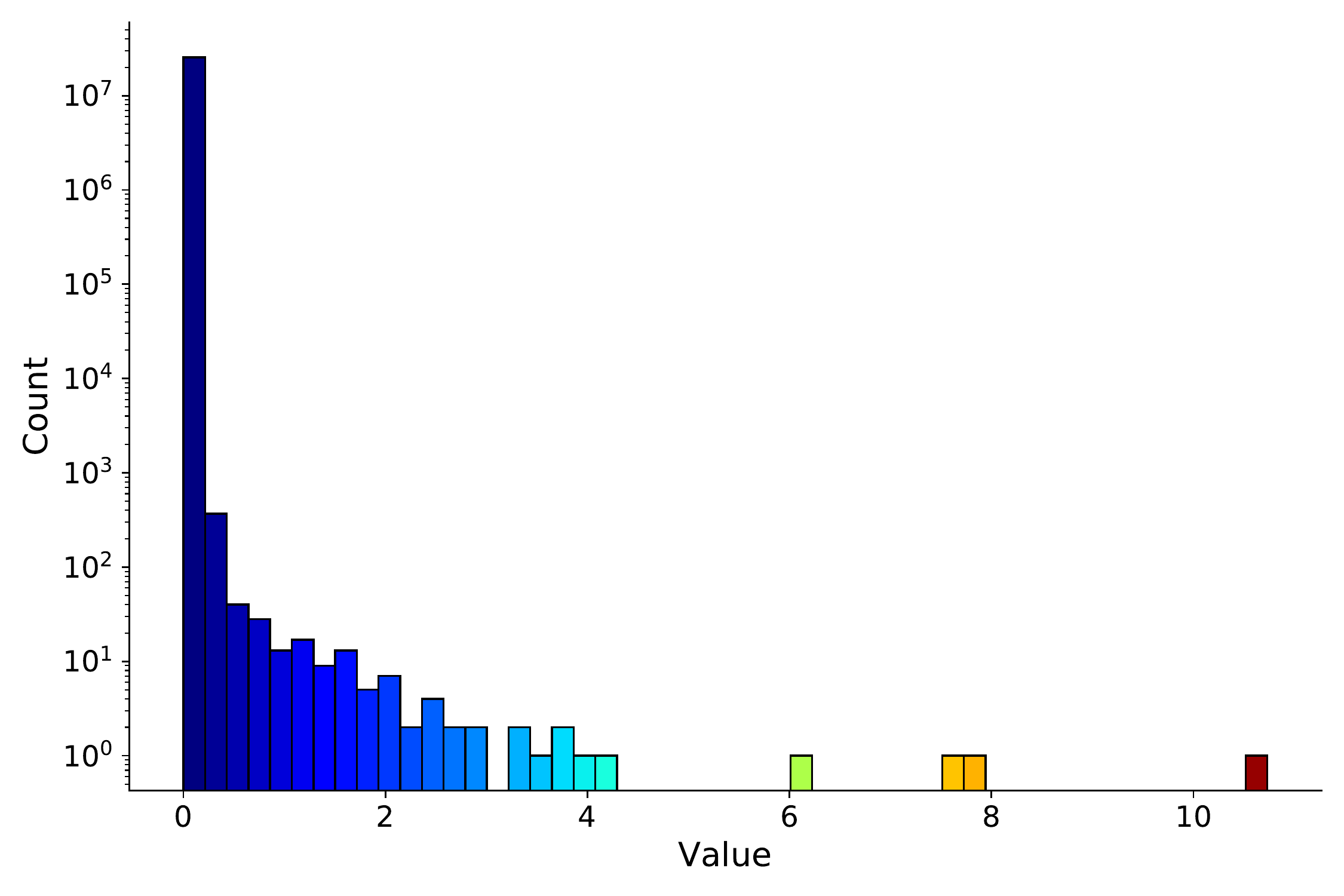}
  \caption{\small{Resnet50}}
  \label{eig:fig5}
\endminipage \hfill
\minipage{0.24\textwidth}%
  \includegraphics[width=\linewidth]{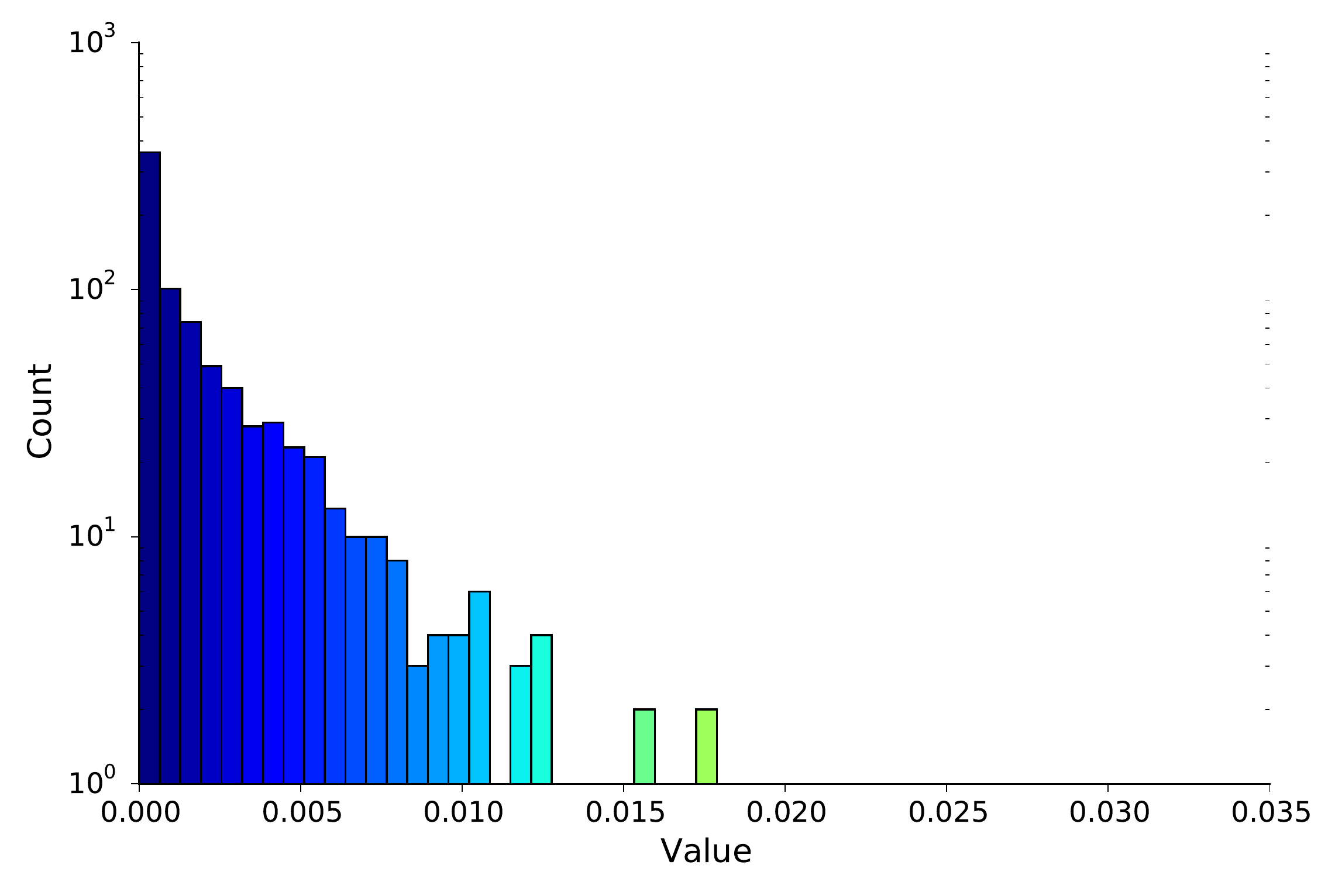}
  \caption{\small{MNIST (layer 1)}}
  \label{eig:fig6}
\endminipage \hfill
\minipage{0.24\textwidth}%
  \includegraphics[width=\linewidth]{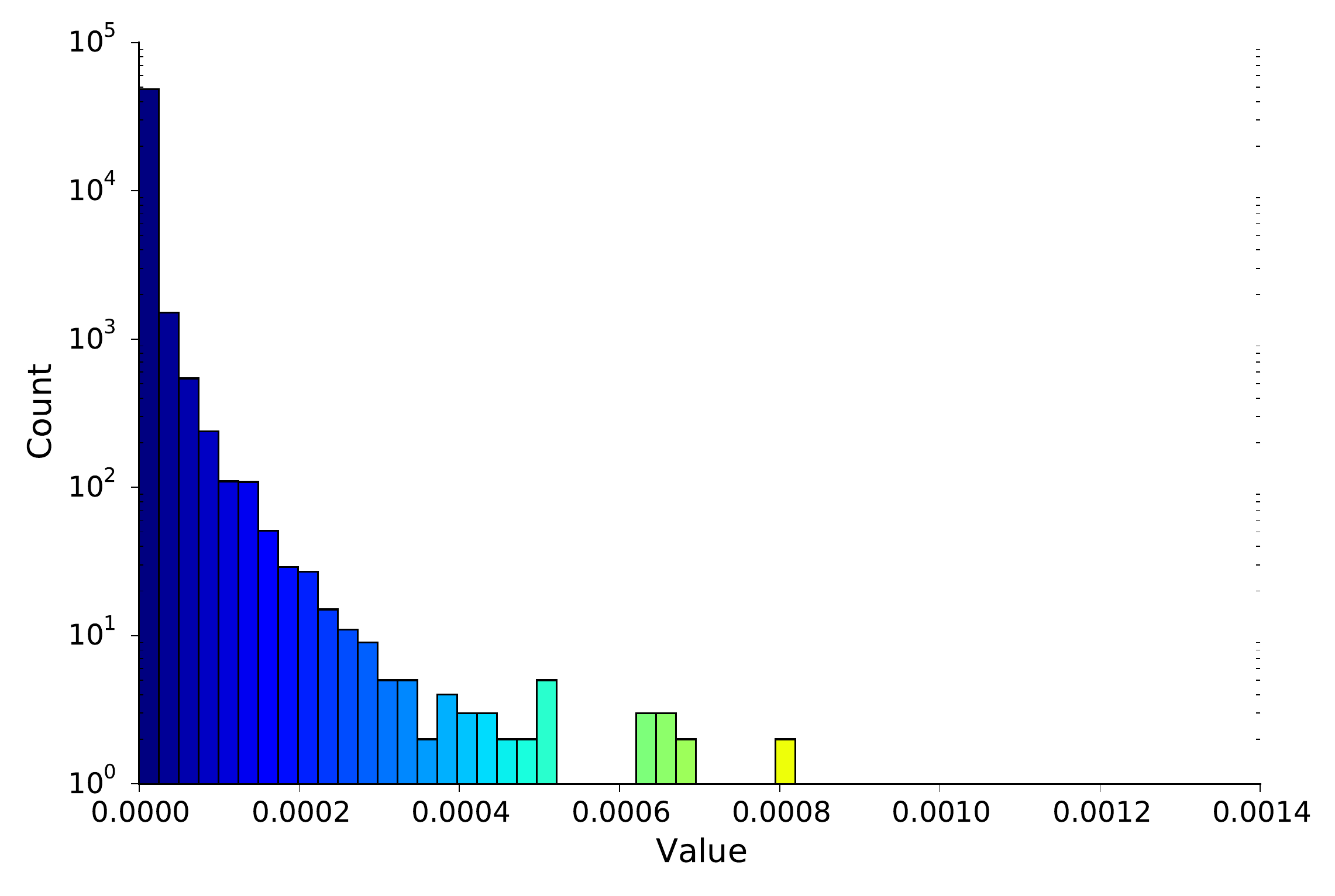}
  \caption{\small{MNIST (layer 2)}}
  \label{eig:fig7}
\endminipage \hfill
\minipage{0.24\textwidth}%
  \includegraphics[width=\linewidth]{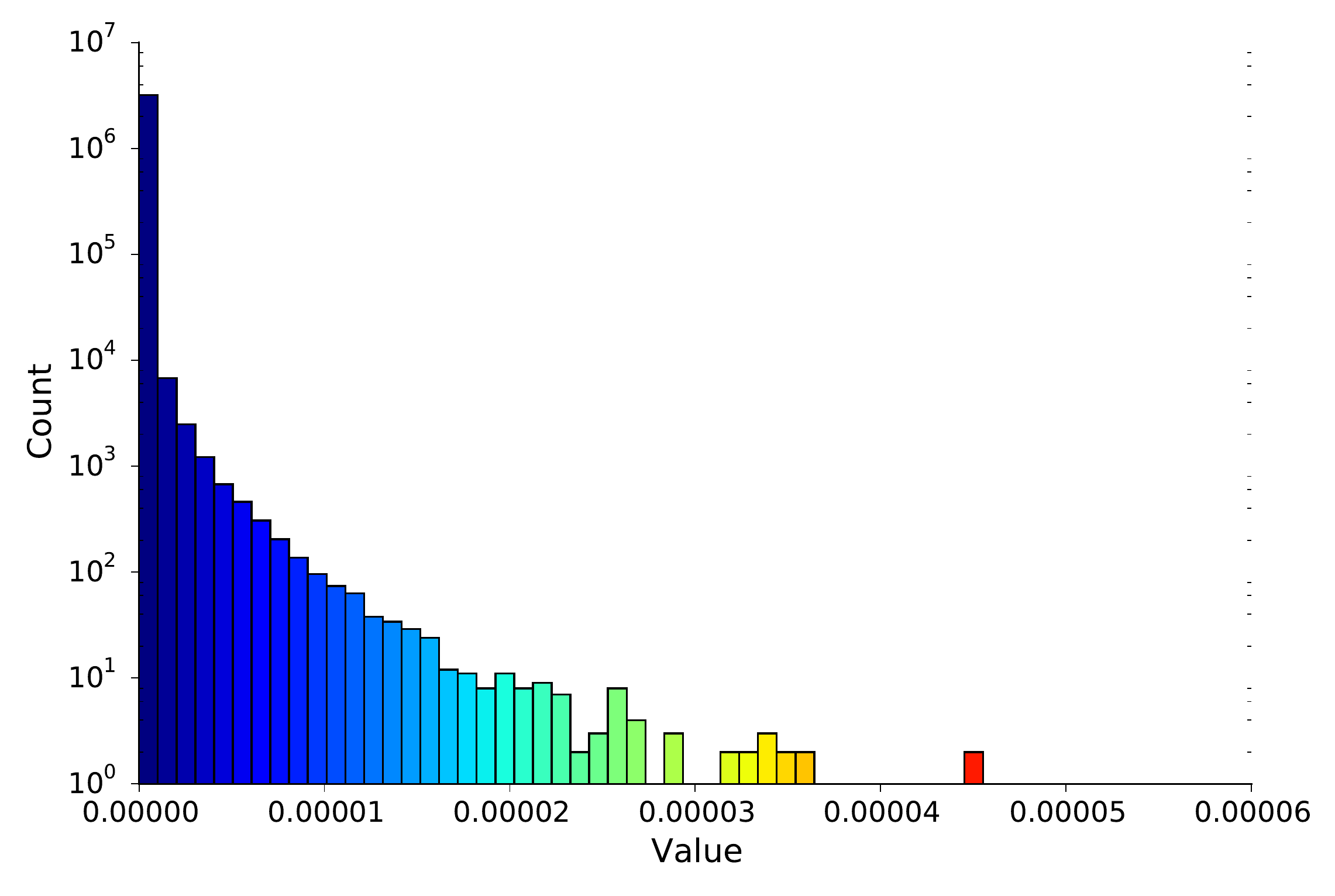}
  \caption{\small{MNIST (layer 3)}}
  \label{eig:fig8}
\endminipage \hfill
\minipage{0.24\textwidth}%
  \includegraphics[width=\linewidth]{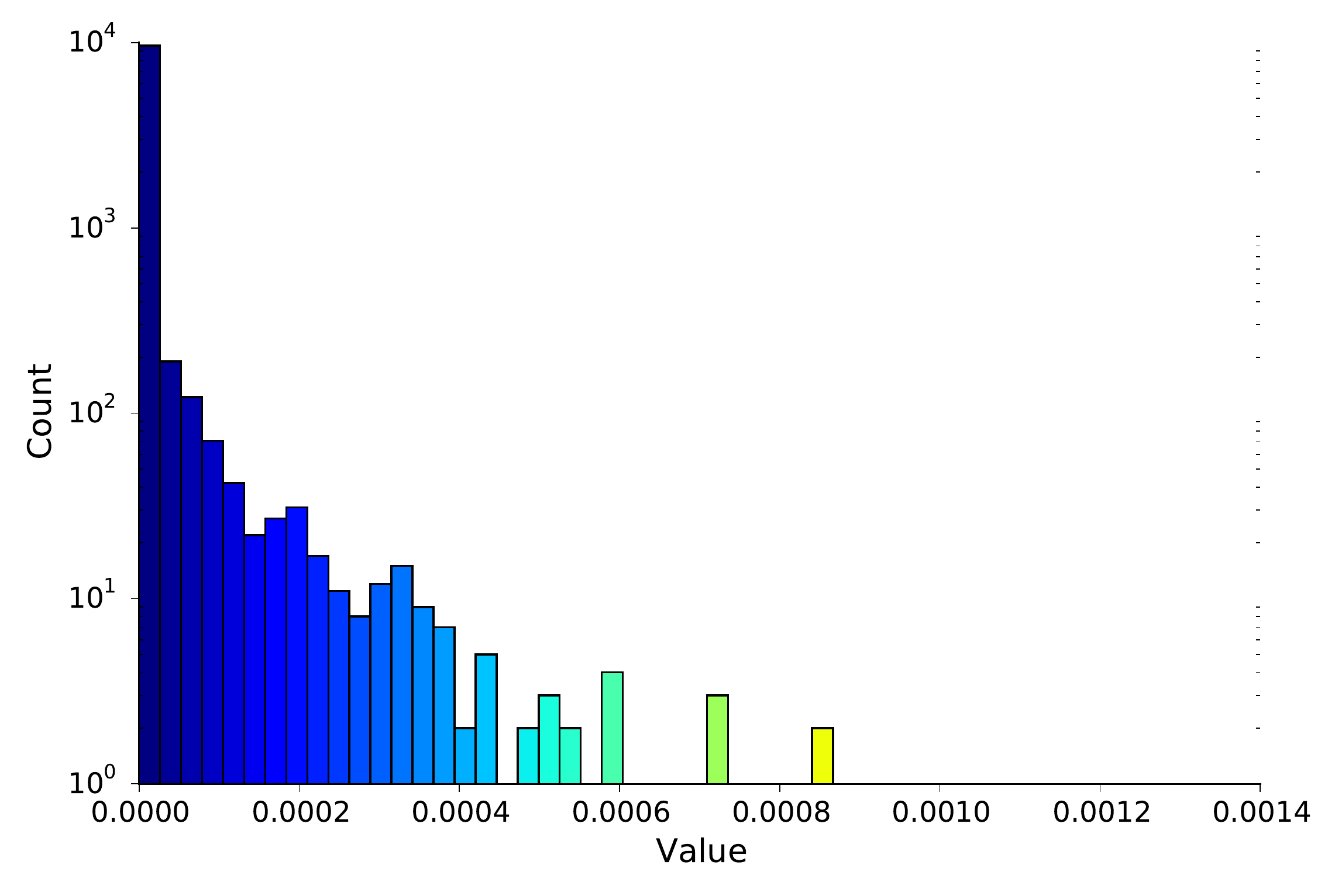}
  \caption{\small{MNIST (layer 4)}}
  \label{eig:fig9}
\endminipage \hfill
\caption{\textbf{Eigenvalue histogram.} For figures \ref{eig:fig1} to \ref{eig:fig5}, the eigenvalues of ImageNET architectures are shown. Here, x-axis plots the values whereas y-axis shows the counts in a log scale. From figure \ref{eig:fig6} to \ref{eig:fig9}, we show the eigenvalues of MNIST (layer-wise differentiated. These figures empirically shows that the spectrum of information matrix is sparse (tend to have many values close to zeros) for all the considered architectures. Furthermore, in the considered set-up for MNIST dataset, more overparameterized layer tends to have more close-to-zero eigenvalues even within the same architecture.}
\label{fig:eig_hist}
\end{figure}

One of the key insight behind our work is that information matrix of overparameterized DNNs tends to have close to zero eigenvalues (equivalently sparse in its spectrum). Is this true for the considered experiments? To answer this question, we plot the eigenvalue histograms in figure \ref{fig:eig_hist}. Figure \ref{fig:eig_hist} shows that the empirical findings of \citet{SagunEGDB18} hold well in our experiment set up. Two concrete observations are found: (i) with varying depth (figures \ref{eig:fig1}, \ref{eig:fig2}, \ref{eig:fig3}, \ref{eig:fig4}, \ref{eig:fig5}), IM showed tendency to get more sparse (especially on the maximum eigenvalue), and (ii) with varying number of parameters in each layers (figures \ref{eig:fig6}, \ref{eig:fig7}, \ref{eig:fig8}, \ref{eig:fig9}) IM showed to be more sparse with the number of parameters. One possible insight is that the information of individual parameters tends to be smaller if there are more parameters for explaining the same amount of data.

\subsection{Effects of Low Rank Approximation}
\label{sm:sec:6:frolra}
We additionally study the effects of LRA on the approximation quality of IM when compared to the exact, block diagonal IM. We include exact diagonal approximation to the true IM while decrease the ranks of INF in steps of $25\%$. The results are depicted in table \ref{results:sm:frolra} which shows that due to the sparsity of IM, the error (with a measure on normalized frobenius norm) does not drastically increase with lower ranks. Diagonal approximation also results in the most severe approximation error on off-diagonal elements. 

\subsection{Additional Results on Toy Regression Experiments}
\label{sm:sec:toy}
\begin{table}[ht]
\scriptsize
\centering
\caption{\textbf{UCI benchmark:} The normalized Frobenius norm of errors for the off-diagonal approximations w.r.t the true Fisher are depicted.}
\label{results:sm:frolra}
\begin{tabular}{ccccc}
\midrule
Dataset &   Off-diagonals      &    &     &  \\
       &  \textbf{Diag}       & \textbf{INF (75$\%$)}    &\textbf{INF (50$\%$)}   & \textbf{INF (25$\%$)} \\
\midrule
\textit{Boston}  &  1.000 $\pm$ 0.000  & 0.524$\pm$0.006  &  0.524$\pm$0.006  &  0.520$\pm$0.006 \\
\textit{Concrete}&  1.000 $\pm$ 0.000  & 0.506$\pm$0.008  &  0.506$\pm$0.008  &  0.508$\pm$0.008 \\
\textit{Energy}  &  1.000 $\pm$ 0.000  & 0.504$\pm$0.006  &  0.504$\pm$0.006  &  0.514$\pm$0.006 \\
\textit{Kin8nm}  &  1.000 $\pm$ 0.000  & 0.526$\pm$0.005  &  0.526$\pm$0.005  &  0.546$\pm$0.005 \\
\textit{Naval}   &  1.000 $\pm$ 0.000  & 0.465$\pm$0.003  &  0.465$\pm$0.003  &  0.465$\pm$0.003 \\
\textit{Power}   &  1.000 $\pm$ 0.000  & 0.492$\pm$0.008  &  0.492$\pm$0.008  &  0.502$\pm$0.008 \\
\textit{Protein} &  1.000 $\pm$ 0.000  & 0.541$\pm$0.021  &  0.541$\pm$0.021  &  0.541$\pm$0.021 \\
\textit{Wine}    &  1.000 $\pm$ 0.000  & 0.535$\pm$0.009  &  0.535$\pm$0.009  &  0.546$\pm$0.009 \\
\textit{Yacht}   &  1.000 $\pm$ 0.000  & 0.516$\pm$0.007  &  0.516$\pm$0.007  &  0.526$\pm$0.007 \\
\bottomrule
\end{tabular}
\end{table}

\begin{figure}
\minipage{0.23\textwidth}%
  \includegraphics[width=\linewidth]{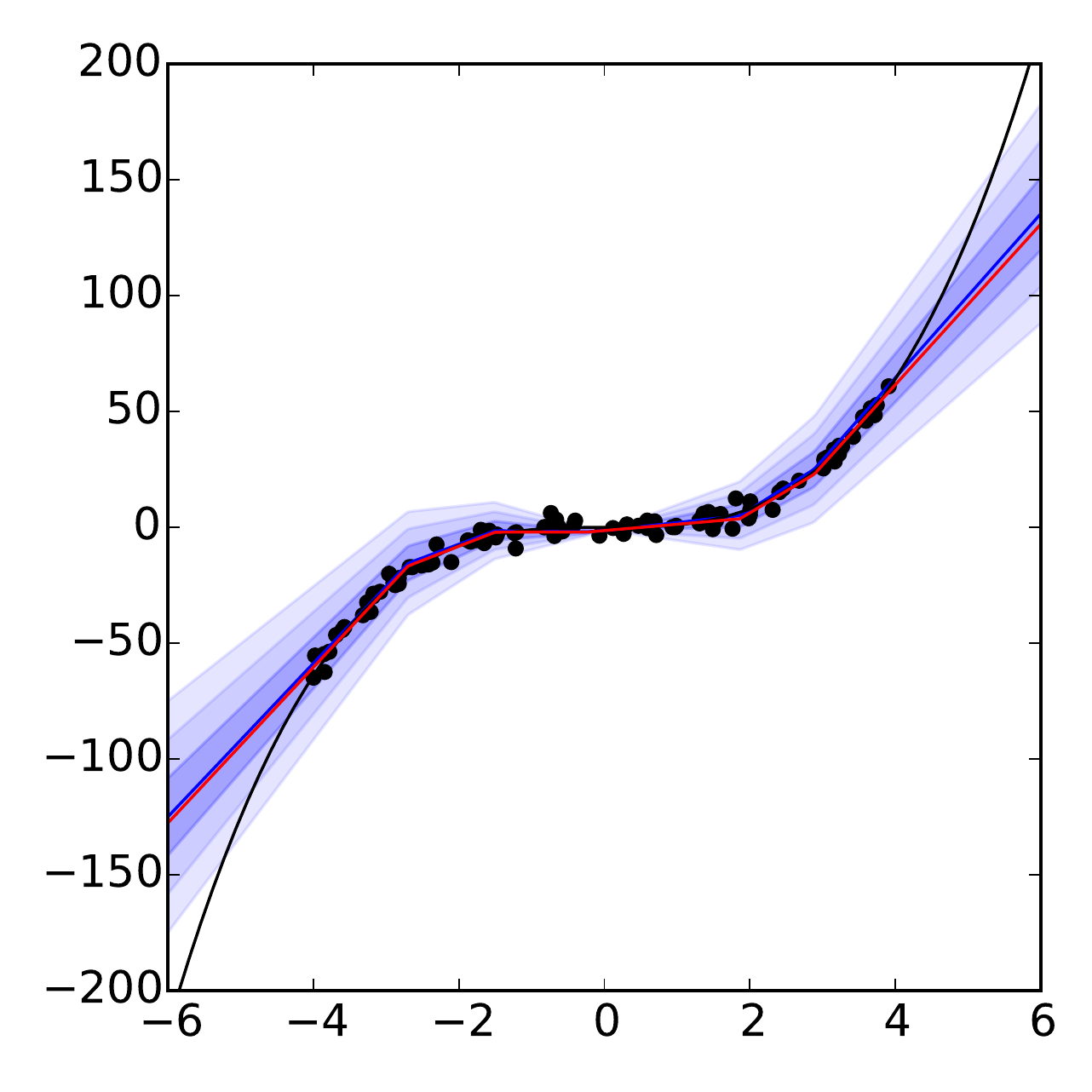}
  \caption{\small{Diag}}
\endminipage \hfill
\minipage{0.23\textwidth}
  \includegraphics[width=\linewidth]{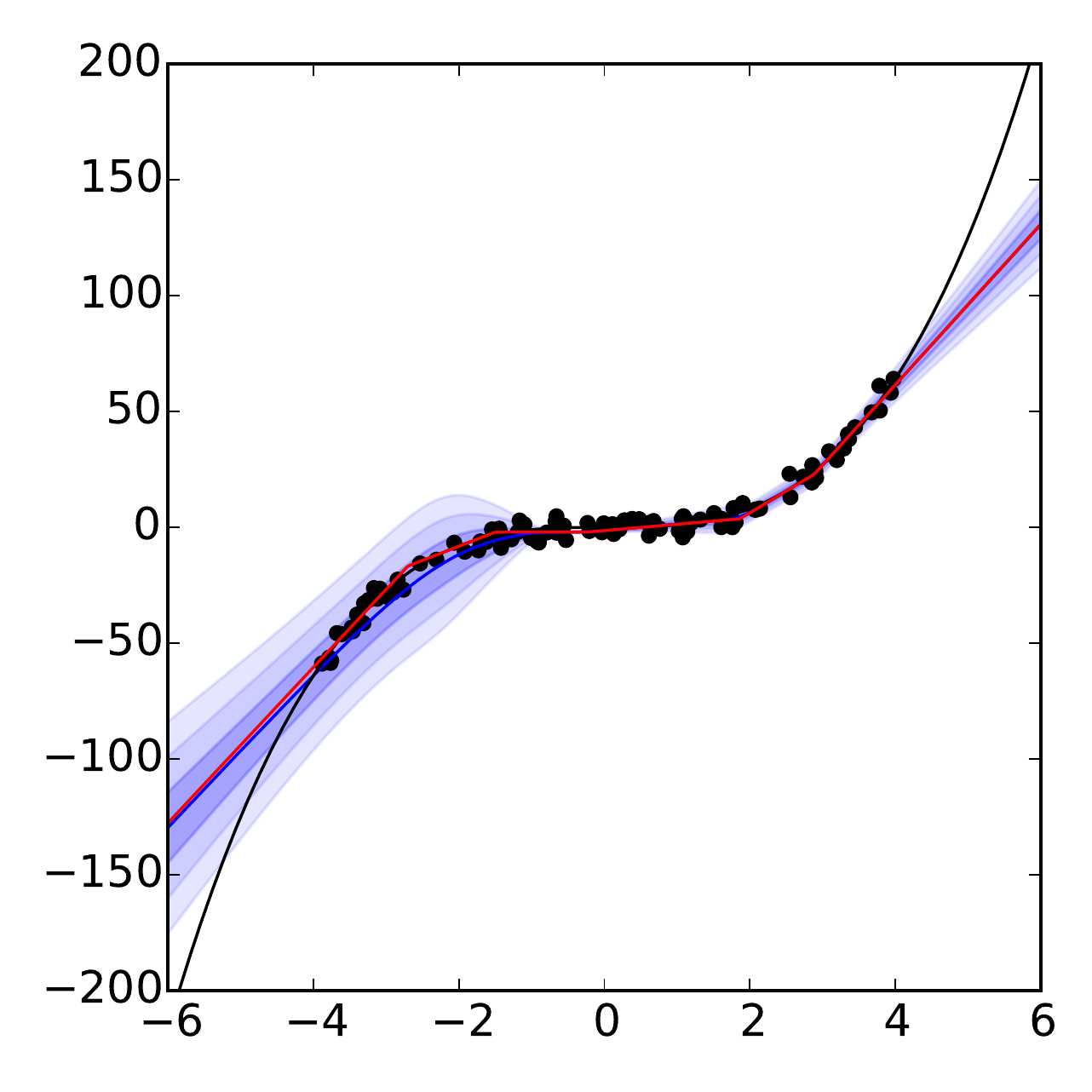}
  \caption{\small{EFB}}
\endminipage \hfill
\minipage{0.23\textwidth}
  \includegraphics[width=\linewidth]{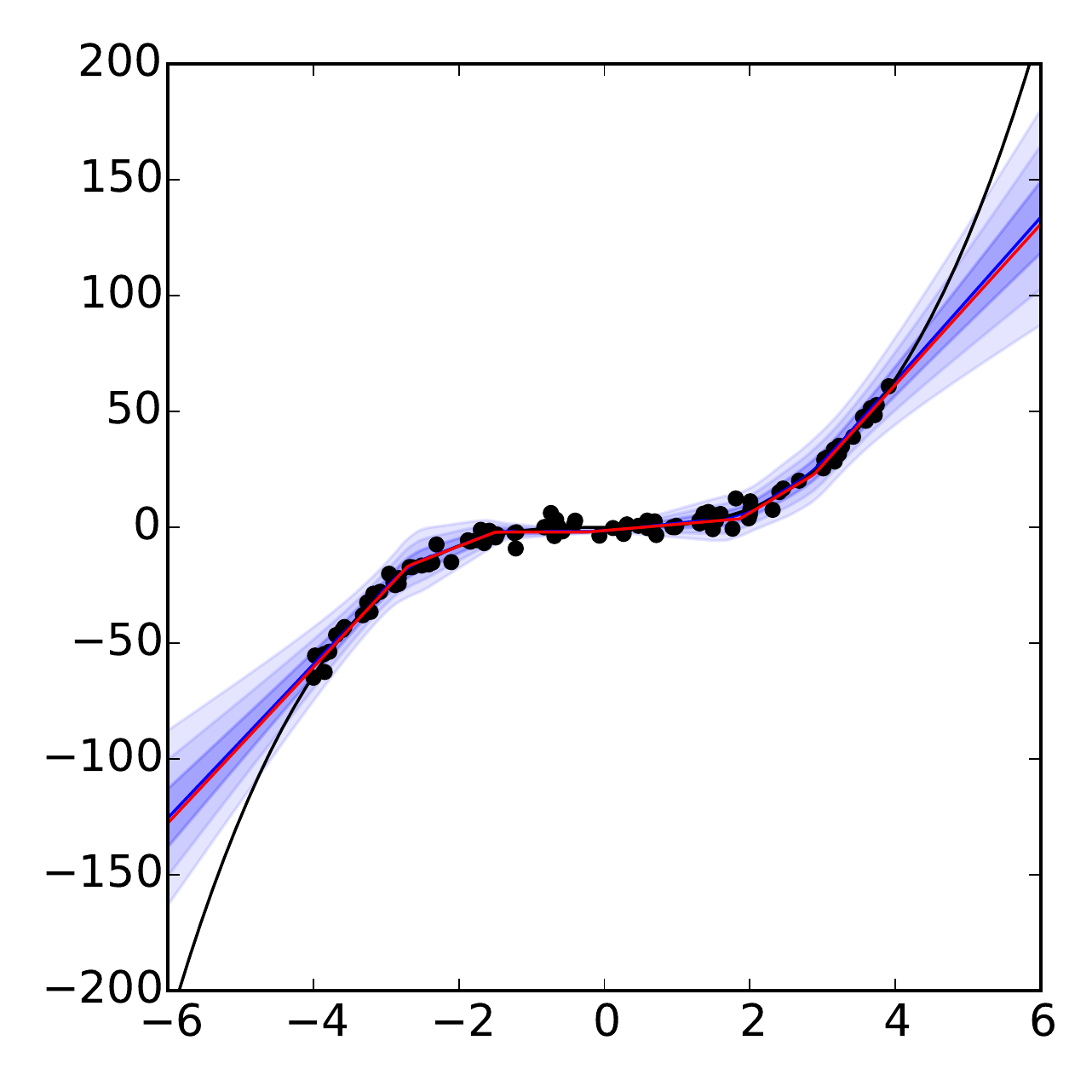}
  \caption{\small{FB}}
\endminipage \hfill
\minipage{0.23\textwidth}%
  \includegraphics[width=\linewidth]{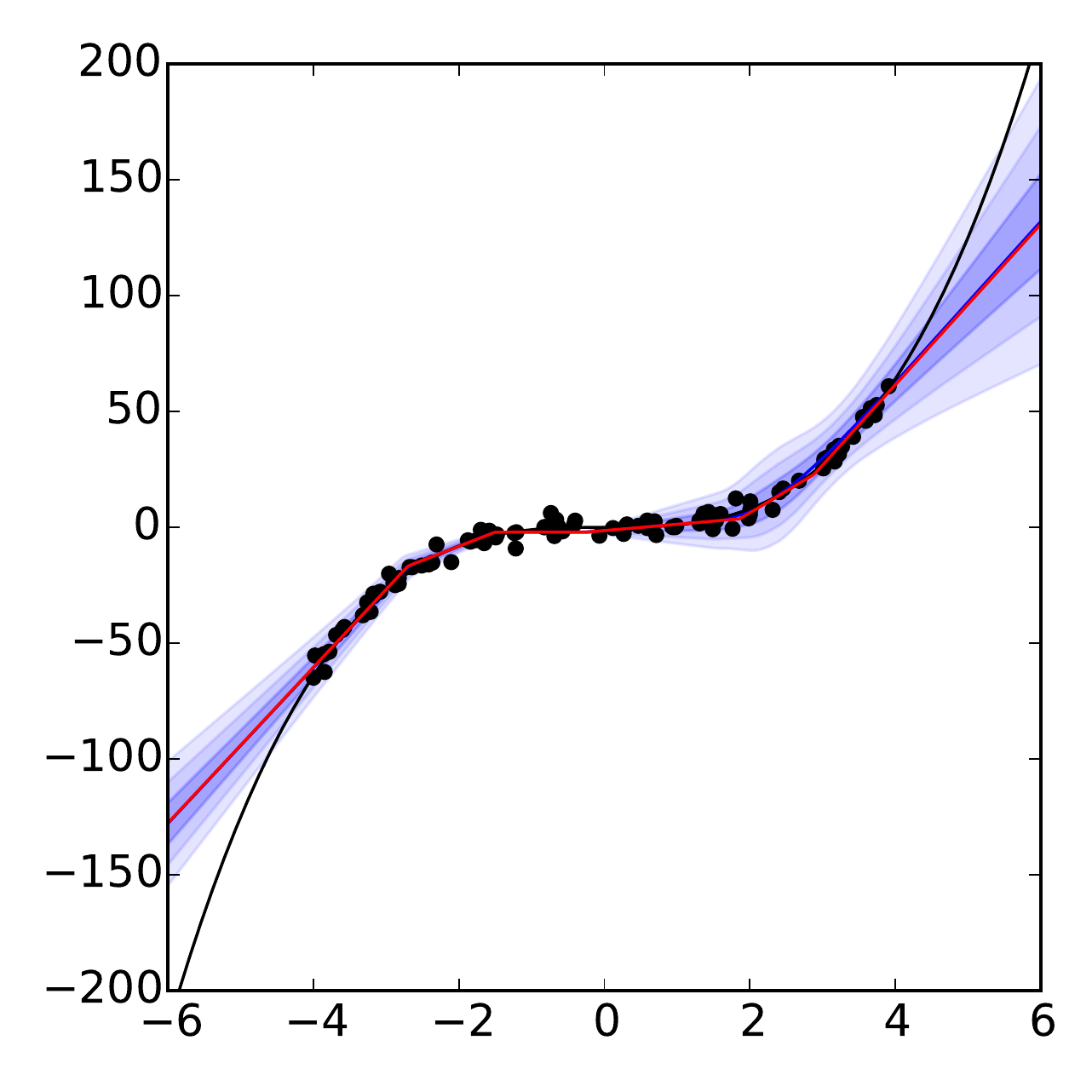}
  \caption{\small{INF with rank 1}}
\endminipage \hfill
\caption{\textbf{Uncertainty on toy regression.} The black dots and the black lines are data points (x, y). The red and blue lines show predictions of the deterministic Neural Network and the mean output respectively. Upto three standard deviations are shown with blue shades.}
\label{fig:toy:additional}
\end{figure}

\begin{figure}
\minipage{0.22\textwidth}%
  \includegraphics[width=\linewidth]{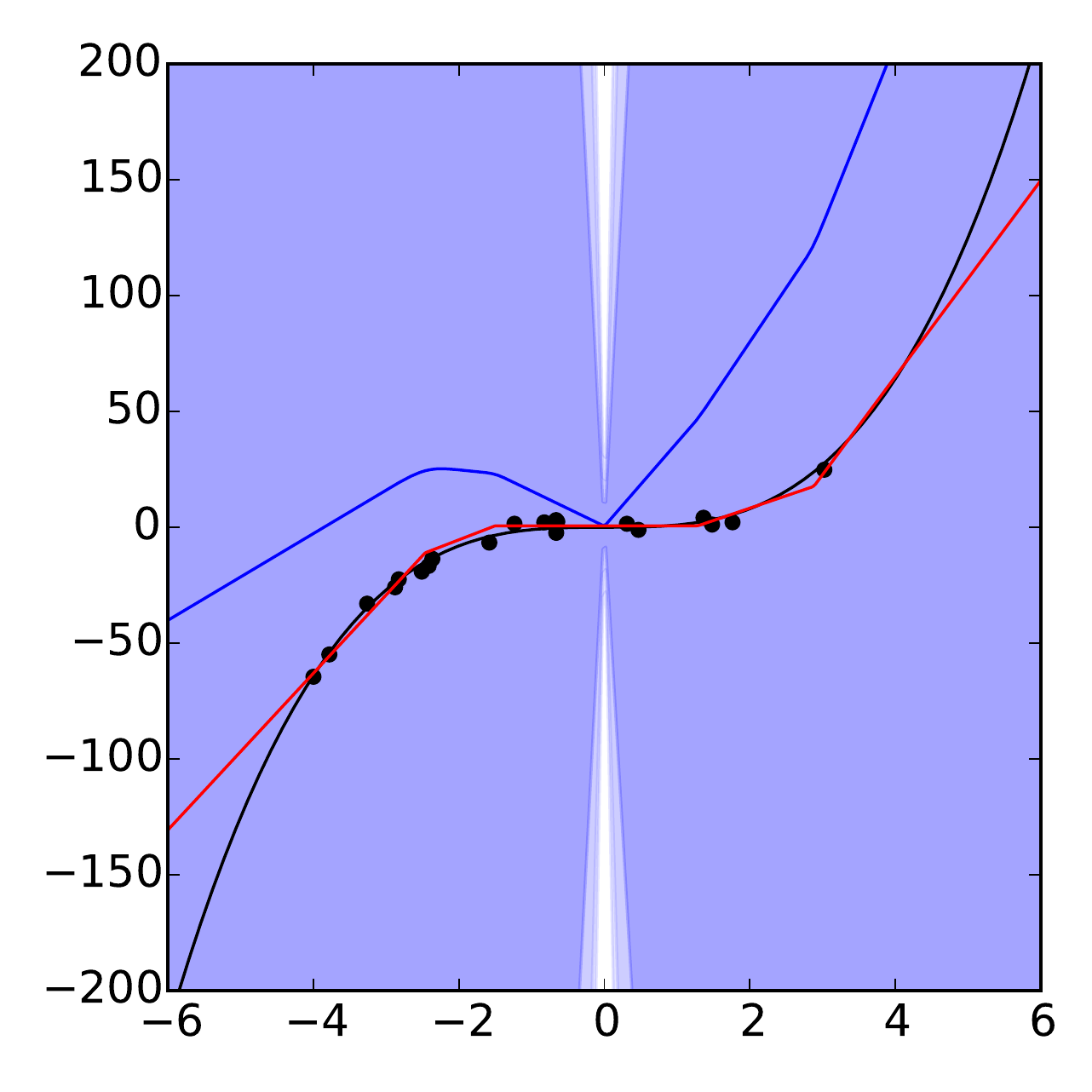}
  \caption{\small{OKF}}
\endminipage \hfill
\minipage{0.22\textwidth}
  \includegraphics[width=\linewidth]{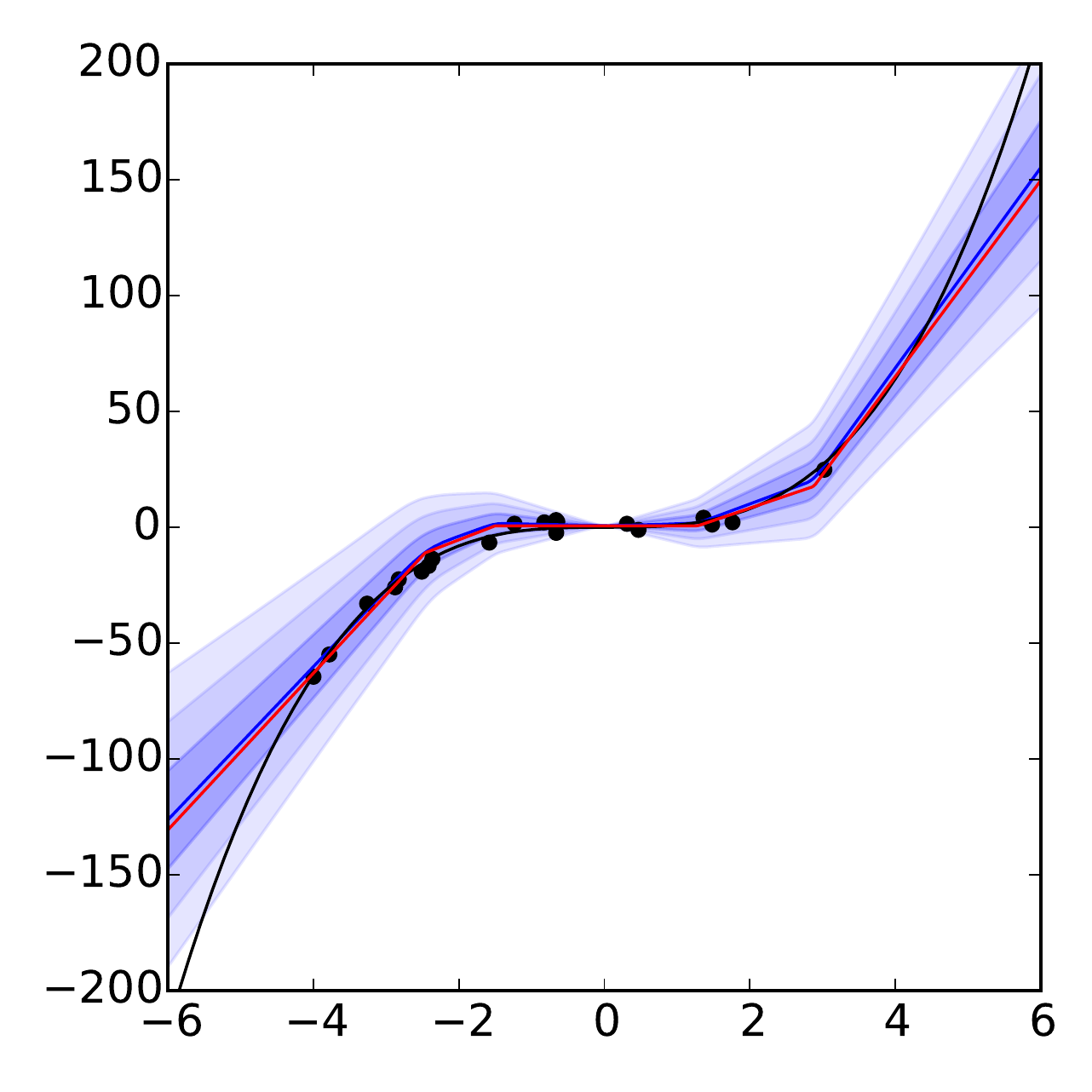}
  \caption{\small{KFAC}}
\endminipage \hfill
\minipage{0.22\textwidth}%
  \includegraphics[width=\linewidth]{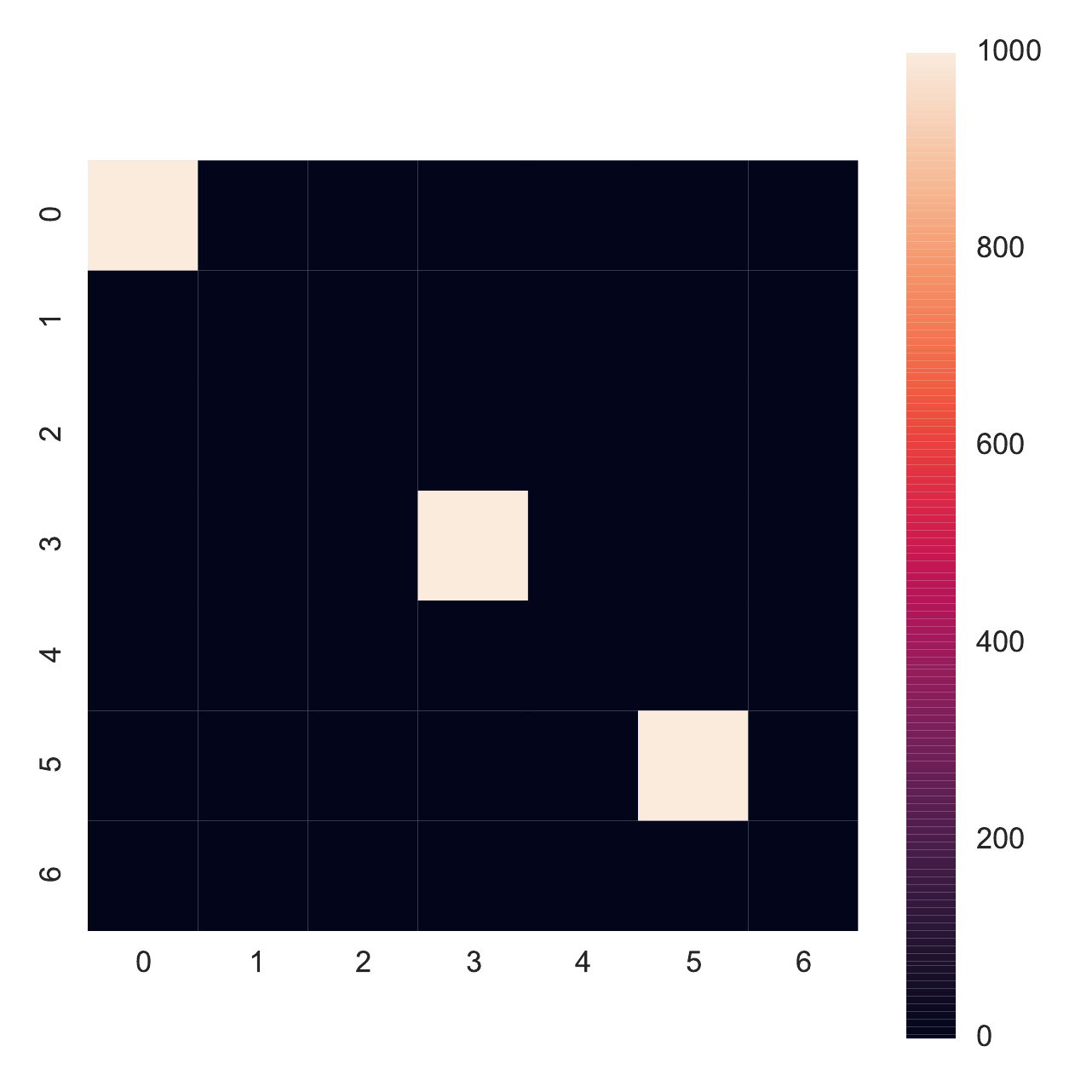}
  \caption{\small{OKF $\Sigma$}}
\endminipage \hfill
\minipage{0.22\textwidth}
  \includegraphics[width=\linewidth]{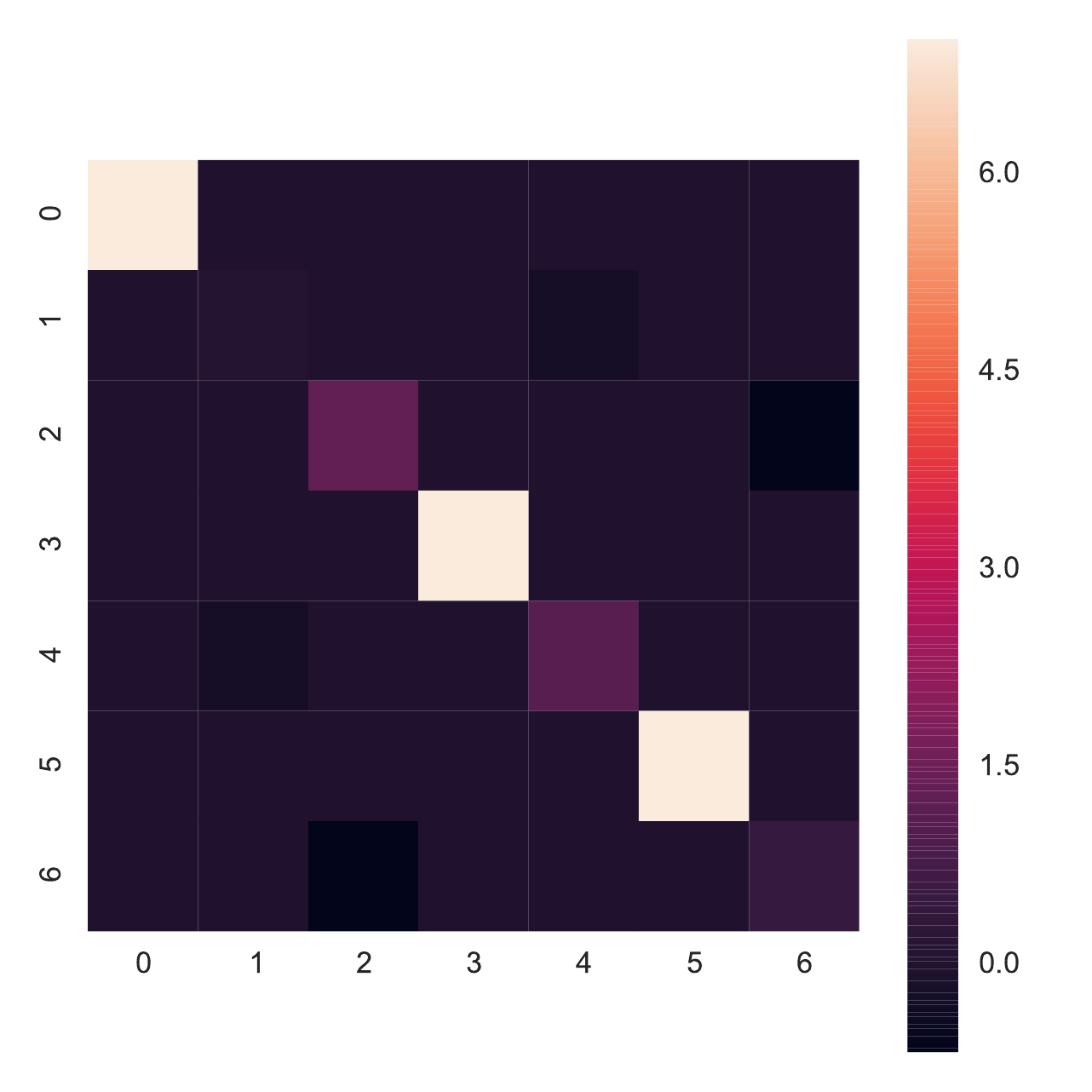}
  \caption{\small{KFAC $\Sigma$}}
\endminipage \hfill
\caption{\textbf{Toy regression uncertainty and covariance visualization} (only the first layer is shown here). OKF Laplace means using equation \ref{sm:sec6:eq:kfacproperprior} without further approximation in equation \ref{sm:sec6:eq:kfacprior}.}
\label{fig:1:sm:4:2}
\end{figure}

\subsubsection{Effects of diagonal correction.}
What is the relation between keeping diagonal elements of IM exact and predictive uncertainty? As effects of regularizing hyperparameters are removed to certain extent in the toy experiment, we study above mentioned question within this limited but controllable set-up.

For this purpose, we depict Diag, EFB, FB (layer-wise true IM) and INF with rank 1. The most comparable fit to HMC is given by FB while INF with rank 1 deteriorates when compared to its full rank counterpart. EFB for this set-up, produces considerable misfit to HMC. Importantly, since the only difference between EFB and DEF Laplace is a diagonal correction term, these results suggest that keeping diagonals of IM exact can result in accurate predictive uncertainty.

\subsubsection{KFAC - a critical analysis.}
\citet{Ritter2017ASL, RitterBB18} reports that KFAC requires smaller sets of hyperparameters than Diag, which may suggest that KFAC produces better fits to the true posterior. Instead, we find that KFAC's approximation step for the prior incorporation may result in this phenomena. Concretely, lets define two variants as:
\begin{equation}
\label{sm:sec6:eq:kfacproperprior}
\textsc{N}\boldsymbol{I}_{\text{kfac}} + \tau I = \textsc{N}\left(A \otimes G \right) + \tau I \ \ \text{or}
\end{equation}
\begin{equation}
\label{sm:sec6:eq:kfacprior}
\textsc{N}\boldsymbol{I}_{\text{kfac}} + \tau I \approx \left(\sqrt{\textsc{N}}A_{\mathfrak{i}-1} + \sqrt{\tau}I \right)\otimes \left(\sqrt{\textsc{N}}G_\mathfrak{i} + \sqrt{\tau}I\right).
\end{equation}
Here, equation \ref{sm:sec6:eq:kfacprior} has been the approximation step of \citet{Ritter2017ASL, RitterBB18} while equation \ref{sm:sec6:eq:kfacproperprior} is an exact variant. We denote the later as OKF. By reproducing the results of \citet{Ritter2017ASL}, we depict the results in figure \ref{fig:1:sm:4:2}, in which we plot the predictive uncertainty obtained from OKF and KFAC under the same hyperparameter settings. Furthermore, a direct plot of covariance matrix can be found as well for the same hyperparameters. These results show that without the approximation step in equation \ref{sm:sec6:eq:kfacprior}, KFAC requires higher regularization hyperparameters, as similar as Diag. Looking into the covariance matrix directly, we also find that the magnitude of KFAC is smaller with this approximation. Therefore, our findings are that KFAC, due to the given approximation in incorporation of prior, requires smaller sets of regularization hyperparameters. Furthermore, as OKF does not seem to result in a similar phenomena, it might be difficult to conclude that KFAC, when compared to Diag, produces better fit to the true posterior.

\begin{figure}
\minipage{0.22\textwidth}%
  \includegraphics[width=\linewidth]{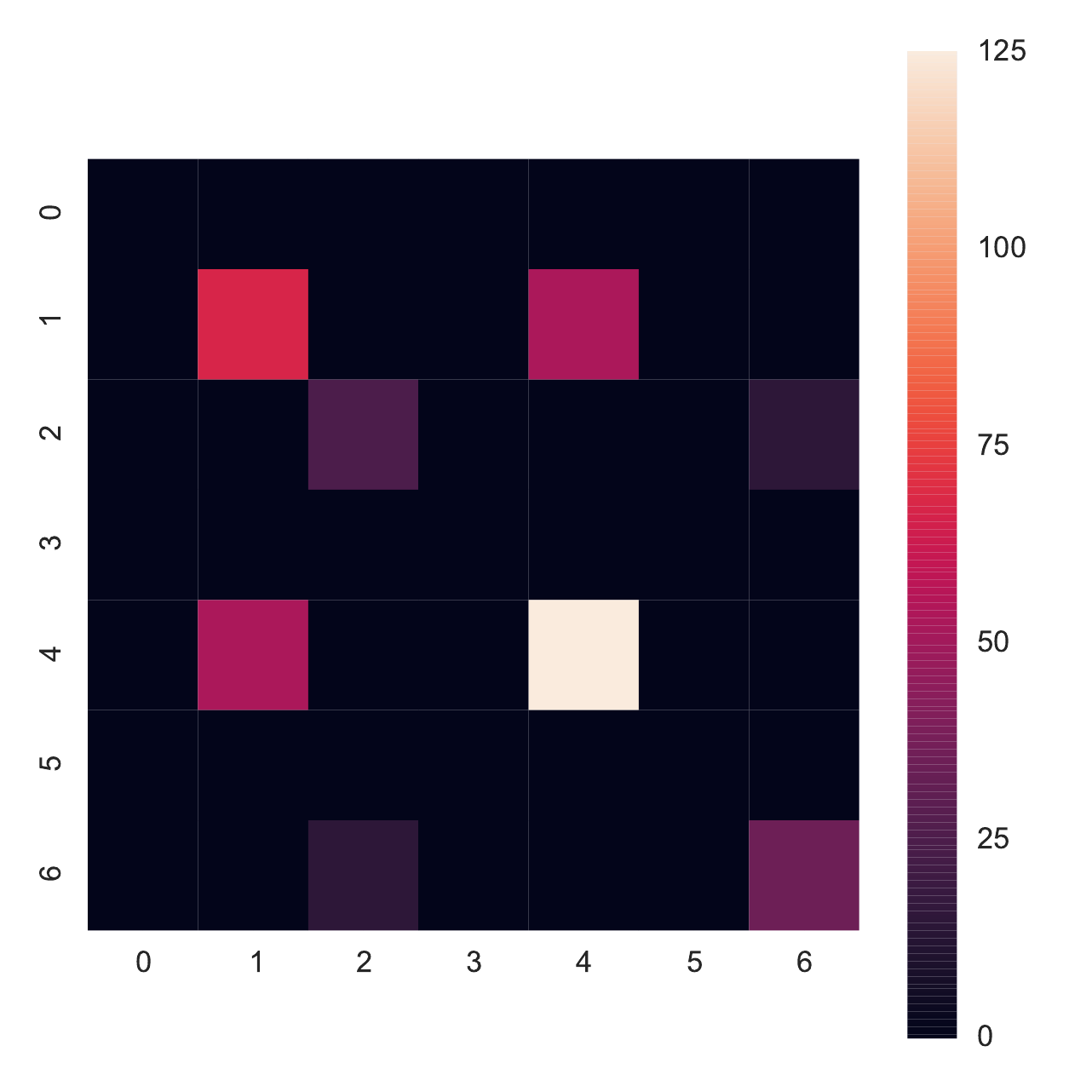}
  \caption{\small{the Hessian [20].}}
\endminipage \hfill
\minipage{0.22\textwidth}
  \includegraphics[width=\linewidth]{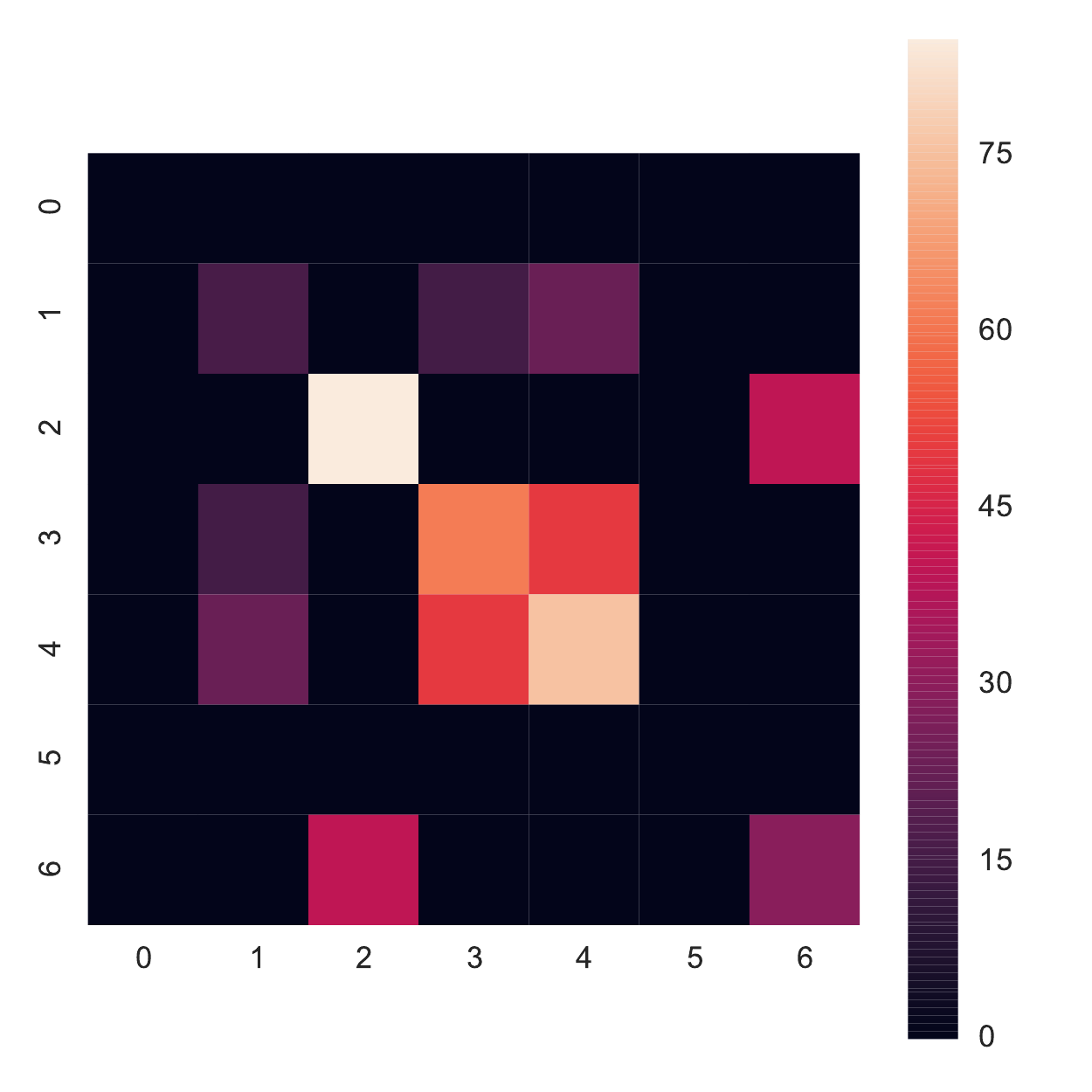}
  \caption{\small{the Hessian [100]}}
\endminipage \hfill
\caption{\textbf{Visualization of the approximate information matrix with different data points.} Only the first layer chosen for the analysis. With increasing data points, the resulting information matrix becomes less degenerate.}
\label{fig:1:sm:4:3}
\end{figure}

\subsubsection{Effects of data points size.}

We now study the effects of dataset size to number of parameters. For this, we compare the dataset size 100 and 20. Results are depicted in figure \ref{fig:1:sm:4:3}. Notably, at using 20 data points resulted in more number of zero diagonal entries and corresponding rows and columns. This might be due to overparameterization of the model which results in under determined Hessian.

\subsection{Effects of hyperparameters - UCI benchmark}
\label{sm:sec:6:hyp}
\begin{figure*}
  \includegraphics[width=\linewidth]{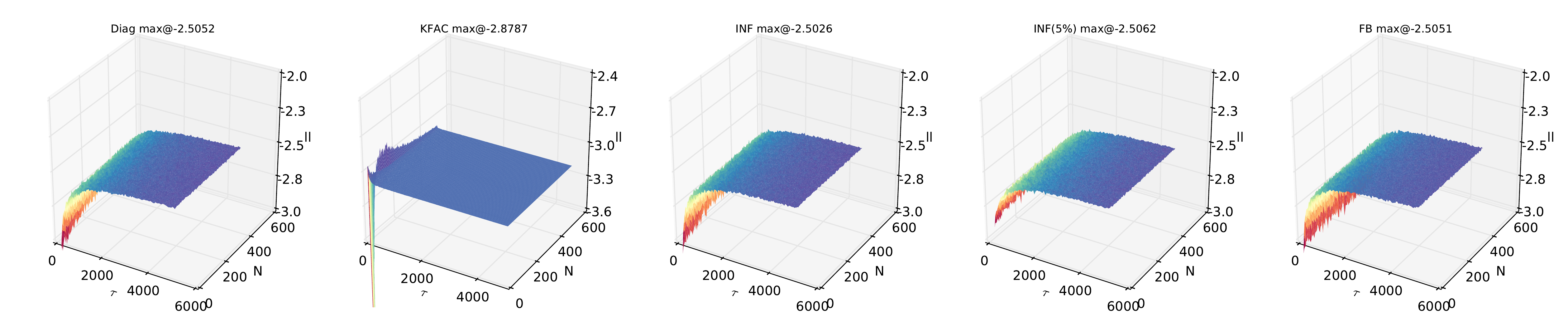}
\caption{\textbf{Boston Hyperparameter Landscape.} Test-log likelihood on the z-axis. Ranging hyperparameters are display on XY plane. Maximum values are also displayed for Diag, KFAC, INF with two different ranks, and FB (true block-diagonal information matrix). Except KFAC, all LA-based approaches show similar behavior. Concrete, energy and protein showed similar tendency. }
\label{fig:boston:hyp_landscape}
\end{figure*}
\begin{figure*}  
\includegraphics[width=\linewidth]{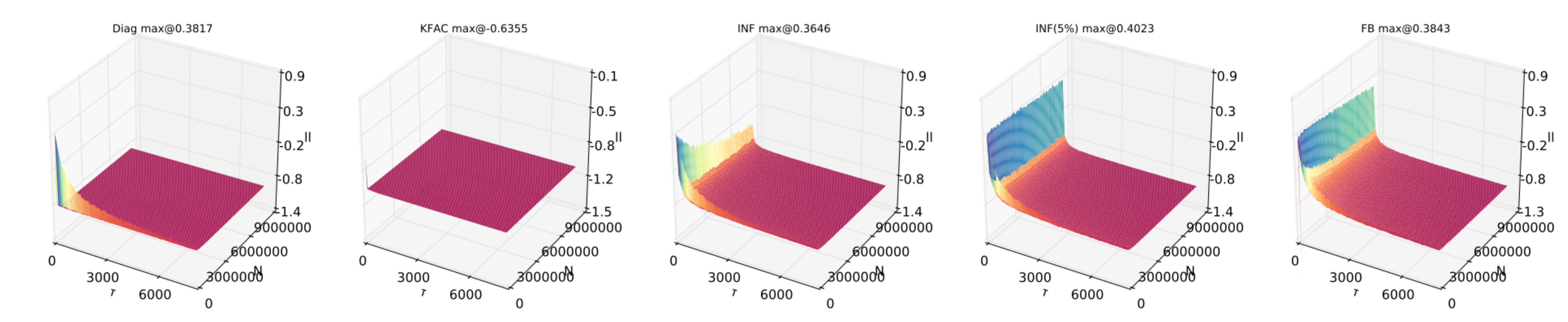}
\caption{\textbf{Kin8nm Hyperparameter Landscape.} Test-log likelihood on the z-axis. Ranging hyperparameters are display on XY plane. Maximum values are also displayed for Diag, KFAC, INF with two different ranks, and FB (true block-diagonal information matrix). All LA-based approaches show different behavior. Naval, power, wine and yacht datasets have similar tendency.}
\label{fig:kin8nm:hyp_landscape}
\end{figure*}

\begin{figure*}
\minipage{0.25\textwidth}%
  \includegraphics[width=\linewidth]{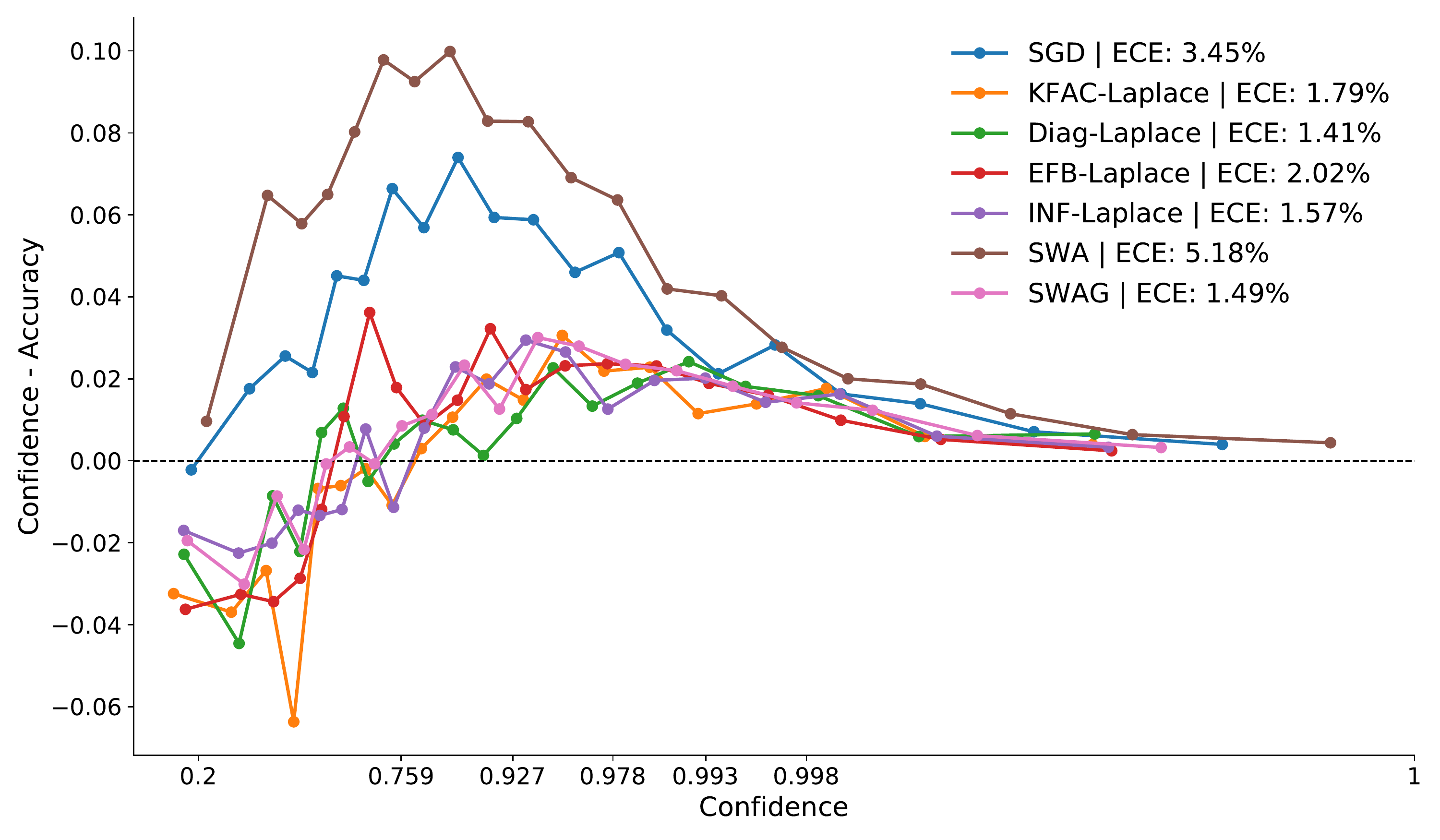}
\endminipage \hfill
\minipage{0.25\textwidth}
  \includegraphics[width=\linewidth]{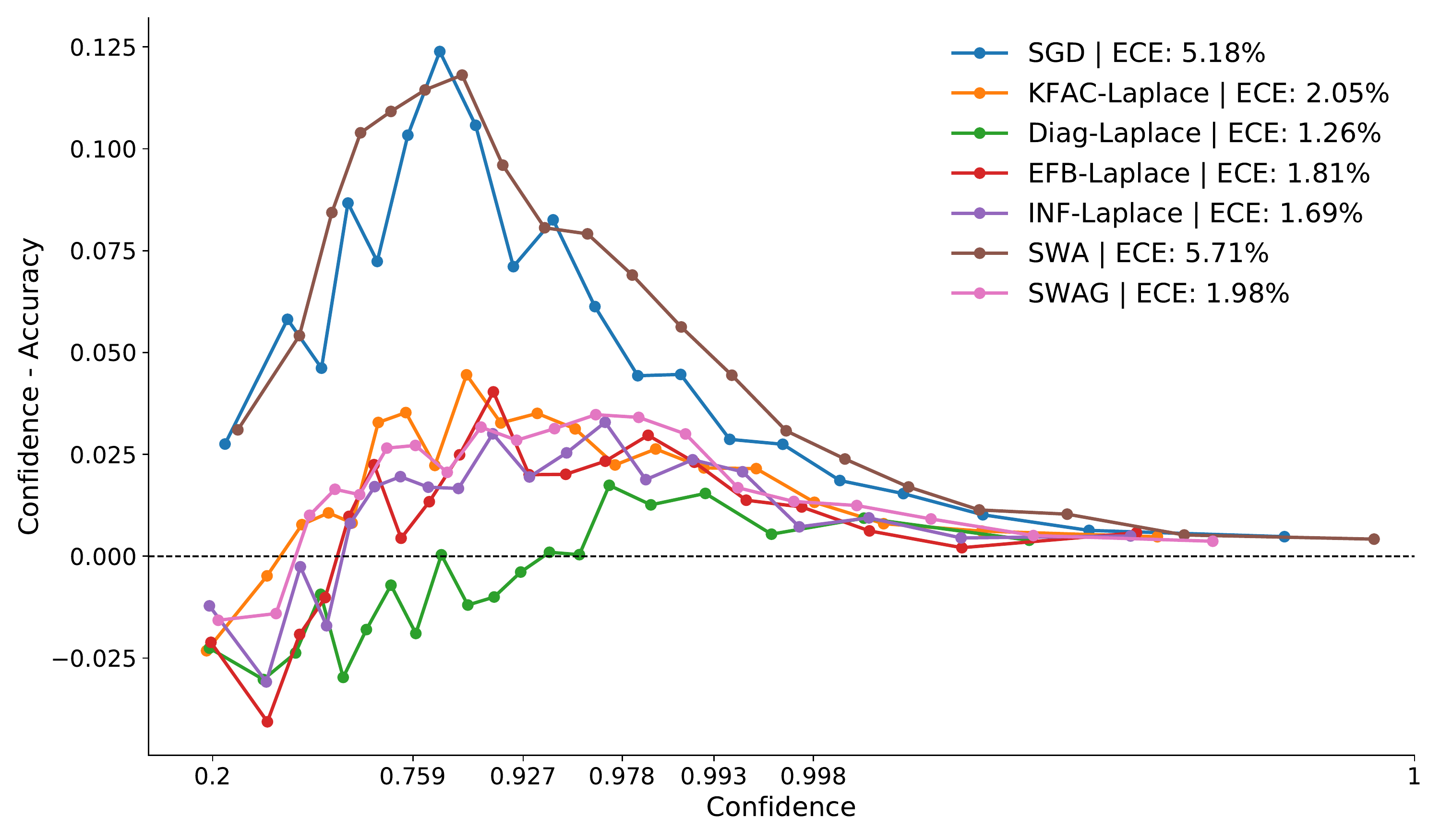}
\endminipage \hfill
\minipage{0.25\textwidth}%
  \includegraphics[width=\linewidth]{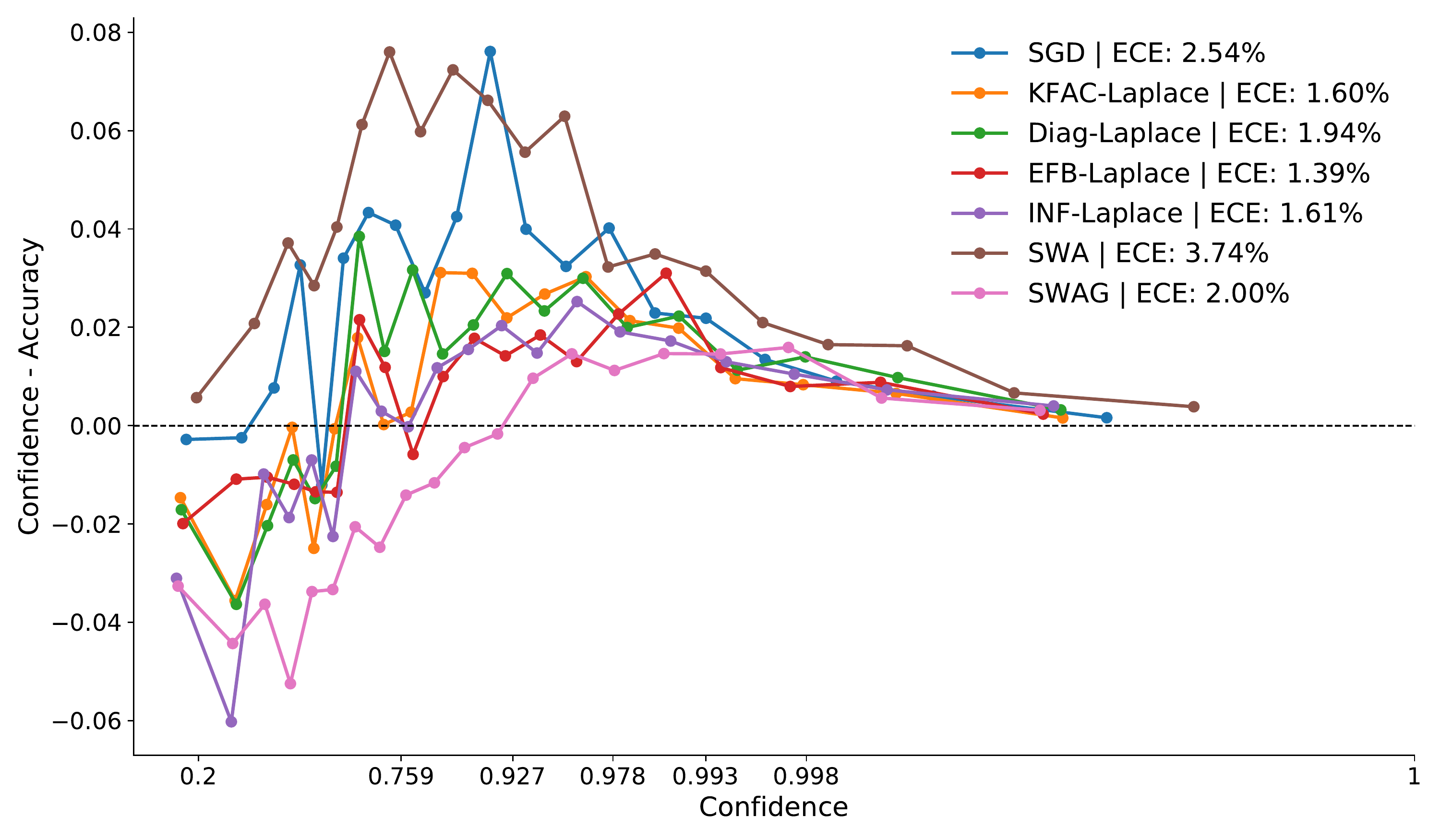}
\endminipage \hfill
\minipage{0.25\textwidth}
  \includegraphics[width=\linewidth]{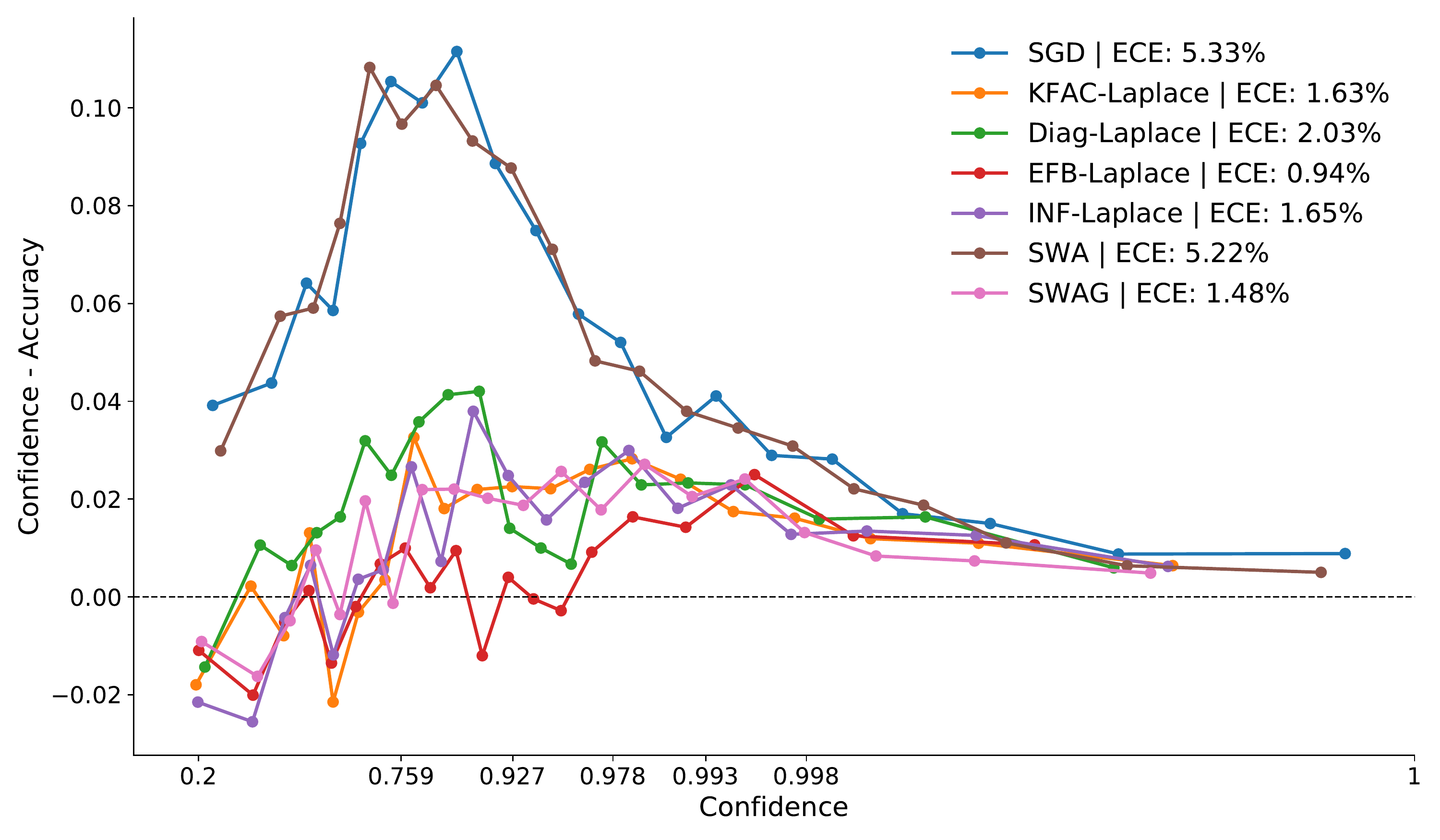}
\endminipage \hfill
\caption{\textbf{Calibration results on large scale experiments:} From left to right: ResNet50, ResNet152, DenseNet121 and DenseNet161. Our method tends to outperform SWA and SWAG while being competitive to other fine tuned LA-based approaches.}
\label{fig:calibration}
\end{figure*}

\begin{figure*}
\minipage{0.25\textwidth}%
  \includegraphics[width=\linewidth]{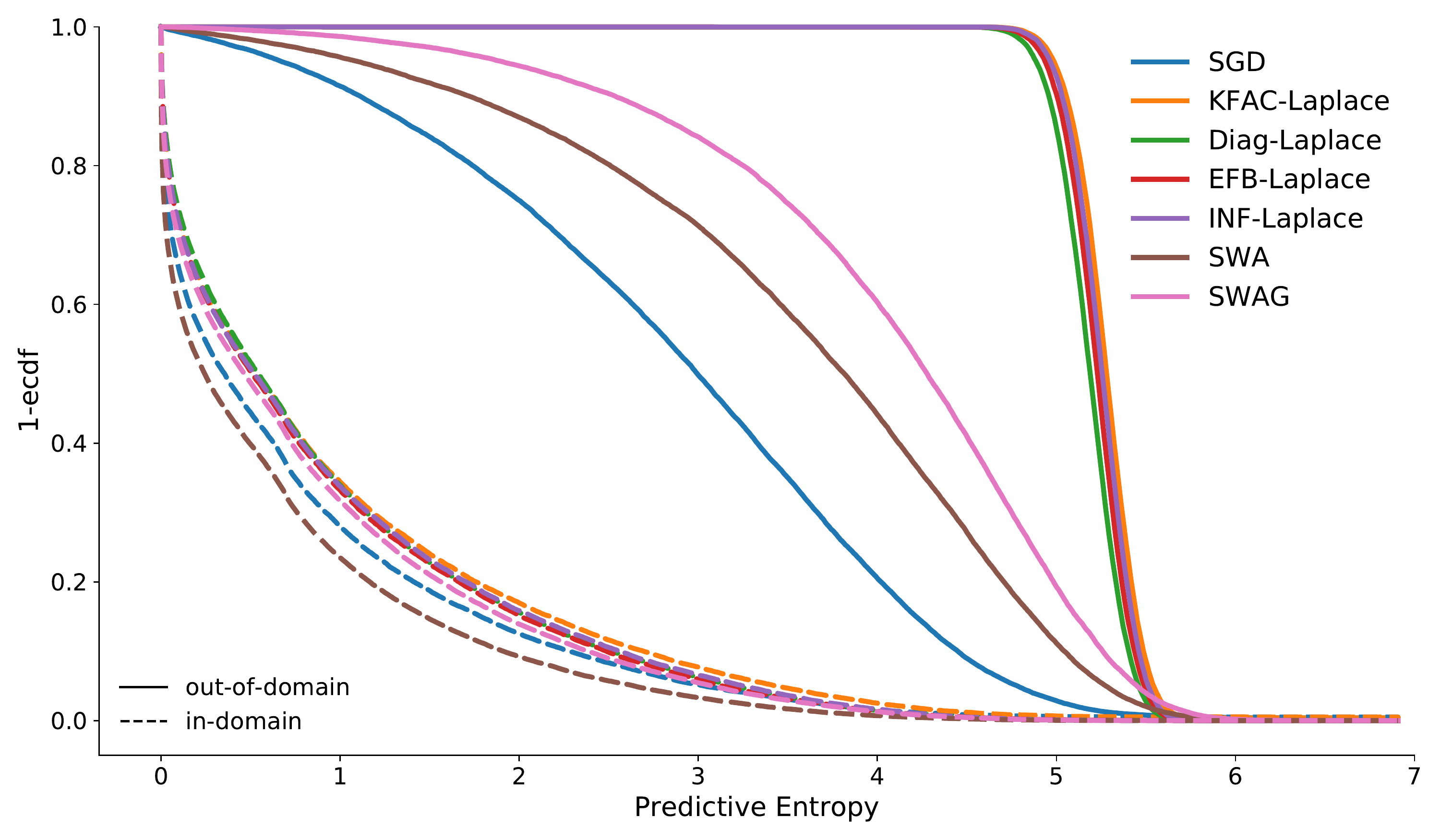}
\endminipage \hfill
\minipage{0.25\textwidth}
  \includegraphics[width=\linewidth]{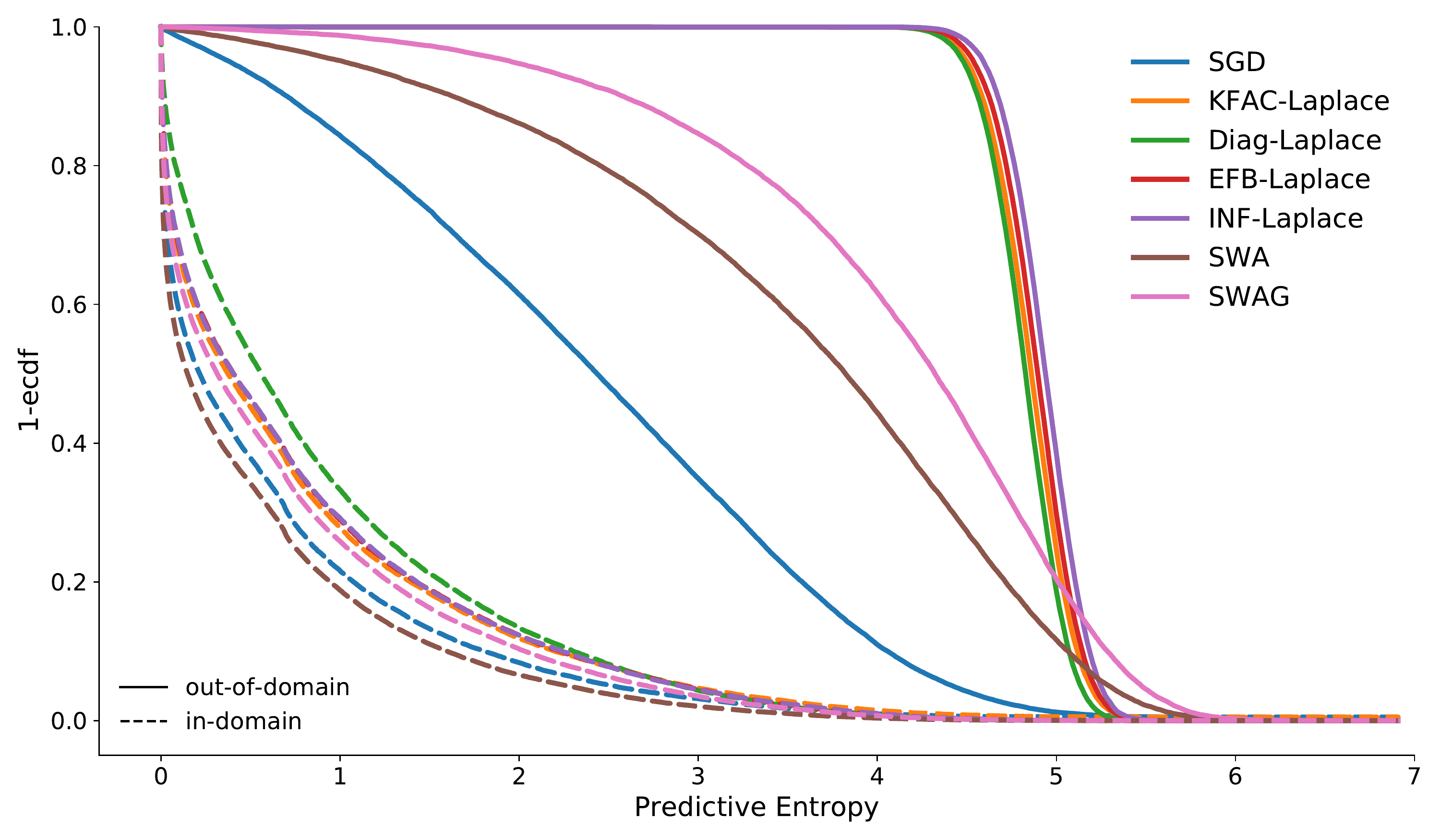}
\endminipage \hfill
\minipage{0.25\textwidth}%
  \includegraphics[width=\linewidth]{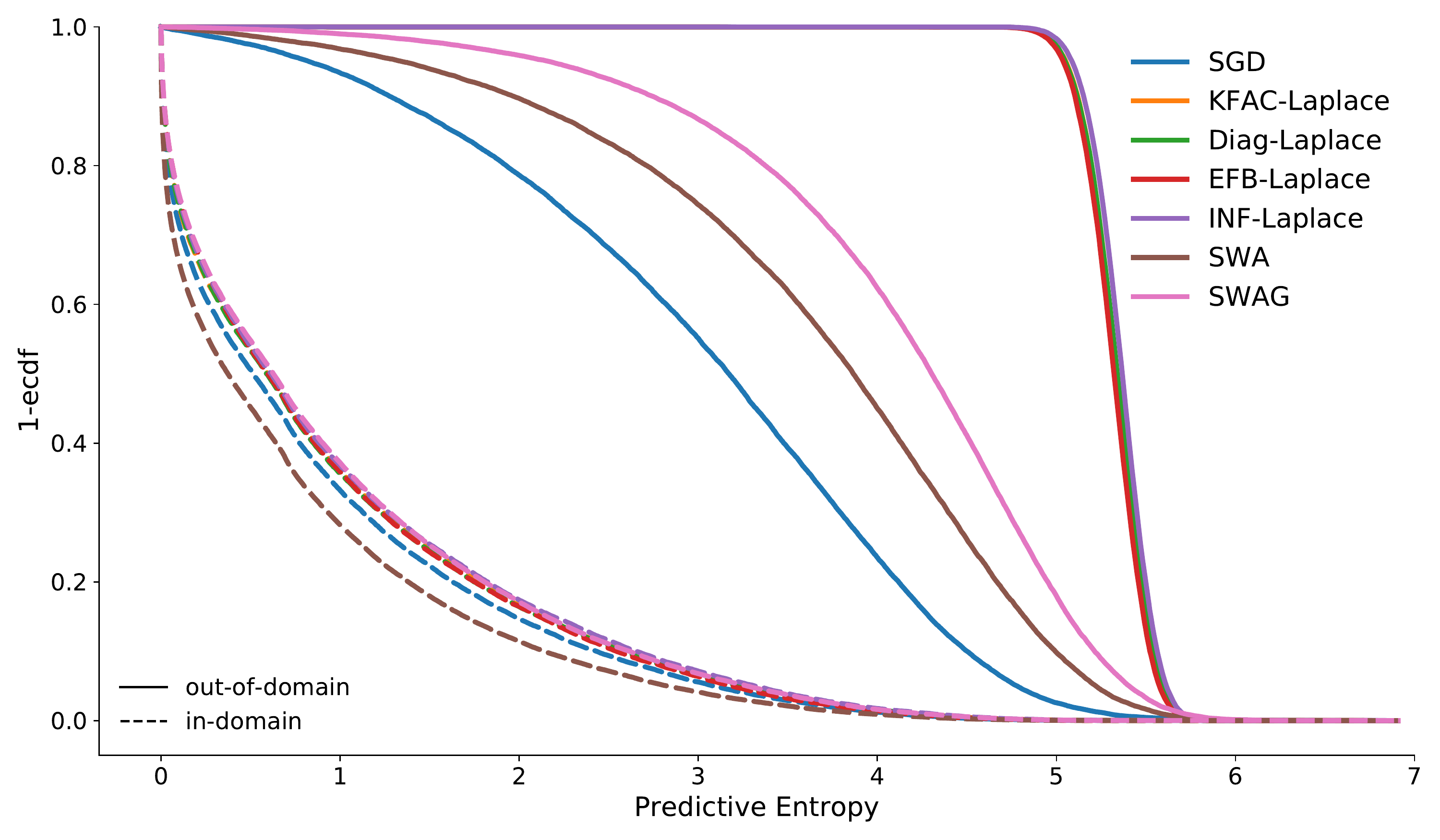}
\endminipage \hfill
\minipage{0.25\textwidth}
  \includegraphics[width=\linewidth]{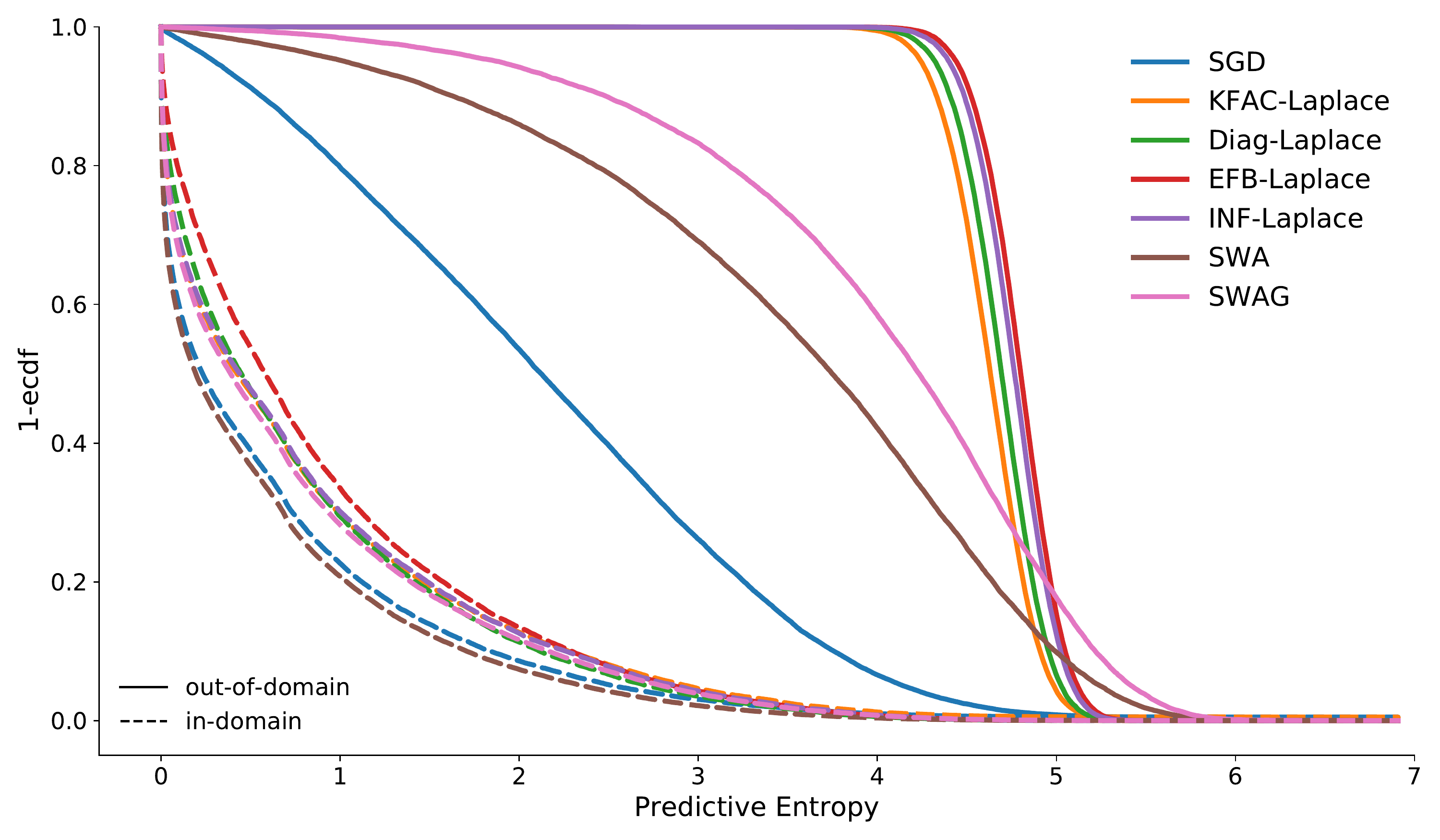}
\endminipage \hfill
\caption{\textbf{Out-of-domain:} From left to right: ResNet50, ResNet152, DenseNet121 and DenseNet161. Our method tends to significantly outperform SWA and SWAG while being competitive to other fine tuned LA-based approaches.}
\label{fig:ood}
\end{figure*}


Instead of linearized LA, we investigate the performance of LA-based methods with a full Bayesian analysis on UCI benchmarks. Rather than reporting the best performance of each methods with a single selected hyperparameter choice, we perform extensive grid searches and show the performance landscape. Such performance landscape can be informative for studying how more accurate approximation of information matrix translates to uncertainty estimation under the effects of hyperparameters. Note that this is possible on UCI datasets due to the small scale of the set up.

To this end, we search 10000 hyperparameters sets for each methods except KFAC, where we increase the size to 20000 hyperparameters \footnote{We have doubled the search space for KFAC as it requires smaller sets of hyperparameters due to equation \ref{sm:sec6:eq:kfacprior} instead of equation \ref{sm:sec6:eq:kfacproperprior}. Following this observation, we further decreased the minimum $\tau$.}. The range of hyperparameters sets have been chosen differently for each datasets so that all the methods produce reliable predictions. We draw 100 $K_{mc}$ samples for each predictions in order to have acceptable range of convergence for monte-carlo integration. 
Results are shown in figures \ref{fig:boston:hyp_landscape} and \ref{fig:kin8nm:hyp_landscape} where we report following observations which falls into two categories.
\begin{itemize}
\item \textbf{Type 1 landscape:} Experiments on datasets namely boston housing, concrete, energy and protein showed similar behaviors. As seen in figure \ref{fig:boston:hyp_landscape}, the performance landscape (test log-likelihood) show similar curves for all the methods except KFAC. The maximum achievable performance have been found also similar with a marginal difference.
\item \textbf{Type 2 landscape:} Experiments on datasets namely kin8nm, naval, power, wine and yacht showed a different tendency than type 1. While no methods significantly outperformed the other uniformly across all the datasets, the curve showed different behavior than type 1 as reported in figure \ref{fig:kin8nm:hyp_landscape}. Each methods also showed different performance landscape.
\end{itemize}

These experiments suggest that for type 1, the benefits from having more accurate Fisher information is marginal. This suggests that improvements on accuracy of IM w.r.t Frobenius norm of error may not directly translate to more accurate uncertainty estimation within this context of LA. 


For type 2 however, interesting differences can be found in the sense that INF variants and FB showed significantly more regimes of hyperparameter sets that outputs higher log-likelihood which can be benefits of having more accurate Fisher information matrix - when only smaller number of hyperparameter searches are possible, more accurate IM can result in better quality of predictive uncertainty. Understanding the causes of these behaviors to full generality seem a challenging research question as LA is tightly coupled with loss landscape of DNNs and further, how optimization affects generality and the shape of true posterior. One possible explanation for type 1 is that maintaining a single $\tau$ and $N$ for all the layers may force all the methods to be regularized for fitting a few sharply peaked local mode of true posterior, hindering the benefits of having more accurate estimates of true Fisher information.

\subsection{Additional ImageNet Results}
\label{sm:sec:ImageNet}
%

The calibration and OOD detection experiments presented in the main text on ResNet18 were performed identically for the four additional architectures. We show the results in figures \ref{fig:calibration} and \ref{fig:ood}. The observations from the main text hold for the additional networks. All LA-based approaches can reduce the calibration error significantly compared to the deterministic network and SWA and are as good or better than SWAG. In out-of-domain separation, we find that the LA-based approaches perform comparably strong and are far superior to the other methods across all considered networks. 

\begin{table}[ht]
\scriptsize
\centering
\caption{\textbf{Wall clock time analysis on sampling.} Mean and standard deviation over 1000 draws are reported with a single thread.}
\label{results:sm:img:complexity1}
\begin{tabular}{ccccc}
\toprule
Architecture       &  \textbf{Diag [ms]}       & \textbf{KFAC [ms]}    & \textbf{EFB [ms]}   & \textbf{INF [ms]} \\
\midrule
\textit{ResNet18}  &  1.24 $\pm$ 0.06  & 8.23 $\pm$ 0.04  &  9.28 $\pm$ 0.06  & 4.74 $\pm$ 0.08 \\
\textit{ResNet50}  &  1.94 $\pm$ 0.11  & 15.47 $\pm$ 0.18  &  16.89 $\pm$ 0.09  & 12 $\pm$  0.25 \\
\textit{ResNet152}  &  5.4 $\pm$ 0.07  & 32.08 $\pm$  0.5  &  35.55 $\pm$ 0.07  & 32.62 $\pm$ 0.15 \\
\textit{DenseNet121}  &  4.41 $\pm$ 0.14  & 8.92 $\pm$ 0.19  &  10.22 $\pm$ 0.13  & 25.03 $\pm$ 0.65 \\
\textit{DenseNet161}  &  5.86 $\pm$  0.11  & 16.48$\pm$ 0.03  &  18.84 $\pm$ 0.54  & 35.95 $\pm$ 0.35 \\
\bottomrule
\end{tabular}
\end{table}

Table \ref{results:sm:img:complexity1} also the wall clock analysis for sampling. Interestingly, for ResNet variants, INF is more efficient than KFAC and EFB due to the effects of low rank approximation. On the other hand, DenseNet variants have many small layers and therefore, rank reduction is less noticeable and cannot outweigh the disadvantage of having a more number of smaller operations in a sampling procedure. While KFAC and EFB maintain similar size matrices, EFB sampling is slower than KFAC, also due to more number of operations. Diag is as expected, the most efficient method. We note however, that Bayesian Neural Networks in general, has a disadvantage that prediction time is atleast 30 times slower (assuming 30 samples are taken) and thus, there may not be any practical advantages.

\begin{table}[ht]
\scriptsize
\centering
\caption{\textbf{Wall clock time analysis on information matrix computation.} Values are rounded to the nearest.}
\label{results:sm:img:complexity2}
\begin{tabular}{ccccc}
\toprule
Architecture       &  \textbf{Diag [min]}       & \textbf{KFAC [min]}    & \textbf{EFB [min]}   & \textbf{INF [min]} \\
\midrule
\textit{ResNet18}  & 30  & 120  & 165 & 165     \\
\textit{ResNet50}  & 82  & 210 & 	300 & 300    \\
\textit{ResNet152}  & 180  & 510  & 720 & 720    \\
\textit{DenseNet121}  & 100 & 360  & 465 & 465    \\
\textit{DenseNet161}  & 180  & 870 & 1060  & 1060 \\
\bottomrule
\end{tabular}
\end{table}

Next, table \ref{results:sm:img:complexity2} reports the wall clock analysis for IM computations. In our implementation, EFB is computed after having KFAC and therefore, it takes more time than KFAC. Original implementation of EFB contains amortize eigendecomposition and can be made more efficient than KFAC. INF is an offline procedure, and provides negligible overhead to EFB. The total computation time for all the methods are less than a day on ImageNet, and thus, this analysis shows practicality and scalability of LA-based approaches.